\documentclass[10pt,twocolumn,letterpaper]{article}

\usepackage[pagenumbers]{cvpr} 

\usepackage[dvipsnames]{xcolor}

\definecolor{cvprblue}{rgb}{0.21,0.49,0.74}
\usepackage[pagebackref,breaklinks,colorlinks,citecolor=cvprblue]{hyperref}
\usepackage{multirow}
\usepackage{makecell}
\usepackage{bbding}
\usepackage{amsthm}
\usepackage{xfrac}
\usepackage{colortbl}

\newtheorem{observation}{Observation}
\newcommand{\first}[1]{\textcolor{Red}{\textbf{#1}}}
\newcommand{\second}[1]{\textcolor{Green}{\textbf{#1}}}
\newcommand{\third}[1]{\textcolor{Cyan}{\textbf{#1}}}
\def\paperID{2050} 
\def\confName{CVPR}
\def\confYear{2024}

\title{NeRF-HuGS: Improved Neural Radiance Fields in Non-static Scenes \\ Using Heuristics-Guided Segmentation\vspace{-3mm}}
\author{Jiahao Chen$^{1}$ \quad\quad Yipeng Qin$^{2}$ \quad\quad Lingjie Liu$^{3}$ \quad\quad Jiangbo Lu$^{4}$ \quad\quad Guanbin Li$^{1}$\thanks{Corresponding author is Guanbin Li. This work was supported in part by the National Natural Science Foundation of China (NO.~62322608), in part by the CAAI-MindSpore Open Fund, developed on OpenI Community, in part by the Open Project Program of State Key Laboratory of Virtual Reality Technology and Systems, Beihang University (No.VRLAB2023A01), in part by Shenzhen Science and Technology Program KQTD20210811090149095 and also the Pearl River Talent Recruitment Program 2019QN01X226.} \vspace{2mm}\\
$^1$Sun Yat-sen University\ \ $^2$Cardiff University\ \ $^3$University of Pennsylvania\ \ $^4$SmartMore Corporation\\
{\tt\small chenjh328@mail2.sysu.edu.cn}, {\tt\small qiny16@cardiff.ac.uk}, {\tt\small lingjie.liu@seas.upenn.edu} \\ {\tt\small jiangbo.lu@gmail.com}, {\tt\small liguanbin@mail.sysu.edu.cn}
\vspace{-3mm}
}

\begin{document}
\maketitle
\begin{abstract}
Neural Radiance Field (NeRF) has been widely recognized for its excellence in novel view synthesis and 3D scene reconstruction. However, their effectiveness is inherently tied to the assumption of static scenes, rendering them susceptible to undesirable artifacts when confronted with transient distractors such as moving objects or shadows.
In this work, we propose a novel paradigm, namely ``Heuristics-Guided Segmentation'' (HuGS), which significantly enhances the separation of static scenes from transient distractors by harmoniously combining the strengths of hand-crafted heuristics and state-of-the-art segmentation models, thus significantly transcending the limitations of previous solutions. 
Furthermore, we delve into the meticulous design of heuristics, introducing a seamless fusion of Structure-from-Motion (SfM)-based heuristics and color residual heuristics, catering to a diverse range of texture profiles. Extensive experiments demonstrate the superiority and robustness of our method in mitigating transient distractors for NeRFs trained in non-static scenes. Project page: \url{https://cnhaox.github.io/NeRF-HuGS/}
\vspace{-6mm}
\end{abstract}    
\section{Introduction}
\label{sec:intro}

Neural Radiance Fields (NeRF)~\cite{mildenhall2020nerf} have garnered significant attention for their remarkable achievements in novel view synthesis. Utilizing multiple-view images, NeRF conceptualizes the 3D scene as a neural field~\cite{xie2022neural} and produces highly realistic renderings through advanced volume rendering techniques.
This capability has opened the door to a wide array of downstream applications including 3D reconstruction~\cite{wang2021neus,sun2022neural,li2023neuralangelo}, 
content generation~\cite{niemeyer2021giraffe,poole2022dreamfusion,lin2023magic3d}, 
semantic understanding~\cite{zhi2021place,kerr2023lerf,siddiqui2023panoptic}, etc.

\begin{figure}[t]
    \centering
    \includegraphics[width=\linewidth]{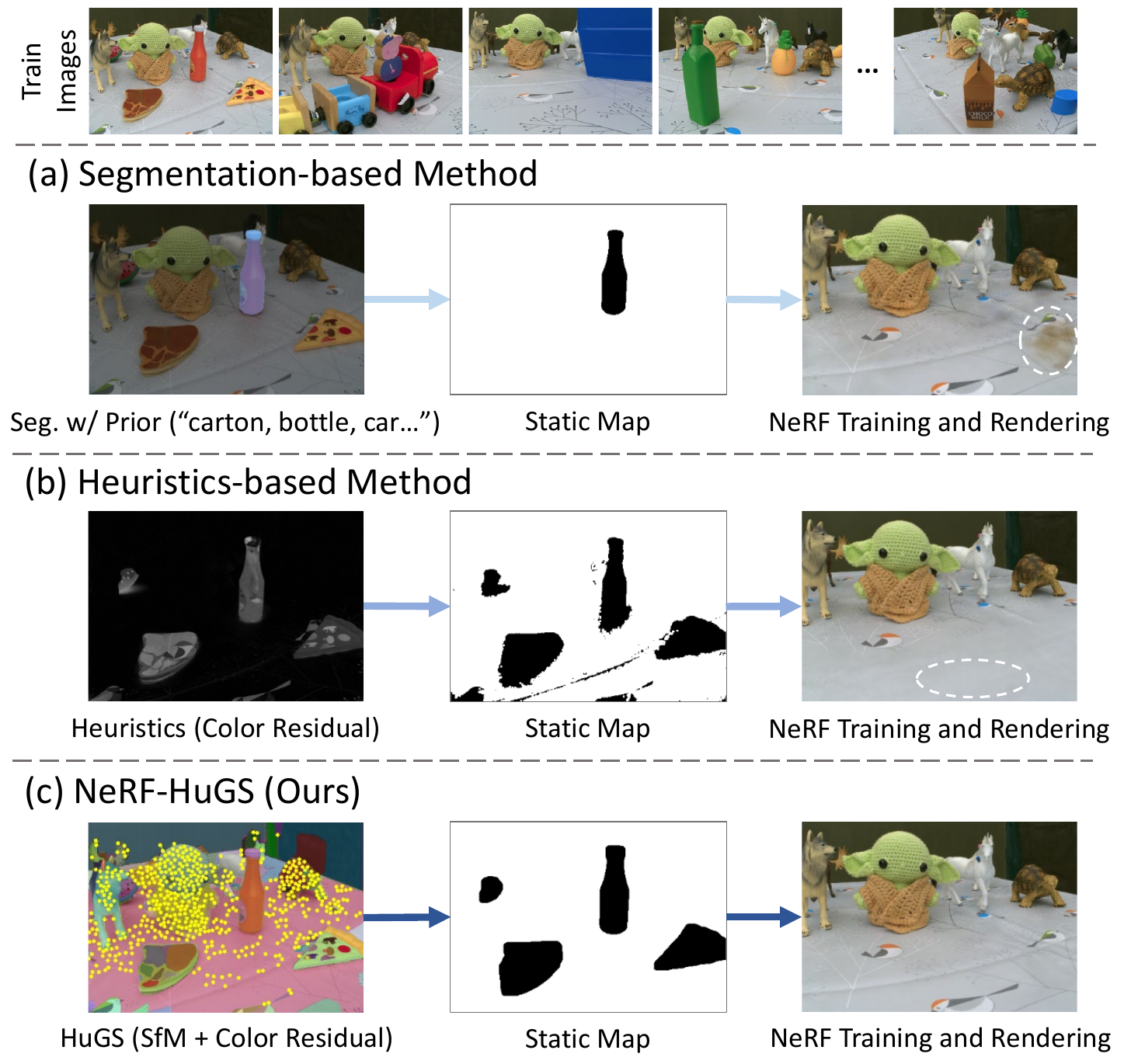}
    \caption{\textbf{Comparison between previous methods and the proposed Heuristics-Guided Segmentation (HuGS) paradigm.} When training NeRF with static scenes disturbed by transient distractors, (a) segmentation-based methods rely on prior knowledge and cannot identify unexpected transient objects (\eg, pizza); (b) heuristics-based methods are more generalizable but inaccurate (\eg, tablecloth textures); (c) our method combines their strengths and produces highly accurate transient \vs static separations, thereby significantly improving NeRF results.\vspace{-6mm}}
    \label{fig:teaser}
\end{figure}

However, the images used as NeRF training data must meet several strict conditions, one of which is the requirement for content consistency and stability. 
In other words, the native NeRF model operates under the assumption of a static scene. 
Any elements that exhibit motion or inconsistency throughout the entire data capture session, which we refer to as ``{\it transient distractors}'', can introduce undesirable artifacts into the reconstructed 3D geometry.
However, the presence of transient distractors is nearly inevitable in real-world scenarios. For instance, in outdoor settings, random appearances of pedestrians and vehicles may occur during image acquisition, while indoor shooting may be affected by shadows cast by the photographer. Furthermore, manually removing these transient distractors from a substantial number of images is a challenging and time-consuming task, often necessitating pixel-by-pixel labeling.

To mitigate the effects of transient distractors, previous research has explored two main paradigms. One paradigm involves leveraging pre-trained segmentation models to detect transient distractors~\cite{tancik2022block,turki2022mega,rematas2022urban,sun2022neural,liu2023robust,karaoglu2023dynamon}. 
Although it can produce accurate results, this approach exhibits limited generality as it relies on additional prerequisites of prior knowledge (\eg, semantic classes of transient objects).
The other strategy aims to separate transient distractors from static scenes through the application of hand-crafted heuristics~\cite{martin2021nerf,chen2022hallucinated,lee2023semantic,sabour2023robustnerf,li2023nerf,kim2023upnerf}. 
Nonetheless, this approach often yields imprecise or erroneous results, primarily attributed to the intricate nature of heuristic design and the inherent ill-posedness of existing heuristics.

In this work, we take the best of both worlds and propose a novel paradigm called ``{\it Heuristics-Guided Segmentation}'' (HuGS) to maximize the accuracy of static {\it vs.} transient object identification for NeRF in non-static scenes (Fig.~\ref{fig:teaser}). 
The rationale of our approach lies in the principle embodied in the British idiom ``horses for courses'', emphasizing the alignment of talents with tasks. 
Specifically, our paradigm harnesses the collective strengths of i) hand-crafted heuristics, adept at discerning rough indicators of static elements, and ii) contemporary segmentation models, like the Segment Anything Model (SAM)~\cite{kirillov2023segment} renowned for their ability to delineate precise object boundaries. 
Furthermore, we delve deeply into the design of heuristics and suggest a seamless fusion of i) our newly devised Structure-from-Motion (SfM)-based heuristics, which efficiently identify static objects characterized by high-frequency texture patterns, with ii) the color residual heuristics derived from a partially trained Nerfacto~\cite{tancik2023nerfstudio}, which excel at detecting static elements marked by low-frequency textures.
This tailored integration of heuristics empowers our method to robustly encompass the full spectrum of static scene elements across a diverse range of texture profiles. Extensive experiments have demonstrated the superiority of our method. Our contributions can be summarized as follows:
\begin{itemize}
    \item We propose a novel paradigm called ``Heuristics-Guided Segmentation'' for improving NeRF trained in non-static scenes, which takes the best of both hand-crafted heuristics and state-of-the-art segmentation models to accurately distinguish static scenes from transient distractors.
    \item We delve into heuristic design and propose the seamless fusion of SfM-based heuristics and color residual ones to capture a wide range of static scene elements across various texture profiles, offering robust performance and superior results in mitigating transient distractors.
    \item Extensive experimental results show that our method produces sharp and accurate static {\it vs.} transient separation results close to the ground truth, and significantly improves NeRFs trained in non-static scenarios.
\end{itemize}

\begin{figure*}
  \centering
  \includegraphics[width=\linewidth]{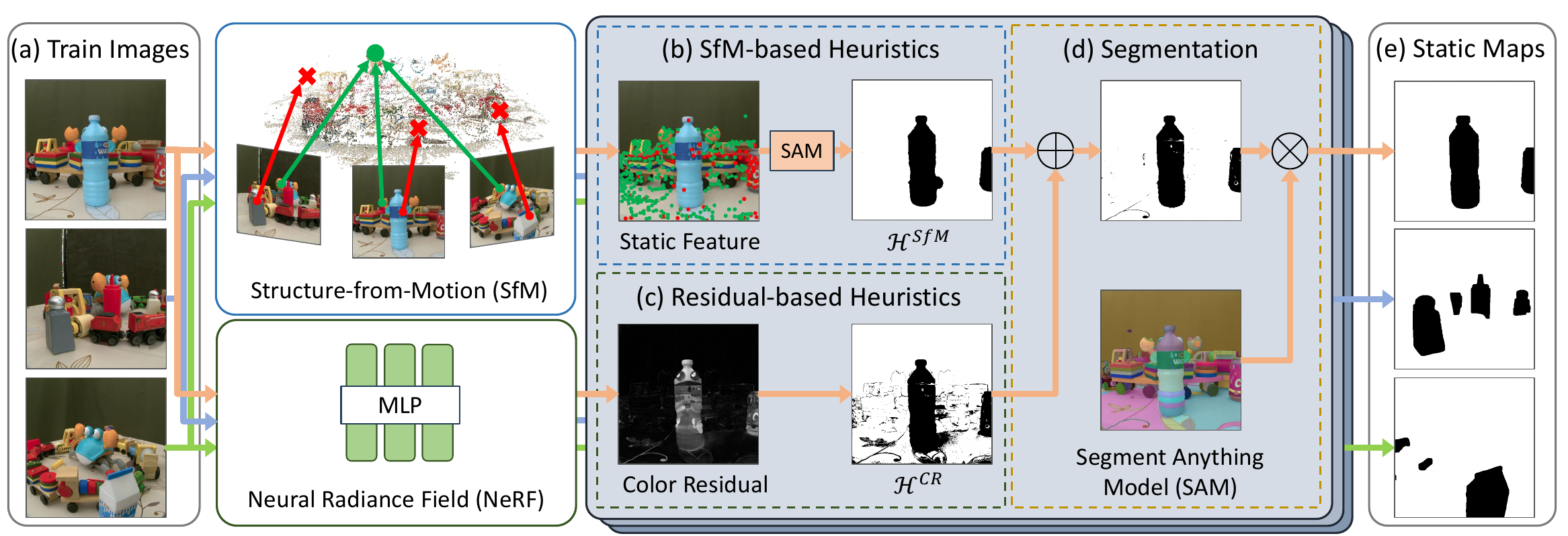}
  \caption{\textbf{Pipeline of HuGS.} (a) Given unordered images of a static scene disturbed by transient distractors as input, we first obtain two types of heuristics. (b) SfM-based heuristics use SfM to distinguish between static (\textcolor{Green}{green}) and transient features (\textcolor{red}{red}). 
  The static features are then employed as point prompts to generate dense masks using SAM. 
  (c) Residual-based heuristics are based on a partially trained NeRF (\ie, trained for several thousands of iterations) that can provide reasonable color residuals. 
  (d) Their combination finally guides SAM again to generate (e) the static map for each input image.\vspace{-6mm}}
  \label{fig:method_pipeline}
\end{figure*}

\section{Related Work}
\label{sec:related}

NeRF~\cite{mildenhall2020nerf} has recently emerged as a promising solution to synthesizing novel photo-realistic views from multiple images, which is a long-standing problem in computer vision.
Although numerous methods~\cite{barron2021mip,barron2022mip2,muller2022instant,barron2023zip,zhang2020nerf++} have been proposed to improve its synthesis quality and training efficiency, most of them assume that the scenes to be reconstructed are static and are therefore not suitable for many real-world scenes (\eg, tourist attractions).

\vspace{1mm}
\noindent \textbf{NeRF in Non-static Scenes.}
In general, there are two major types of non-static scenes that present challenges for NeRF:
i) Dynamic scenes that change over time, where the model needs to render consistent novel views of the scene as it evolves~\cite{gao2021dynamic,li2021neural,park2021nerfies,li2022neural,wu2022d,liu2023robust,park2023temporal}, \eg, scenes with moving objects or environmental effects like lighting or weather changes.
ii) Static scenes disturbed by transient distractors, where the model should exclude dynamic objects like tourists walking through static attractions as background scenes. 
Our work focuses on ii), where existing solutions can be roughly grouped into two main paradigms:
\begin{itemize}
    \item {\it Segmentation-based methods}~\cite{tancik2022block,turki2022mega,rematas2022urban,sun2022neural,liu2023robust,karaoglu2023dynamon} use pre-trained semantic or video segmentation models to identify transient distractors {\it vs.} static scenes and use the information obtained to facilitate NeRF training.
    These models can produce accurate results but have some key limitations: i) They require additional priors like the semantic class of transient distractors or the temporal relationships of the images as video frames, which are hard to satisfy in practice as it is intractable to enumerate all possible distractor classes, and images may be unordered.
    ii) Semantic segmentation cannot distinguish between static and transient objects of the same class.
    \item {\it Heuristics-based methods}~\cite{martin2021nerf,chen2022hallucinated,lee2023semantic,sabour2023robustnerf,li2023nerf,kim2023upnerf} use hand-crafted heuristics to separate transient distractors from static scenes during NeRF training, making themselves more generalizable as they require no prior.
    However, heuristics that enable accurate separation are difficult to design.
    For example, NeRF-W~\cite{martin2021nerf} observes that the density of transient objects is usually small and uses this to regularize NeRF training. However, it can easily produce foggy residuals with small densities that are not transient objects.
    RobustNeRF~\cite{sabour2023robustnerf} distinguishes transient pixels from static ones through color residuals as transient pixels are more difficult to fit during NeRF training.
    However, high-frequency details of static objects are also difficult to fit, causing RobustNeRF to easily ignore them when dealing with transient distractors.
\end{itemize}
In our work, we propose a new paradigm called HuGS that takes the best of both worlds. In short, we match talents to tasks and propose to use heuristics only as rough cues to guide the segmentation and produce highly accurate transient {\it vs.} static separations that are close to the ground truth. 
We also investigate heuristics design and propose to use a combination of heuristics based on color residuals and SfM.

\vspace{1mm}
\noindent \textbf{SfM in NeRF.}
SfM is a technique for reconstructing the corresponding 3D geometry from a set of 2D images.
In NeRF, SfM is typically used to estimate the camera pose of an image.
Recent works have also used it to estimate the scene depth~\cite{sun2022neural}, locate target objects~\cite{wang2023autorecon,yin2023or} or initialize the set of 3D Gaussians~\cite{kerbl20233d}.
In addition to estimating camera poses, our method also uses SfM to design novel heuristics for static {\it vs.} transient object identification. Specifically, we leverage the insight that only feature points belonging to static scene elements can be reliably matched and triangulated across multiple views in the SfM pipeline. 
To the best of our knowledge, we are the first to exploit this property of SfM for NeRF in non-static scenes. 

\section{Preliminaries}
\label{sec:preliminary}

Let $\mathcal{I}=\left\{I_i\left\vert\ i=1, 2, \dots, N_I\right.\right\}$ be a set of multi-view input images with transient objects, we have:

\vspace{2mm}
\noindent \textbf{Structure-from-Motion (SfM).}
SfM first extracts a set of 2D local feature points $\mathcal{F}_i$ for each $I_i$:
\begin{equation}
    \mathcal{F}_i=\left\{\left(\mathbf{x}_{i}^j, \mathbf{f}_{i}^j\right)\left\vert\ j=1, 2, \dots, N_{F_i}\right.\right\},
\end{equation}
where $\mathbf{f}_{i}^j$ is an appearance descriptor and $\mathbf{x}_{i}^j\in\mathbb{R}^2$ denotes its coordinates in $I_i$.
Then, SfM uses the $\mathcal{F}_i$ of all images to reconstruct a sparse point cloud $\mathcal{C}$ representing the 3D structure of the target scene, where the correspondence between 2D feature points (\ie, {\it matching} points) in different images is determined by whether or not they correspond to the same 3D point in $\mathcal{C}$.
For each 2D feature point $\left(\mathbf{x}_{i}^j, \mathbf{f}_{i}^j\right)$, we denote the number of its {\it matching} points in $\mathcal{I}$ as $n_i^j$.

\vspace{2mm}
\noindent \textbf{Neural Radiance Field (NeRF).}
In short, NeRF represents a static scene with a multi-layer perceptron (MLP) parameterized by $\boldsymbol{\theta}$.
Specifically, given a 3D position $\mathbf{p}\in \mathbb{R}^3$ and its viewing direction $\mathbf{d}\in \mathbb{S}^2$, NeRF outputs its corresponding color $\mathbf{c}\in \mathbb{R}^3$ and density $\sigma \in \mathbb{R}$ as:
\begin{equation}
    (\mathbf{c}, \sigma) = \mathop{\text{MLP}}(\mathbf{p}, \mathbf{d}; \boldsymbol{\theta}).
\end{equation}
This allows NeRF to render each pixel color $\hat{\mathbf{C}}(\mathbf{r})$ in a 2D projection by applying volume rendering along its corresponding camera ray $\mathbf{r}$ with multiple sample points.
During training, the parameters $\boldsymbol{\theta}$ are optimized by minimizing the error between $\hat{\mathbf{C}}(\mathbf{r})$ and the ground truth color $\mathbf{C}(\mathbf{r})$ in input images using loss function:
\begin{equation}
    \mathcal{L}(\mathbf{r}) = \mathcal{L}_{\text{recon}}(\hat{\mathbf{C}}(\mathbf{r}),\mathbf{C}(\mathbf{r})),
    \label{eq:nerf}
\end{equation}
where $\mathcal{L}_{\text{recon}}$ is a reconstruction loss whose popular choices include the MSE loss and the Charbonnier loss~\cite{charbonnier1994two}.

\vspace{2mm}
\noindent \textbf{NeRF in Static Scenes.}
Let $M_i$ be the static map corresponding to image $I_i$ which labels pixels of transient objects with 0 and pixels of static objects with 1, we modify Eq.~\ref{eq:nerf} in a straightforward way by using $M_i$ as the loss weight to avoid the interference of transient pixels:
\begin{equation}
    \mathcal{L}(\mathbf{r}) = M(\mathbf{r})\mathcal{L}_{\text{recon}}(\hat{\mathbf{C}}(\mathbf{r}),\mathbf{C}(\mathbf{r})), 
    \label{eq:nerf_static_scenes}
\end{equation}
where we omit $i$ and use $\mathbf{r}$ instead for simplicity.

\section{Method}
\label{sec:methods}

\begin{figure}
  \centering
  \begin{subfigure}{0.315\linewidth}
    \includegraphics[width=\linewidth]{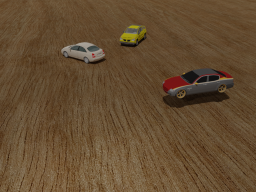}
    \caption{Training Image}
    \label{fig:sam_with_existing_method_img}
  \end{subfigure}
  \hfill
  \rotatebox{90}{\scriptsize ~~~~~~~~~~~~~~~~NeRF-W}
  \hfill
  \begin{subfigure}{0.315\linewidth}
    \includegraphics[width=\linewidth]{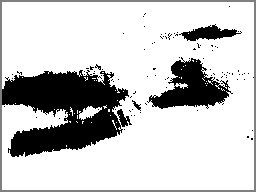}
    \caption{Heuristics}
    \label{fig:sam_with_existing_method_nerfw_heuristic}
  \end{subfigure}
  \hfill
  \begin{subfigure}{0.315\linewidth}
    \includegraphics[width=\linewidth]{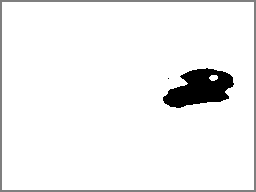}
    \caption{Static Map}
    \label{fig:sam_with_existing_method_nerfw_static}
  \end{subfigure}\\
  \begin{subfigure}{0.315\linewidth}
    \includegraphics[width=\linewidth]{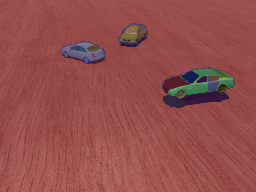}
    \caption{Seg. Masks}
    \label{fig:sam_with_existing_method_sam}
  \end{subfigure}
  \hfill
  \rotatebox{90}{\scriptsize ~~~~~~~~~~~~~RobustNeRF}
  \hfill
  \begin{subfigure}{0.315\linewidth}
    \includegraphics[width=\linewidth]{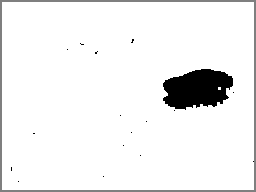}
    \caption{Heuristics}
    \label{fig:sam_with_existing_method_robust_heuristic}
  \end{subfigure}
  \hfill
  \begin{subfigure}{0.315\linewidth}
    \includegraphics[width=\linewidth]{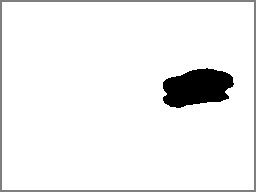}
    \caption{Static Map}
    \label{fig:sam_with_existing_method_robust_static}
  \end{subfigure}
  \caption{\textbf{Performance of HuGS using existing heuristics.} (a) is an example training image with a moving red car, and (d) is its segmentation result using SAM. (b, e) are heuristic maps obtained from different partially trained models. (c, f) are static maps produced by our method, where inaccurate heuristics may lead to incorrect results (NeRF-W).\vspace{-6mm}}
  \label{fig:sam_with_existing_method}
\end{figure}

As Eq.~\ref{eq:nerf_static_scenes} implies, the more accurate the static maps $M_i$ are, the higher the quality of the trained NeRF.
To maximize the accuracy of $M_i$, we follow the British idiom ``horses for courses'', which suggests matching talents to tasks, and approach the problem through a novel framework called {\it Heuristics-Guided Segmentation (HuGS)} (Sec.~\ref{sec:heuristics-guided_segmentation}).
As Fig.~\ref{fig:method_pipeline} shows, HuGS combines the strengths of both hand-crafted heuristics in identifying coarse cues of static objects and the capabilities of state-of-the-art segmentation models in producing sharp and accurate object boundaries.
Furthermore, we conduct an in-depth analysis of the choice of heuristics (Sec.~\ref{sec:heuristics_development}). 
Our solution combines novel SfM-based heuristics, which effectively identify static objects with high-frequency texture patterns, with the color residual heuristics from a partially trained Nerfacto~\cite{tancik2023nerfstudio}, which excel at detecting static objects characterized by low-frequency textures. This tailored integration of heuristics allows our method to robustly capture the full range of static scene elements across diverse texture profiles.

\subsection{Heuristics-Guided Segmentation (HuGS)}
\label{sec:heuristics-guided_segmentation}

While humans can easily distinguish between transient and static objects, providing a rigorous mathematical definition of this distinction has proven elusive thus far due to the high diversity of real-world scenes.
To this end, the most effective existing solutions rely heavily on hand-crafted heuristics to make this distinction.
For example, NeRF-W~\cite{martin2021nerf} employs the heuristics that transient objects usually have lower density than their static counterparts and incorporates this as a regularization term during NeRF training;  RobustNeRF~\cite{sabour2023robustnerf} leverages the observation that transient objects are typically harder to fit during optimization and uses it to produce the static maps used in Eq.~\ref{eq:nerf_static_scenes}.
However, despite their success, these methods implicitly make the strong assumption that distinguishing between transient and static rays/pixels can be determined {\it solely} based on simple hand-crafted heuristics, which does not hold up while handling the diverse shapes and appearances of real-world objects.
As a result, these methods are prone to produce errors and/or ambiguous object boundaries (\cref{fig:sam_with_existing_method_nerfw_heuristic,fig:sam_with_existing_method_robust_heuristic}).

To address this limitation, we propose a novel framework HuGS that avoids fully relying on hand-crafted heuristics to differentiate between transient and static objects. 
Instead, our approach leverages heuristics to provide {\it only} rough hints $\mathcal{H}_{i}$ about potential static objects in each image $I_{i}$, and then refines those imprecise cues into accurate static maps $M_{i}$ using the segmentation masks of $I_{i}$ provided by model $S$. Specifically, let $S(I_i) = \{m_i^1, m_i^2, ..., m_i^{N_{M_i}}\}$, where $m_i^j$ denotes the segmentation mask of the $j$-th object (instance) and $N_{M_i}$ is the number of masks, we have:
\begin{align}
    M_{i} = \bigcup m_i^j, \ \forall \frac{m_i^j \cap \mathcal{H}_{i}}{m_i^j} \geq \mathcal{T}_{m} , 
    \label{eq:heuristic_guided_segmentation}
\end{align}
where $\mathcal{T}_{m}$ is a user-specified threshold and we implement $S$ using the state-of-the-art SAM~\cite{kirillov2023segment}.
As shown in \cref{fig:sam_with_existing_method}, our framework can produce static maps with sharp object boundaries even when using partially trained (10\%) models of previous methods~\cite{martin2021nerf,sabour2023robustnerf} as heuristics (\cref{fig:sam_with_existing_method_nerfw_heuristic,fig:sam_with_existing_method_robust_heuristic}). 
However, despite the relaxation, the success of our framework is based on the assumption that rough but accurate $\mathcal{H}_i$ about static objects are available (\cref{fig:sam_with_existing_method_nerfw_static,fig:sam_with_existing_method_robust_static}).

\subsection{Heuristics Development}
\label{sec:heuristics_development}

\begin{figure}
  \centering
  \begin{subfigure}{0.325\linewidth}
    \includegraphics[width=\linewidth]{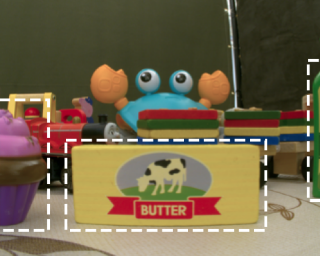}\\
    \includegraphics[width=\linewidth]{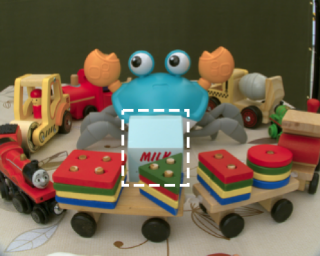}
    \caption{Training Image}
    \label{fig:robustnerf_with_diff_threshold_train_images}
  \end{subfigure}
  \hfill
  \begin{subfigure}{0.325\linewidth}
    \includegraphics[width=\linewidth]{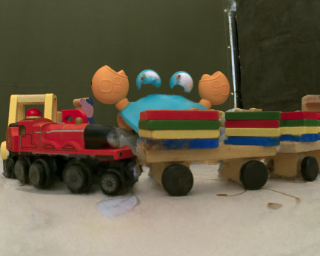}\\
    \includegraphics[width=\linewidth]{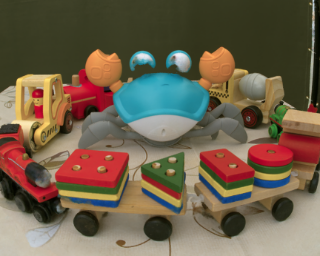}
    \caption{$\text{Quantile}=0.7$}
    \label{fig:robustnerf_with_diff_threshold_7}
  \end{subfigure}
  \hfill
  \begin{subfigure}{0.325\linewidth}
    \includegraphics[width=\linewidth]{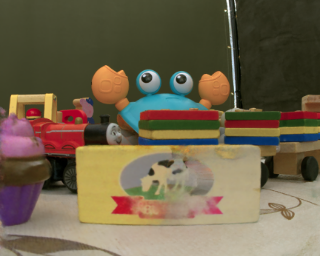}\\
    \includegraphics[width=\linewidth]{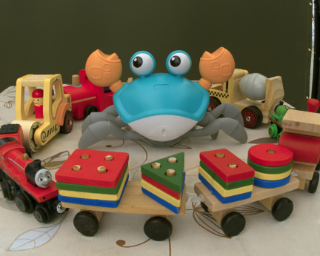}
    \caption{$\text{Quantile}=0.9$}
    \label{fig:robustnerf_with_diff_threshold_9}
  \end{subfigure}
  \caption{\textbf{Performance of RobustNeRF \vs transient objects of different sizes.} Transient distractors in the training images are framed in white.
  A lower quantile (threshold) causes the model to miss small-sized static objects, while a higher quantile prevents the removal of large-sized transient objects.\vspace{-6mm}}
  \label{fig:robustnerf_with_diff_threshold}
\end{figure}

To provide rough but accurate heuristics $\mathcal{H}_i$ of static objects, we use a combination of two complementary heuristics, \ie our novel SfM-based heuristics and the color residual heuristics from a partially trained Nerfacto~\cite{tancik2023nerfstudio}, which excel in detecting statics objects with high-frequency and low-frequency textures respectively.

\begin{figure}
  \centering
  \leftline{\scriptsize ~~~~~~(a) Training Image~~~~~~~~~~~(b) Static Map ($\mathcal{H}_i^{SfM}$)~~~(c) Static Map (Combined $\mathcal{H}_i$)}

  \includegraphics[width=\linewidth]{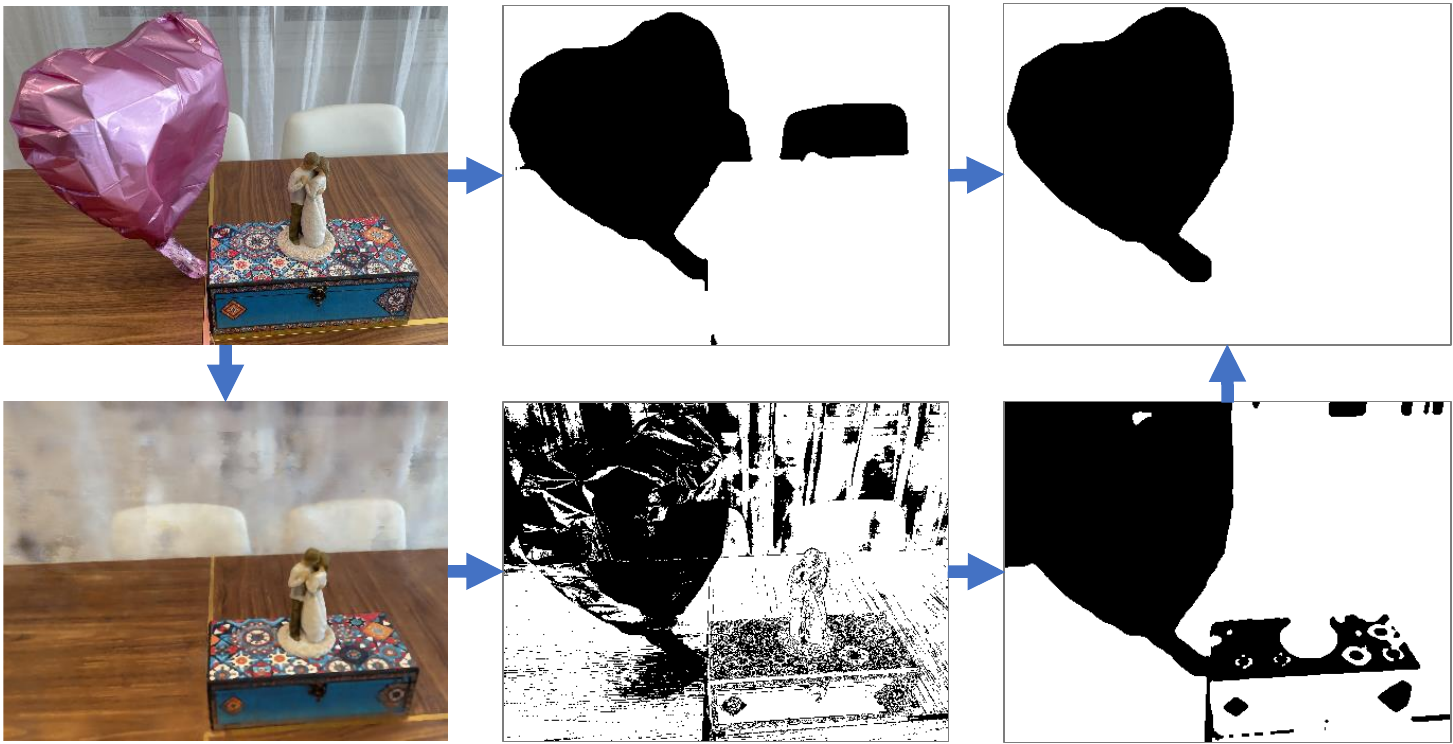}

  \vspace{-1mm}
  \leftline{\scriptsize ~~~~~~~~~(d) Rendering~~~~~~~~~~~~~~~~~~~~~~~~~~~~~(e) $\mathcal{H}_i^{CR}$~~~~~~~~~~~~~~~~~~~~~~(f) Static Map ($\mathcal{H}_i^{CR}$)}
  \caption{\textbf{Heuristics combination.} (b) The SfM-based heuristics $\mathcal{H}_i^{SfM}$ alone captures high-frequency static details (\eg, box textures) well but misses smooth ones (\eg, white chairs). This could be complemented by incorporating (e) residual-based heuristics $\mathcal{H}_i^{CR}$ from a (d) Nerfacto with $5k$ training iterations which does the opposite (f). 
  Their combination (c) covers the full spectrum of static scenes and identifies transient objects (\eg, pink balloon). }
  \vspace{-6mm}
  \label{fig:combine_example}
\end{figure}

\vspace{2mm}
\noindent \textbf{SfM-based Heuristics.} 
As mentioned above in Sec.~\ref{sec:preliminary}, SfM reconstruction relies on matching distinct, identifiable features across images. This makes it well-suited for detecting objects characterized by high-frequency textures, since these distinctive textures provide abundant unique features to match.
To distinguish between static and transient objects, our SfM-based heuristics share a similar high-level intuition to previous methods in that transient objects are considered a minority compared to static ones and their positions are constantly changing. However, ours has a different interpretation of ``minority". Specifically, our method defines it as the {\it frequency of occurrence} across input images, which aligns well with the temporal meaning of ``transient''. In contrast, NeRF-W~\cite{martin2021nerf} and RobustNeRF~\cite{sabour2023robustnerf} interpret ``minority'' in terms of total density or quantile of color residual respectively, which relate more to {\it spatial area coverage}. As a result, their methods struggle with transient objects of varying sizes (\cref{fig:robustnerf_with_diff_threshold}), since the area-based definitions do not fully capture the temporal aspect of identifying transient objects.
Recalling the definition of $n_i^j$ (the number of matching points) and $N_I$ (the number of input images) in Sec.~\ref{sec:preliminary}, we have the following observation:
\begin{observation}
\label{ob:feat}
The SfM features of static objects have much larger $n_i^j$ than those of transient ones in image $I_i$.
\end{observation}
\noindent Note that the already smaller $n_i^j$ of transient objects could be further reduced by their constantly changing positions (\ie, less likely to get matched during SfM reconstruction), making them easier to distinguish.
Accordingly, we set a threshold $\mathcal{T}_{SfM}$ to obtain set $\mathcal{X}_i$ of the coordinates of static feature points for each image $I_i$:
\begin{equation}
    \mathcal{X}_i = \left\{\mathbf{x}_{i}^j\left\vert\ \mathbf{x}_{i}^j\in\mathcal{F}_i~\text{and}~\frac{n_i^j}{N_I} \geq \mathcal{T}_{SfM}\right.\right\}.
\end{equation}
where we set $\mathcal{T}_{SfM}$ based on $\frac{n_i^j}{N_I}$ rather than $n_i^j$ as the former better represents the \textit{frequency of occurrence} across the whole scene.
However, $\mathcal{X}_i$ is a relatively sparse point set, so we need to convert it to a pixel-wise static map for NeRF training.
Fortunately, SAM~\cite{kirillov2023segment} is a promptable segmentation model that can accept points as prompts and output their corresponding segmentation masks.
Therefore, we feed $\mathcal{X}_i$ into SAM to obtain heuristics $\mathcal{H}_i^{SfM}$, which is a static map where a pixel value of 1 means that it belongs to a static object and 0 means that it is a transient one. 

\vspace{2mm}
\noindent \textbf{Combined Heuristics.}
Although effective, our SfM-based heuristics $\mathcal{H}^{SfM}$ may neglect low-frequency static objects due to their lack of distinctive features (Fig.~\ref{fig:combine_example}).
To address this limitation, we propose an integrated approach that incorporates the complementary strength of another heuristic: the color residual of a partially trained Nerfacto~\cite{tancik2023nerfstudio}, which effectively identifies smooth transient objects but struggles with textured objects.
Specifically, we first train a Nerfacto for several thousands of iterations and construct its color residual map $\mathcal{R}_i$ using the color residual for each ray $\mathbf{r}$ as $\epsilon(\mathbf{r}) = \lVert \hat{\mathbf{C}}(\mathbf{r}) - \mathbf{C}(\mathbf{r}) \rVert_2$.
We then combine $\mathcal{H}_i^{SfM}$ with residual-based heuristics $\mathcal{H}_i^{CR}$ to get heuristics $\hat{\mathcal{H}}_i$:
\begin{equation}
    \hat{\mathcal{H}}_i = \mathcal{H}_i^{SfM} \cup \mathcal{H}_i^{CR},
\label{eq:heuristics_full}
\end{equation}
where $\mathcal{H}_i^{CR} = \mathcal{R}_i \leq \text{mean}(\mathcal{R}_i)$. 
While in practice, our $\mathcal{H}_i^{SfM}$ may occasionally incorrectly include some transient objects due to misclassification of feature points or SAM segmentation errors. 
To eliminate them, we apply an upper bound defined by $\hat{\mathcal{H}}_i^{CR}$ to Eq.~\ref{eq:heuristics_full} as additional insurance:
\begin{equation}
    \mathcal{H}_i = \hat{\mathcal{H}}_i \cap \hat{\mathcal{H}}_i^{CR},
\label{eq:heuristics_full_bounded}
\end{equation}
where $\hat{\mathcal{H}}_i^{CR} = \mathcal{R}_i \leq \text{quantile}(\mathcal{R}_i, \mathcal{T}_{CR})$.
$\mathcal{T}_{CR}$ is a high threshold that ensures $\hat{\mathcal{H}}_i^{CR}$ include all static objects.

\vspace{2mm}
\noindent \textbf{Remark.} 
We use Nerfacto~\cite{tancik2023nerfstudio} to generate residual maps as it can be trained quickly with much fewer computational resources and still producing reasonable results (Fig.~\ref{fig:combine_example}). 
Although relatively low, this level of performance is sufficient to satisfy our requirement for rough heuristic cues, which further demonstrates the superiority of our paradigm.

\section{Experiments}
\label{sec:expriment}

\subsection{Experimental Setup}
\label{sec:expriment_setup}
\noindent \textbf{Datasets.} We use three datasets in our experiments: 

\begin{figure*}
  \centering
  \setlength{\tabcolsep}{3pt}
  \resizebox{\linewidth}{!}{
  \begin{tabular}{@{}l|ccc|ccc|ccc|ccc|ccc|ccc@{}}
      \toprule
      \multirow{2}*{Method} & \multicolumn{3}{c|}{Car} & \multicolumn{3}{c|}{Cars} & \multicolumn{3}{c|}{Bag} & \multicolumn{3}{c|}{Chairs} & \multicolumn{3}{c|}{Pillow} & \multicolumn{3}{c}{Avg.}\\
      & PSNR$\uparrow$ & SSIM$\uparrow$ & LPIPS$\downarrow$ & PSNR$\uparrow$ & SSIM$\uparrow$ & LPIPS$\downarrow$ & PSNR$\uparrow$ & SSIM$\uparrow$ & LPIPS$\downarrow$ & PSNR$\uparrow$ & SSIM$\uparrow$ & LPIPS$\downarrow$ & PSNR$\uparrow$ & SSIM$\uparrow$ & LPIPS$\downarrow$ & PSNR$\uparrow$ & SSIM$\uparrow$ & LPIPS$\downarrow$ \\
      \midrule
      Nerfacto~\cite{tancik2023nerfstudio} & 30.71 & .862 & .227 & 29.50 & .809 & .316 & 31.32 & .917 & .103 & 27.41 & .811 & .258 & 29.86 & .906 & .148 & 29.76 & .861 & .210 \\
      Mip-NeRF 360~\cite{barron2022mip2} & 26.67 & .846 & .238 & 28.88 & .822 & .281 & 32.81 & .948 & .053 & 27.07 & .839 & .179 & 29.66 & .919 & .123 & 29.02 & .875 & .175 \\
      \midrule
      NeRF-W~\cite{martin2021nerf} & 29.44 & .901 & .124 & 28.34 & .867 & .186 & 34.49 & .946 & .045 & 22.75 & .826 & .187 & 29.04 & .915 & .142 & 28.81 & .891 & .137 \\
      HA-NeRF~\cite{chen2022hallucinated} & 28.69 & .915 & .124 & 31.95 & .903 & .143 & 38.48 & .969 & \third{.021} & 33.48 & .922 & .071 & 31.66 & .946 & .083 & 32.85 & \third{.931} & .089 \\
      D$^2$NeRF~\cite{wu2022d} & 34.03 & .874 & .099 & 33.67 & .844 & .123 & 33.77 & .889 & .118 & 32.77 & .875 & .113 & 29.49 & .907 & .139 & 32.75 & .878 & .118 \\
      RobustNeRF~\cite{sabour2023robustnerf} & \third{37.31} & \second{.968} & \second{.040} & \second{40.52} & \second{.963} & \third{.047} & \third{40.50} & \third{.976} & .026 & \second{38.56} & \third{.958} & \third{.037} & \third{41.31} & \third{.980} & \third{.028} & \third{39.64} & \second{.969} & \third{.036} \\
      \midrule
      Ours (Nerfacto) & \second{39.49} & \third{.964} & \third{.042} & \third{39.95} & \third{.958} & \first{.045} & \second{41.39} & \second{.980} & \first{.017} & \third{38.48} & \second{.962} & \second{.036} & \second{42.70} & \second{.982} & \second{.025} & \second{40.40} & \second{.969} & \second{.033} \\
      Ours (Mip-NeRF 360) & \first{39.75} & \first{.972} & \first{.036} & \first{40.74} & \first{.966} & \second{.046} & \first{42.32} & \first{.983} & \second{.019} & \first{39.32} & \first{.968} & \first{.033} & \first{43.90} & \first{.986} & \first{.023} & \first{41.21} & \first{.975} & \first{.032} \\
      \bottomrule
  \end{tabular}}\\
  \vspace{1mm}
  \begin{subfigure}{0.012\linewidth}
  \rotatebox{90}{\scriptsize ~~~~~~~~~~~~~~~~~~~~~~~~~~~~~~~~~~~Car}
  \end{subfigure}
  \hfill
  \begin{subfigure}{0.16\linewidth}
    \includegraphics[width=\linewidth]{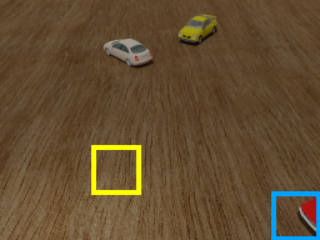}\\
    \includegraphics[width=0.48\linewidth]{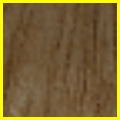}
    \hfill
    \includegraphics[width=0.48\linewidth]{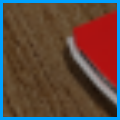}
    \caption*{Test Image}
  \end{subfigure}
  \hfill
  \begin{subfigure}{0.16\linewidth}
    \includegraphics[width=\linewidth]{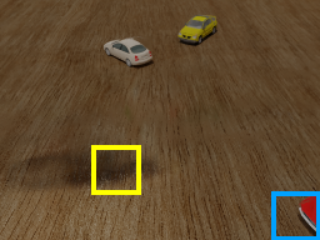}\\
    \includegraphics[width=0.48\linewidth]{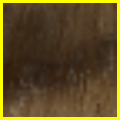}
    \hfill
    \includegraphics[width=0.48\linewidth]{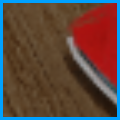}
    \caption*{NeRF-W}
  \end{subfigure}
  \hfill
  \begin{subfigure}{0.16\linewidth}
    \includegraphics[width=\linewidth]{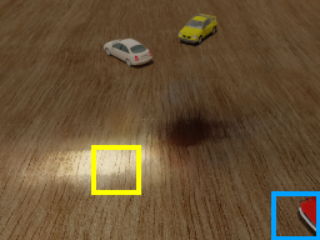}\\
    \includegraphics[width=0.48\linewidth]{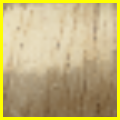}
    \hfill
    \includegraphics[width=0.48\linewidth]{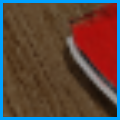}
    \caption*{HA-NeRF}
  \end{subfigure}
  \hfill
  \begin{subfigure}{0.16\linewidth}
    \includegraphics[width=\linewidth]{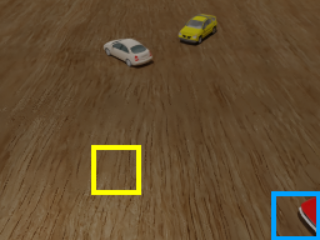}\\
    \includegraphics[width=0.48\linewidth]{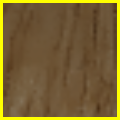}
    \hfill
    \includegraphics[width=0.48\linewidth]{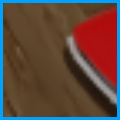}
    \caption*{D$^2$NeRF}
  \end{subfigure}
  \hfill
  \begin{subfigure}{0.16\linewidth}
    \includegraphics[width=\linewidth]{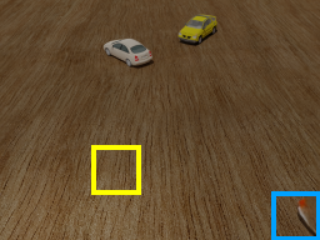}\\
    \includegraphics[width=0.48\linewidth]{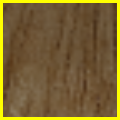}
    \hfill
    \includegraphics[width=0.48\linewidth]{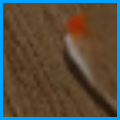}
    \caption*{RobustNeRF}
  \end{subfigure}
  \hfill
  \begin{subfigure}{0.16\linewidth}
    \includegraphics[width=\linewidth]{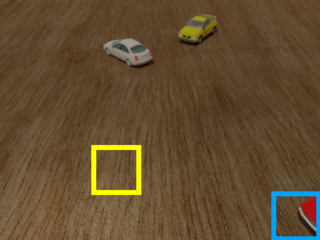}\\
    \includegraphics[width=0.48\linewidth]{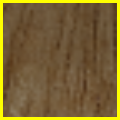}
    \hfill
    \includegraphics[width=0.48\linewidth]{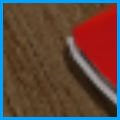}
    \caption*{Ours (Mip-NeRF 360)}
  \end{subfigure}
  \caption{\textbf{Quantitative and qualitative results on the Kubric dataset. } The \first{1st}, \second{2nd} and \third{3rd} best results are highlighted. Quantitatively, our method not only significantly improves the performance of Nerfacto and Mip-NeRF 360, but also helps Mip-NeRF 360 outperform the previous methods and become the SOTA. Qualitatively, our method can better preserve static details while ignoring transient distractors.}
  \label{fig:compare_kubric}
\end{figure*}
\begin{figure*}
  \centering
  \setlength{\tabcolsep}{3pt}
  \resizebox{\linewidth}{!}{
  \begin{tabular}{@{}l|ccc|ccc|ccc|ccc|ccc@{}}
    \toprule
    \multirow{2}*{Method} & \multicolumn{3}{c|}{Statue} & \multicolumn{3}{c|}{Android} & \multicolumn{3}{c|}{Crab} & \multicolumn{3}{c|}{BabyYoda} & \multicolumn{3}{c}{Avg.}\\
    & PSNR$\uparrow$ & SSIM$\uparrow$ & LPIPS$\downarrow$ & PSNR$\uparrow$ & SSIM$\uparrow$ & LPIPS$\downarrow$ & PSNR$\uparrow$ & SSIM$\uparrow$ & LPIPS$\downarrow$ & PSNR$\uparrow$ & SSIM$\uparrow$ & LPIPS$\downarrow$ & PSNR$\uparrow$ & SSIM$\uparrow$ & LPIPS$\downarrow$\\
    \midrule
    Nerfacto~\cite{tancik2023nerfstudio} & 18.21 & .591 & .336 & 21.34 & .617 & .226 & 26.62 & .864 & .135 & 22.06 & .697 & .267 & 22.06 & .692 & .241 \\
    Mip-NeRF 360~\cite{barron2022mip2} & \third{19.86} & .690 & .233 & 21.81 & .695 & .176 & 29.25 & .918 & .086 & 23.75 & .770 & .216 & 23.67 & .768 & .178 \\
    \midrule
    NeRF-W~\cite{martin2021nerf} & 18.91 & .616 & .369 & 20.62 & .664 & .258 & 26.91 & .866 & .157 & 28.64 & .752 & .260 & 23.77 & .725 & .261 \\
    HA-NeRF~\cite{chen2022hallucinated} & 18.67 & .616 & .367 & 22.03 & .706 & .203 & 28.58 & .901 & .116 & \third{29.28} & .779 & .208 & 24.64 & .750 & .224 \\
    RobustNeRF~\cite{sabour2023robustnerf} & \second{20.60} & \second{.758} & \second{.154} & \second{23.28} & \second{.755} & \third{.126} & \second{32.22} & \second{.945} & \third{.060} & \second{29.78} & \second{.821} & \third{.155} & \second{26.47} & \second{.820} & \third{.124} \\
    \midrule
    Ours (Nerfacto) & 19.18 & \third{.703} & \third{.183} & \third{22.59} & \third{.720} & \first{.120} & \third{32.11} & \third{.939} & \second{.033} & 28.77 & \third{.802} & \first{.087} & \third{25.66} & \third{.791} & \second{.106} \\
    Ours (Mip-NeRF 360) & \first{21.00} & \first{.774} & \first{.135} & \first{23.32} & \first{.763} & \second{.123} & \first{34.16} & \first{.956} & \first{.032} & \first{30.70} & \first{.834} & \second{.124} & \first{27.29} & \first{.832} & \first{.103} \\
    \bottomrule
  \end{tabular}}\\
  \vspace{1mm}
  \begin{subfigure}{0.012\linewidth}
  \rotatebox{90}{\scriptsize ~~~~~~~~~~~~~~~~~~~~~~~~~~~BabyYoda~~~~~~~~~~~~~~~~~~~~~~~~~~~~~~~~~~~~~~~~~Android}
  \end{subfigure}
  \hfill
  \begin{subfigure}{0.16\linewidth}
    \includegraphics[width=\linewidth]{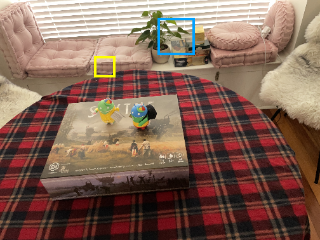}\\
    \includegraphics[width=0.48\linewidth]{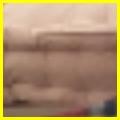}
    \hfill
    \includegraphics[width=0.48\linewidth]{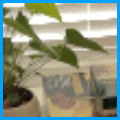}\\
    \includegraphics[width=\linewidth]{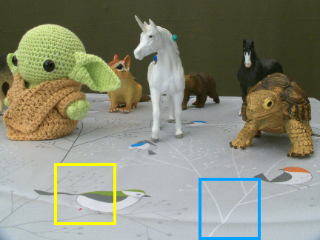}\\
    \includegraphics[width=0.48\linewidth]{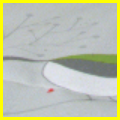}
    \hfill
    \includegraphics[width=0.48\linewidth]{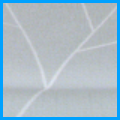}
    \caption*{Test Image}
  \end{subfigure}
  \hfill
  \begin{subfigure}{0.16\linewidth}
    \includegraphics[width=\linewidth]{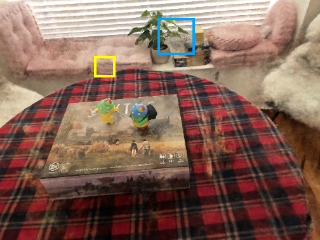}\\
    \includegraphics[width=0.48\linewidth]{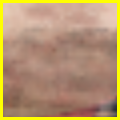}
    \hfill
    \includegraphics[width=0.48\linewidth]{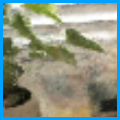}\\
    \includegraphics[width=\linewidth]{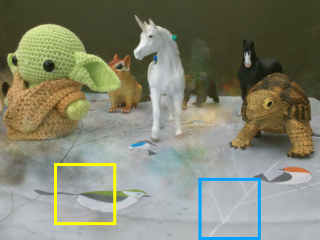}\\
    \includegraphics[width=0.48\linewidth]{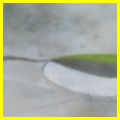}
    \hfill
    \includegraphics[width=0.48\linewidth]{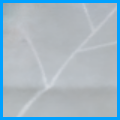}
    \caption*{Mip-NeRF 360}
  \end{subfigure}
  \hfill
  \begin{subfigure}{0.16\linewidth}
    \includegraphics[width=\linewidth]{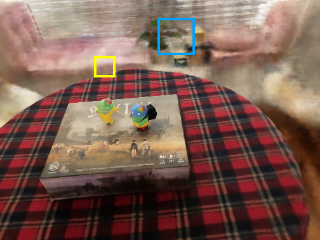}\\
    \includegraphics[width=0.48\linewidth]{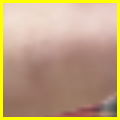}
    \hfill
    \includegraphics[width=0.48\linewidth]{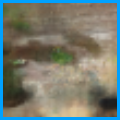}\\
    \includegraphics[width=\linewidth]{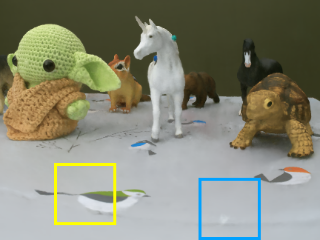}\\
    \includegraphics[width=0.48\linewidth]{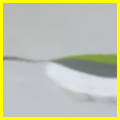}
    \hfill
    \includegraphics[width=0.48\linewidth]{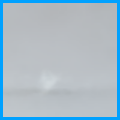}
    \caption*{NeRF-W}
  \end{subfigure}
  \hfill
  \begin{subfigure}{0.16\linewidth}
    \includegraphics[width=\linewidth]{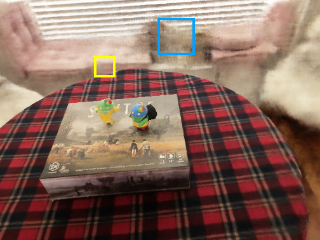}\\
    \includegraphics[width=0.48\linewidth]{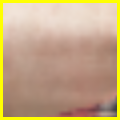}
    \hfill
    \includegraphics[width=0.48\linewidth]{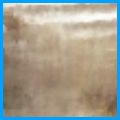}\\
    \includegraphics[width=\linewidth]{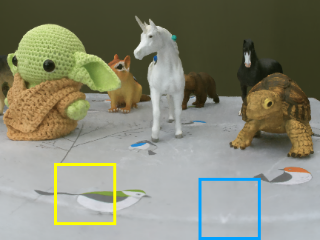}\\
    \includegraphics[width=0.48\linewidth]{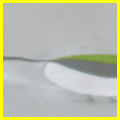}
    \hfill
    \includegraphics[width=0.48\linewidth]{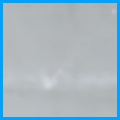}
    \caption*{HA-NeRF}
  \end{subfigure}
  \hfill
  \begin{subfigure}{0.16\linewidth}
    \includegraphics[width=\linewidth]{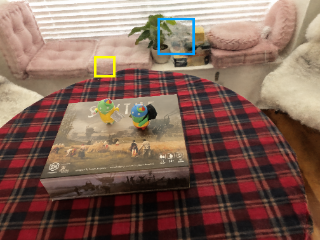}\\
    \includegraphics[width=0.48\linewidth]{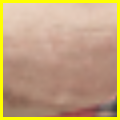}
    \hfill
    \includegraphics[width=0.48\linewidth]{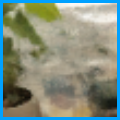}\\
    \includegraphics[width=\linewidth]{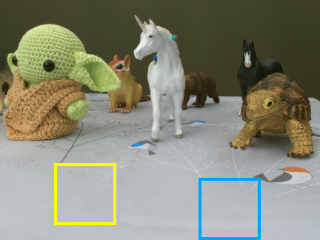}\\
    \includegraphics[width=0.48\linewidth]{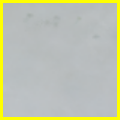}
    \hfill
    \includegraphics[width=0.48\linewidth]{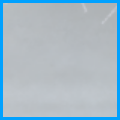}
    \caption*{RobustNeRF}
  \end{subfigure}
  \hfill
  \begin{subfigure}{0.16\linewidth}
    \includegraphics[width=\linewidth]{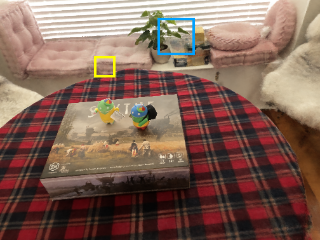}\\
    \includegraphics[width=0.48\linewidth]{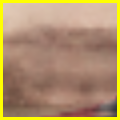}
    \hfill
    \includegraphics[width=0.48\linewidth]{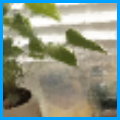}\\
    \includegraphics[width=\linewidth]{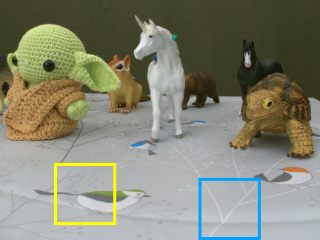}\\
    \includegraphics[width=0.48\linewidth]{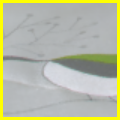}
    \hfill
    \includegraphics[width=0.48\linewidth]{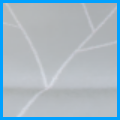}
    \caption*{Ours (Mip-NeRF 360)}
  \end{subfigure}
  \caption{\textbf{Quantitative and qualitative results on the Distractor dataset.} The \first{1st}, \second{2nd} and \third{3rd} best results are highlighted. Our method applied to Mip-NeRF 360 is the best in most quantitative results, with our method applied to Nerfacto leading the rest. Qualitatively, our method captures scene details better compared to other baselines, which suffer from missing or disturbed details. }
  \label{fig:compare_distractor}
\end{figure*}
\begin{figure*}
  \centering
  \setlength{\tabcolsep}{3pt}
  \resizebox{\linewidth}{!}{
  \begin{tabular}{@{}l|ccc|ccc|ccc|ccc|ccc@{}}
    \toprule
    \multirow{2}*{Method} & \multicolumn{3}{c|}{Brandenburg Gate} & \multicolumn{3}{c|}{Sacre Coeur} & \multicolumn{3}{c|}{Taj Mahal} & \multicolumn{3}{c|}{Trevi Fountain} & \multicolumn{3}{c}{Avg.}\\
    & PSNR$\uparrow$ & SSIM$\uparrow$ & LPIPS$\downarrow$ & PSNR$\uparrow$ & SSIM$\uparrow$ & LPIPS$\downarrow$ & PSNR$\uparrow$ & SSIM$\uparrow$ & LPIPS$\downarrow$ & PSNR$\uparrow$ & SSIM$\uparrow$ & LPIPS$\downarrow$ & PSNR$\uparrow$ & SSIM$\uparrow$ & LPIPS$\downarrow$\\
    \midrule

    Nerfacto~\cite{tancik2023nerfstudio} & 24.69 & .890 & .132 & 21.09 & .819 & .169 & 22.80 & .804 & .237 & 22.63 & .748 & .195 & 22.80 & .815 & .183 \\
    Mip-NeRF 360~\cite{barron2022mip2} & 25.59 & .904 & .121 & \third{21.46} & .818 & .175 & \third{24.33} & .828 & .202 & \third{23.25} & .767 & .189 & 23.66 & .829 & .172 \\
    \midrule
    NeRF-W~\cite{martin2021nerf} & 23.62 & .851 & .171 & 19.75 & .749 & .233 & 22.23 & .772 & .317 & 20.80 & .664 & .299 & 21.60 & .759 & .255 \\
    HA-NeRF~\cite{chen2022hallucinated} & 23.93 & .881 & .140 & 19.85 & .808 & .175 & 20.70 & .811 & .234 & 20.07 & .713 & .223 & 21.14 & .803 & .193 \\
    RobustNeRF~\cite{sabour2023robustnerf} & \third{25.79} & \second{.923} & \third{.094} & 20.94 & \third{.852} & \third{.137} & \second{24.64} & \first{.859} & \first{.173} & \first{23.58} & \second{.785} & \second{.170} & \third{23.73} & \second{.855} & \second{.144} \\
    \midrule
    Ours (Nerfacto) & \second{25.99} & \third{.919} & \second{.087} & \second{22.03} & \second{.856} & \second{.132} & 24.06 & \third{.836} & \third{.198} & 22.90 & \third{.770} & \third{.173} & \second{23.74} & \third{.845} & \third{.147} \\
    Ours (Mip-NeRF 360) & \first{27.17} & \first{.929} & \first{.083} & \first{22.23} & \first{.862} & \first{.124} & \first{24.92} & \second{.857} & \second{.176} & \second{23.41} & \first{.788} & \first{.165} & \first{24.43} & \first{.859} & \first{.137} \\
    \bottomrule
  \end{tabular}}\\
  \vspace{1mm}
  \begin{subfigure}{0.012\linewidth}
  \rotatebox{90}{\scriptsize ~~~~~~~~~~~~~~~~~~~~~~~Trevi Fountain}
  \end{subfigure}
  \hfill
  \begin{subfigure}{0.16\linewidth}
    \includegraphics[width=\linewidth]{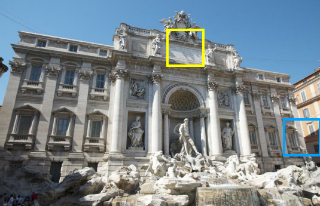}\\
    \includegraphics[width=0.48\linewidth]{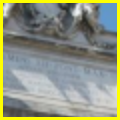}
    \hfill
    \includegraphics[width=0.48\linewidth]{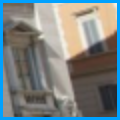}
    \caption*{Test Image}
  \end{subfigure}
  \hfill
  \begin{subfigure}{0.16\linewidth}
    \includegraphics[width=\linewidth]{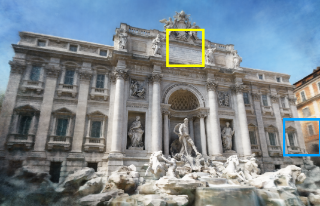}\\
    \includegraphics[width=0.48\linewidth]{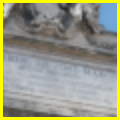}
    \hfill
    \includegraphics[width=0.48\linewidth]{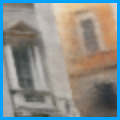}
    \caption*{Mip-NeRF 360}
  \end{subfigure}
  \hfill
  \begin{subfigure}{0.16\linewidth}
    \includegraphics[width=\linewidth]{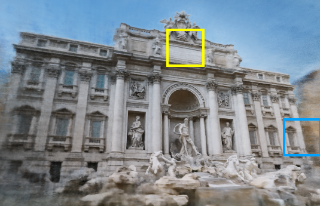}\\
    \includegraphics[width=0.48\linewidth]{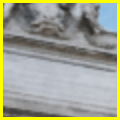}
    \hfill
    \includegraphics[width=0.48\linewidth]{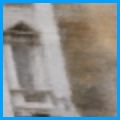}
    \caption*{NeRF-W}
  \end{subfigure}
  \hfill
  \begin{subfigure}{0.16\linewidth}
    \includegraphics[width=\linewidth]{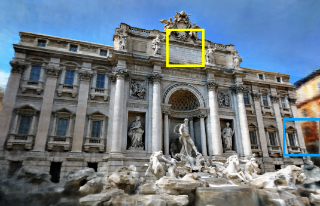}\\
    \includegraphics[width=0.48\linewidth]{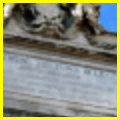}
    \hfill
    \includegraphics[width=0.48\linewidth]{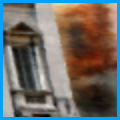}
    \caption*{HA-NeRF}
  \end{subfigure}
  \hfill
  \begin{subfigure}{0.16\linewidth}
    \includegraphics[width=\linewidth]{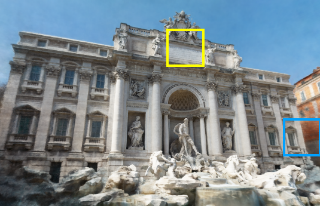}\\
    \includegraphics[width=0.48\linewidth]{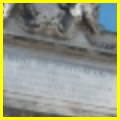}
    \hfill
    \includegraphics[width=0.48\linewidth]{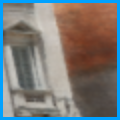}
    \caption*{RobustNeRF}
  \end{subfigure}
  \hfill
  \begin{subfigure}{0.16\linewidth}
    \includegraphics[width=\linewidth]{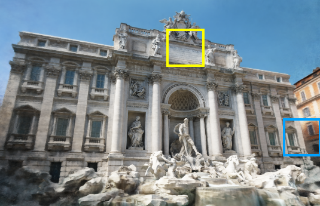}\\
    \includegraphics[width=0.48\linewidth]{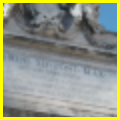}
    \hfill
    \includegraphics[width=0.48\linewidth]{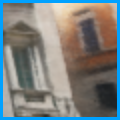}
    \caption*{Ours (Mip-NeRF 360)}
  \end{subfigure}
  \caption{\textbf{Quantitative and qualitative results on the Phototourism dataset.} The \first{1st}, \second{2nd} and \third{3rd} best results are highlighted.
  Note that the main content of test images in this dataset is the upper part of the building, which is less affected by transient distractors (\eg, tourists).
  Thus, our method brings less improvement but still yields competitive results against the SOTA.
  }
  \vspace{-6mm}
  \label{fig:compare_phototourism}
\end{figure*}

\begin{itemize}
    \item \textit{Kubric Dataset~\cite{wu2022d}.} Generated by Kubric~\cite{greff2022kubric}, this synthetic dataset contains five scenes with simple geometries placed in an empty room. The frames have temporal relationships and a subset of geometries serves as transient distractors that move between frames. 
    \item \textit{Distractor Dataset~\cite{sabour2023robustnerf}.}
    This real-world dataset has four controlled indoor scenes with 1-150 distractors per scene.
    \item \textit{Phototourism Dataset~\cite{martin2021nerf}.}
    This real-world dataset has scenes of four cultural landmarks, each with photos collected online containing various transient distractors.
    The landmark and distractor appearances vary across images due to shooting differences.
\end{itemize}

\vspace{2mm}
\noindent \textbf{Implementation Details.} Please see the supplementary materials for more details.
\begin{itemize}
\item \textit{HuGS.}
We use COLMAP~\cite{schonberger2016structure} for SfM reconstruction and SAM~\cite{kirillov2023segment} as the segmentation model.
COLMAP uses SIFT~\cite{lowe2004distinctive} to extract image features, and we set COLMAP's parameters to default values.
We set $\mathcal{T}_{m}$ to a common $0.5$.
The values of $\mathcal{T}_{SfM}$ and $\mathcal{T}_{CR}$ depend on the complexity of the scene, so we empirically set them to $0.2$ and $0.9$ for Kubric, $0.01$ and $0.95$ for Distractor, $0.01$ and $0.97$ for Phototourism datasets.

\item \textit{NeRF Training.}
We apply our method to two baseline NeRF models, Nerfacto~\cite{tancik2023nerfstudio} and Mip-NeRF 360~\cite{barron2022mip2}, to show its generalizability. 
We did not test it on the vanilla NeRF~\cite{mildenhall2020nerf} because the vanilla NeRF has difficulty handling the unbounded scenes in the Distractor dataset~\cite{barron2022mip2}.
\end{itemize}

\begin{figure*}
  \centering
  \setlength{\tabcolsep}{3pt}
  \resizebox{\linewidth}{!}{
  \begin{tabular}{@{}lcc|cc|cc|cc|cc|cc|cc@{}}
    \toprule
    \multirow{2}*{Method} & \multirow{2}*{Seg. Type} & \multirow{2}*{Require Prior} & \multicolumn{2}{c|}{Car} & \multicolumn{2}{c|}{Cars} & \multicolumn{2}{c|}{Bag} & \multicolumn{2}{c|}{Chairs} & \multicolumn{2}{c|}{Pillow} & \multicolumn{2}{c}{Avg.}\\
    & & & mIoU$\uparrow$ & F1$\uparrow$ & mIoU$\uparrow$ & F1$\uparrow$ & mIoU$\uparrow$ & F1$\uparrow$ & mIoU$\uparrow$ & F1$\uparrow$ & mIoU$\uparrow$ & F1$\uparrow$ & mIoU$\uparrow$ & F1$\uparrow$ \\
    \midrule
    DeepLabv3+~\cite{chen2018encoder} & \textit{Semantic} & \Checkmark & .604 & .378 & .578 & .293 & .501 & .048 & .564 & .239 & .535 & .149 & .556 & .221 \\
    Mask2Former~\cite{cheng2022masked} & \textit{Semantic} & \Checkmark & .664 & .520 & .622 & .376 & .513 & .071 & .707 & .561 & .653 & .377 & .632 & .381 \\
    Grounded-SAM~\cite{kirillov2023segment,liu2023grounding} & \textit{Open-Set} & \Checkmark & .888 & .877 & .640 & .406 & .755 & .671 & .603 & .373 & \third{.851} & \third{.828} & .747 & .631 \\
    DINO~\cite{caron2021emerging} & \textit{Video} & \Checkmark & \second{.947} & \second{.947} & .720 & .557 & .591 & .367 & \second{.777} & \second{.703} & \second{.911} & \second{.904} & .789 & .695 \\
    \midrule
    NeRF-W~\cite{martin2021nerf} & / & \XSolidBrush & .682 & .575 & .584 & .328 & .526 & .099 & .547 & .298 & .557 & .296 & .579 & .319 \\
    HA-NeRF~\cite{chen2022hallucinated} & / & \XSolidBrush & .869 & .852 & \third{.823} & \third{.724} & \second{.813} & \second{.771} & .729 & .595 & .819 & .766 & \third{.811} & \third{.742} \\
    D$^2$NeRF~\cite{wu2022d} & / & \XSolidBrush & \third{.912} & \third{.909} & \second{.895} & \second{.867} & \third{.794} & \third{.727} & .660 & .507 & .800 & .760 & \second{.812} & \second{.754} \\
    RobustNeRF~\cite{sabour2023robustnerf} & / & \XSolidBrush & .813 & .784 & .718 & .547 & .731 & .633 & \third{.731} & \third{.638} & .724 & .633 & .743 & .647 \\
    \midrule
    HuGS (Ours) & / & \XSolidBrush & \first{.963} & \first{.964} & \first{.940} & \first{.907} & \first{.939} & \first{.935} & \first{.937} & \first{.927} & \first{.940} & \first{.937} & \first{.944} & \first{.934} \\
    \bottomrule
  \end{tabular}}\\
  \vspace{1mm}
  \begin{subfigure}{0.012\linewidth}
  \rotatebox{90}{\scriptsize ~~~~~~~~~~~~~~~~Cars}
  \end{subfigure}
  \hfill
  \begin{subfigure}{0.118\linewidth}
    \includegraphics[width=\linewidth]{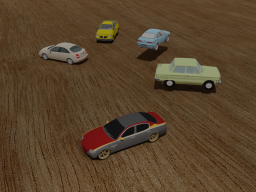}
    \caption*{Training Image}
  \end{subfigure}
  \hfill
  \begin{subfigure}{0.118\linewidth}
    \includegraphics[width=\linewidth]{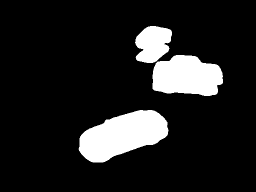}
    \caption*{GT}
  \end{subfigure}
  \hfill
  \begin{subfigure}{0.118\linewidth}
    \includegraphics[width=\linewidth]{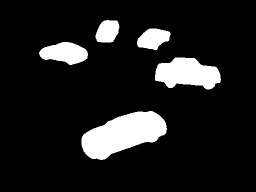}
    \caption*{Mask2former}
  \end{subfigure}
  \hfill
  \begin{subfigure}{0.118\linewidth}
    \includegraphics[width=\linewidth]{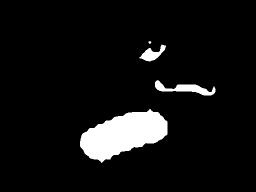}
    \caption*{DINO}
  \end{subfigure}
  \hfill
  \begin{subfigure}{0.118\linewidth}
    \includegraphics[width=\linewidth]{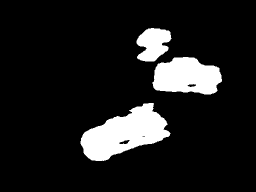}
    \caption*{HA-NeRF}
  \end{subfigure}
  \hfill
  \begin{subfigure}{0.118\linewidth}
    \includegraphics[width=\linewidth]{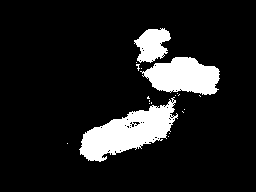}
    \caption*{D$^2$NeRF}
  \end{subfigure}
  \hfill
  \hfill
  \begin{subfigure}{0.118\linewidth}
    \includegraphics[width=\linewidth]{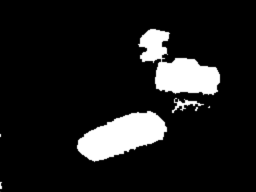}
    \caption*{RobustNeRF}
  \end{subfigure}
  \hfill
  \begin{subfigure}{0.118\linewidth}
    \includegraphics[width=\linewidth]{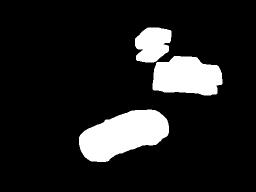}
    \caption*{Ours}
  \end{subfigure}
  \caption{\textbf{Quantitative and qualitative segmentation results on the Kubric dataset. } The \first{1st}, \second{2nd} and \third{3rd} best results are highlighted. }
  \vspace{-4mm}
  \label{fig:compare_seg}
\end{figure*}

\subsection{Evaluation on View Synthesis}
\label{sec:evaluation_view_synthesis}
\noindent \textbf{Baselines.}
In addition to baseline models, we compare our method to three other state-of-the-art heuristics-based methods: NeRF-W~\cite{martin2021nerf}, HA-NeRF~\cite{chen2022hallucinated} and RobustNeRF~\cite{sabour2023robustnerf}, which design heuristics based on scene density, pixel visible possibility, and color residuals, respectively.
We also compare our method to D$^2$NeRF~\cite{wu2022d} on the Kubric dataset, which is a dynamic NeRF that works well on monocular videos.
Segmentation-based methods are not included in our comparison because they rely on priors of transient distractors, which cannot be satisfied in most scenes.

\vspace{1mm}
\noindent \textbf{Comparisons.}
Both the above models and ours are trained on images disturbed by transient distractors and evaluated on images with only static scenes. 
We report image synthesis qualities based on PSNR, SSIM~\cite{wang2004image} and LPIPS~\cite{zhang2018unreasonable}.
\begin{itemize}
    \item {\it Kubric dataset} (\cref{fig:compare_kubric}).
    Compared to the native ones, applying our method leads to substantial PSNR improvements of $8.78$ to $12.84$dB for Nerfacto, and $9.51$ to $14.24$dB for Mip-NeRF 360.
    Our method achieves this by generating high-quality static maps that effectively shield the native models from pixels disturbed by transient distractors.
    Compared to the other baselines, our method achieves the highest quantitative results and maintains a good balance between ignoring transient distractors and preserving static details.
    Specifically, NeRF-W and HA-NeRF fail due to incorrect decoupling of transient distractors from the static scenes; D$^2$NeRF and RobustNeRF achieve better decoupling but lose static details such as ground textures and the red car.
    \item {\it Distractor dataset} (\cref{fig:compare_distractor}). The results and conclusions are similar to those on the Kubric dataset.
    \item {\it Phototourism dataset} (\cref{fig:compare_phototourism}).
    Its training and test sets share a unique feature: the landmark \textit{main bodies} are not deeply disturbed by transient distractors (\eg nearby tourists) and can be well reconstructed even without the removal of transient distractors. Thus, the improvement from our method mainly focus on the landmark \textit{boundaries} and is relatively less compared to the above datasets.
    Nonetheless, our results remain quantitatively competitive and qualitatively recover more details compared to prior works.
    Please see the supplement for more details.
\end{itemize}

\subsection{Evaluation on Segmentation}
\label{sec:evaluation_segmentation}
\noindent \textbf{Baselines.}
We perform the comparison on the Kubric dataset, which is synthetic and has ground truth segmentation data.
We compare our method with various existing segmentation models, including semantic segmentation models~\cite{chen2018encoder,cheng2022masked}, open-set segmentation models~\cite{kirillov2023segment,liu2023grounding} and video segmentation models~\cite{caron2021emerging}. 
The baseline NeRF models mentioned above are also compared by using the static maps generated after they are fully trained.

\vspace{1mm}
\noindent \textbf{Comparisons (\cref{fig:compare_seg}).}
We report segmentation qualities based on the mIoU and F1 score. 
Interestingly, we observe that even when prior knowledge is provided, the performance of existing segmentation models is limited because they are not designed for this specific task.
On the other hand, heuristics-based methods can roughly localize transient distractors but cannot provide accurate segmentation results.
By combining heuristics and the segmentation model together, our method takes the best of both worlds and can accurately segment transient distractors from static scenes without any prior knowledge.

\vspace{1mm}
\noindent \textbf{Verification of \cref{ob:feat}.} 
Please see the supplementary materials for additional experiments in which we verify the correctness of \cref{ob:feat}. 


\begin{figure}
  \centering
  \setlength{\tabcolsep}{3pt}
  \resizebox{\linewidth}{!}{
  \begin{tabular}{@{}lccc|ccc|ccc@{}}
    \toprule
    & \multirow{2}*{$\mathcal{H}^{SfM}$} & \multirow{2}*{$\mathcal{H}^{CR}$} & \multirow{2}*{SAM} & \multicolumn{3}{c|}{Kubric (Avg.)} & \multicolumn{3}{c}{Distractor (Avg.)}\\
    & & & & PSNR$\uparrow$ & SSIM$\uparrow$ & LPIPS$\downarrow$ & PSNR$\uparrow$ & SSIM$\uparrow$ & LPIPS$\downarrow$ \\
    \midrule
    (a) & -          & -          & -          & 29.76 & .861 & .210 & 22.06 & .692 & .241 \\
    (b) & \Checkmark & -          & -          & 38.14 & .962 & .054 & \first{25.67} & \second{.788} & \second{.108} \\
    (c) & -          & \Checkmark & -          & 39.86 & \third{.967} & \third{.036} & 24.95 & .767 & .146 \\
    (d) & -          & \Checkmark & \Checkmark & \second{40.12} & \second{.968} & \first{.033} & 24.58 & .779 & .126 \\
    (e) & \Checkmark & \Checkmark & -          & \third{40.11} & \second{.968} & \second{.035} & \third{25.40} & \third{.786} & \third{.116} \\
    (f) & \Checkmark & \Checkmark & \Checkmark & \first{40.40} & \first{.969} & \first{.033} & \second{25.66} & \first{.791} & \first{.106} \\
    \bottomrule
  \end{tabular}}\\
  \vspace{1mm}
  
  \begin{subfigure}{0.158\linewidth}
    \includegraphics[width=\linewidth]{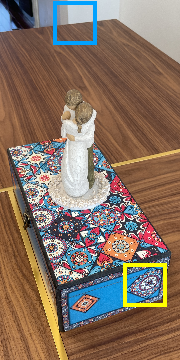}
    \caption*{Test Image}
  \end{subfigure}
  \hfill
  \begin{subfigure}{0.158\linewidth}
    \includegraphics[width=\linewidth]{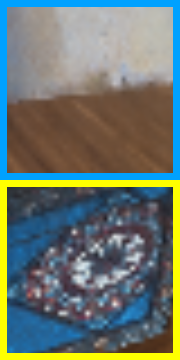}
    \caption*{(b)}
  \end{subfigure}
  \hfill
  \begin{subfigure}{0.158\linewidth}
    \includegraphics[width=\linewidth]{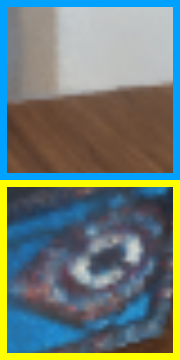}
    \caption*{(c)}
  \end{subfigure}
  \hfill
  \begin{subfigure}{0.158\linewidth}
    \includegraphics[width=\linewidth]{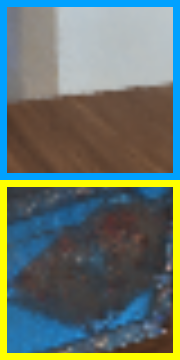}
    \caption*{(d)}
  \end{subfigure}
  \hfill
  \begin{subfigure}{0.158\linewidth}
    \includegraphics[width=\linewidth]{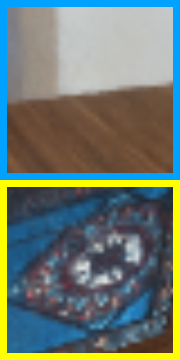}
    \caption*{(e)}
  \end{subfigure}
  \hfill
  \begin{subfigure}{0.158\linewidth}
    \includegraphics[width=\linewidth]{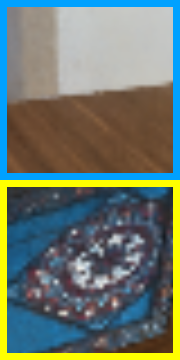}
    \caption*{(f)}
  \end{subfigure}
  \caption{\textbf{Ablation results.} 
  The patches in blue frames denote the smooth wall, and those in yellow frames denote complex textures. The \first{1st}, \second{2nd} and \third{3rd} best results are highlighted. 
  \vspace{-6mm}
  }
  \label{fig:ablation_results}
\end{figure}

\subsection{Ablation Study}
\label{sec:ablation_study}
Based on Nerfacto, we remove different components of our method to study their effects on two different datasets.
As shown in \cref{fig:ablation_results}, method (a) without any static maps, \ie, the native Nerfacto, performs the worst.
Using SfM-based heuristics or residual-based heuristics alone has limited improvement because the former cannot capture smooth surfaces and the latter has difficulty handling high-frequency details.
The complete method (f), which combines them with the segmentation model, achieves the best results.
\section{Conclusions}

In this work, we propose a novel heuristics-guided segmentation paradigm that effectively addresses the prevalent issue of transient distractors in real-world NeRF training. 
By strategically combining the complementary strengths of hand-crafted heuristics and state-of-the-art semantic segmentation models, our method achieves highly accurate segmentation of transient distractors across diverse scenes without any prior knowledge. Through meticulous heuristic design, our method can capture both high and low-frequency static scene elements robustly.
Extensive experiments demonstrate the superiority of our approach over existing methods. 
Please see the supplementary details for {\bf limitations and future work}.

{
    \small
    \bibliographystyle{ieeenat_fullname}
    \bibliography{main}
}

\clearpage
\setcounter{page}{1}
\maketitlesupplementary


\begin{figure}
  \centering
  \setlength{\tabcolsep}{4pt}
    \resizebox{\linewidth}{!}{
    \begin{tabular}{@{}lc|ccc|ccc@{}}
    \toprule
    \multirow{2}*{Method} & \multirow{2}*{Cor.} & \multicolumn{3}{c|}{Cars} & \multicolumn{3}{c}{Chairs}\\
    & & PSNR$\uparrow$ & SSIM$\uparrow$ & LPIPS$\downarrow$ & PSNR$\uparrow$ & SSIM$\uparrow$ & LPIPS$\downarrow$ \\
    \midrule
    \multirow{2}*{D$^2$NeRF} & - & 26.01 & .816 & .140 & 24.55 & .536 & .171 \\
    & \Checkmark & \textbf{33.67} & \textbf{.844} & \textbf{.123} & \textbf{32.77} & \textbf{.875} & \textbf{.113} \\
    \midrule
    \multirow{2}*{RobustNeRF} & - & 26.60 & .946 & .064 & 24.84 & .517 & .117 \\
    & \Checkmark & \textbf{40.52} & \textbf{.963} & \textbf{.047} & \textbf{38.56} & \textbf{.958} & \textbf{.037} \\
    \midrule
    \multirow{2}*{Ours (Mip-NeRF 360)} & - & 26.54 & .948 & .064 & 24.84 & .518 & .116 \\
    & \Checkmark & \textbf{40.74} & \textbf{.966} & \textbf{.046} & \textbf{39.32} & \textbf{.968} & \textbf{.033} \\
    \bottomrule
  \end{tabular}}\\
  \vspace{1mm}
   \rotatebox{90}{\scriptsize ~~~~~~~~~~~~~~~~~~~~Cars}
  \begin{subfigure}{0.315\linewidth}
    \includegraphics[width=\linewidth]{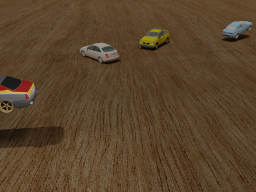}
    \caption*{Training Image}
  \end{subfigure}
  \hfill
  \begin{subfigure}{0.315\linewidth}
    \includegraphics[width=\linewidth]{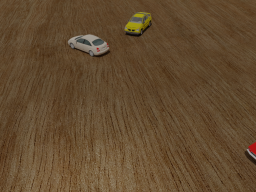}
    \caption*{Test Image (w/o Cor.)}
  \end{subfigure}
  \hfill
  \begin{subfigure}{0.315\linewidth}
    \includegraphics[width=\linewidth]{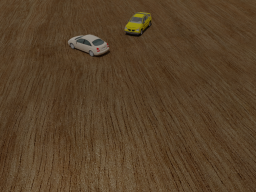}
    \caption*{Test Image (w/ Cor.)}
  \end{subfigure}\\
  \rotatebox{90}{\scriptsize ~~~~~~~~~~~~~~~~~~Chairs}
  \begin{subfigure}{0.315\linewidth}
    \includegraphics[width=\linewidth]{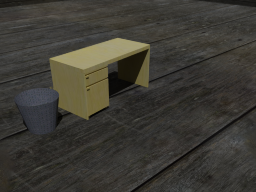}
    \caption*{Rendered Image}
  \end{subfigure}
  \hfill
  \begin{subfigure}{0.315\linewidth}
    \includegraphics[width=\linewidth]{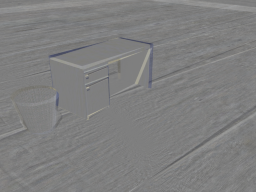}
    \caption*{Difference (w/o Cor.)}
  \end{subfigure}
  \hfill
  \begin{subfigure}{0.315\linewidth}
    \includegraphics[width=\linewidth]{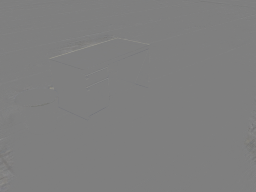}
    \caption*{Difference (w/ Cor.)}
  \end{subfigure}
  \caption{\textbf{Quantitative results and visualization of the correction on the Kubric dataset. } \textbf{Better results} are highlighted. After correction, the resynthesized test images keep the same illumination as the training images in the Cars scene and are aligned with the camera poses in the Chairs scene, resulting in smaller difference between the test images and the rendered images from NeRF. ``Cor.'': Correction.
  \vspace{-4mm}
  }
  \label{fig:kubric_correction}
\end{figure}
  
\section{Implementation Details}
\label{sec:supplement_implement_details}

\subsection{Datasets}

\vspace{1mm}
\noindent \textbf{Kubric Dataset~\cite{chen2022hallucinated}.}
Each scene of this dataset contains 200 noisy images for training
and 100 clean images for evaluation, and we downsample these images by 2x following the prior works.
As mentioned in \cref{sec:expriment_setup} of the main paper, images of this dataset are ordered in time, which is a requirement for dynamic NeRFs and video segmentation models to work.
In addition, we uncovered some inconsistencies in the test set of the released Kubric dataset~\cite{wu2022d} that affect the evaluation.
Specifically, i) in the Cars scene, the global illumination of the test set is inconsistent with that of the training set, and a red car that acts as a transient distractor is not removed from the test images; 
ii) in the Chairs scene, there is no alignment between the camera poses and corresponding images of the test set.
Therefore, we use the originally released script files to resynthesize the above scenes and correct the test images.
After correction, the quantitative results of different models have improved and are now consistent in magnitude with the results of other scenes (\cref{fig:kubric_correction}).

\vspace{1mm}
\noindent \textbf{Distractor Dataset~\cite{sabour2023robustnerf}.}
Each scene of this dataset contains 72-255 noisy images for training and 19-202 clean images for evaluation.
Images of this dataset are unordered, and we downsample them by 8x following \cite{sabour2023robustnerf}. 

\vspace{1mm}
\noindent \textbf{Phototourism Dataset~\cite{martin2021nerf}.}
Each scene of this dataset contains 859-1716 unordered noisy images for training and 10-27 nearly clean images for evaluation, and we downsample these images by 2x.
Different from other datasets, images of this dataset vary greatly in appearance (\eg, scene illumination, image style, sky and building color) due to diverse shooting environments and shooting equipment.
To address such appearance variations, we follow the prior works~\cite{martin2021nerf,barron2022mip2} and assign a 48-dimensional {\it appearance embedding} (a.k.a., {\it GLO vector}) to each image when training baseline models and ours.
Except for HA-NeRF, other models for comparison require finetuning appearance embeddings on new images. 
We therefore optimize the embeddings with $5k$ iterations using the left half of the test images after training and report metrics on the right half.

\subsection{Segment Anything Model (SAM)} 

The Segment Anything Model (SAM)~\cite{kirillov2023segment} is a promptable segmentation model that unifies the 2D segmentation task by introducing a prompt-based segmentation paradigm.
Given an image $I$ and a set of prompts $\mathcal{P}$ as input, SAM can output the corresponding 2D segmentation binary mask $M$:
\begin{align}
    M=\mathop{\text{SAM}}\left(I, \mathcal{P}\right),
\end{align}
where $\mathcal{P}$ can be points, boxes or masks.
Point and box prompts are the most common forms, while mask prompts are used to assist them.

In our method, we use SAM twice for each image:
\begin{enumerate}
    \item The first is to convert static SfM feature points (as input prompt $\mathcal{P}$) into their corresponding static map $\mathcal{H}_i^{SfM}$ (\cref{sec:heuristics_development} of the main paper).
    Specifically, SAM converts each of them (or with its nearest static neighbors) to a sub-map and unifies them into $\mathcal{H}_i^{SfM}$.
    \item The second is to generate instance segmentation masks $S(I_i)$ (\cref{sec:heuristics-guided_segmentation} of the main paper). SAM achieves this by taking a regular point grid as input prompt $\mathcal{P}$, and the default grid resolution of our method is $64\times 64$.
\end{enumerate}

\begin{figure*}
  \centering
  \setlength{\tabcolsep}{3pt}
  \resizebox{\linewidth}{!}{
  \begin{tabular}{@{}c|ccc|ccc|ccc|ccc|ccc@{}}
    \toprule
    \multirow{2}*{Quantile (Threshold)} & \multicolumn{3}{c|}{Statue} & \multicolumn{3}{c|}{Android} & \multicolumn{3}{c|}{Crab} & \multicolumn{3}{c|}{BabyYoda} & \multicolumn{3}{c}{Avg.}\\
    & PSNR$\uparrow$ & SSIM$\uparrow$ & LPIPS$\downarrow$ & PSNR$\uparrow$ & SSIM$\uparrow$ & LPIPS$\downarrow$ & PSNR$\uparrow$ & SSIM$\uparrow$ & LPIPS$\downarrow$ & PSNR$\uparrow$ & SSIM$\uparrow$ & LPIPS$\downarrow$ & PSNR$\uparrow$ & SSIM$\uparrow$ & LPIPS$\downarrow$\\
    \midrule
    $0.5$ & 16.80 & .670 & .303 & 19.46 & .695 & .212 & 16.48 & .751 & .329 & 17.59 & .630 & .395 & 17.58 & .687 & .310 \\
    $0.6$ & 18.38 & .707 & .243 & 20.52 & .722 & .187 & 21.09 & .857 & .177 & 20.54 & .736 & .259 & 20.13 & .755 & .216 \\
    $0.7$ & 20.06 & .750 & .175 & 21.45 & .738 & .156 & 26.17 & .922 & .092 & 25.58 & .798 & .186 & 23.32 & .802 & .152 \\
    $0.8$ & \textbf{20.60} & \textbf{.758} & \textbf{.154} & 23.28 & \textbf{.755} & \textbf{.126} & \textbf{32.22} & \textbf{.945} & \textbf{.060} & 29.78 & .821 & .155 & 26.47 & \textbf{.820} & \textbf{.124} \\
    $0.9$ & 20.54 & .745 & .171 & \textbf{23.31} & .753 & .128 & 32.16 & .944 & \textbf{.060} & \textbf{30.35} & \textbf{.828} & \textbf{.145} & \textbf{26.59} & .817 & .126 \\
    $1.0$ & 19.86 & .690 & .233 & 21.81 & .695 & .176 & 29.25 & .918 & .086 & 23.75 & .770 & .216 & 23.67 & .768 & .178 \\
    \bottomrule
  \end{tabular}}\\
  \vspace{1mm}
  \rotatebox{90}{\scriptsize ~~~~~~~~~~~~~~~~~~Statue}
  \begin{subfigure}{0.136\linewidth}
    \includegraphics[width=\linewidth]{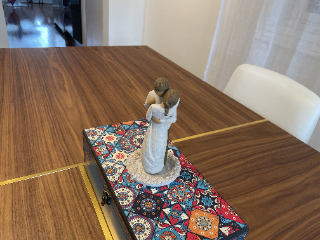}
    \caption*{Test Image}
  \end{subfigure}
  \hfill
  \begin{subfigure}{0.136\linewidth}
    \includegraphics[width=\linewidth]{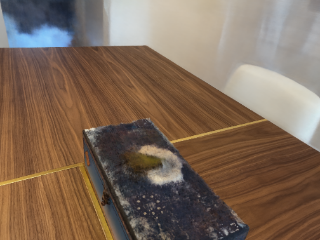}
    \caption*{$\text{Quantile}=0.5$}
  \end{subfigure}
  \hfill
  \begin{subfigure}{0.136\linewidth}
    \includegraphics[width=\linewidth]{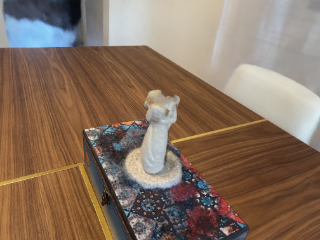}
    \caption*{$\text{Quantile}=0.6$}
  \end{subfigure}
  \hfill
  \begin{subfigure}{0.136\linewidth}
    \includegraphics[width=\linewidth]{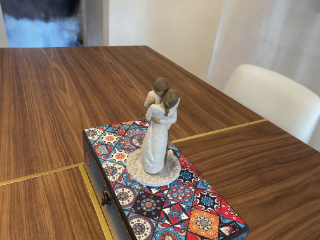}
    \caption*{$\text{Quantile}=0.7$}
  \end{subfigure}
  \hfill
  \begin{subfigure}{0.136\linewidth}
    \includegraphics[width=\linewidth]{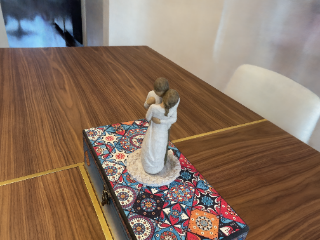}
    \caption*{$\text{Quantile}=0.8$}
  \end{subfigure}
  \hfill
  \begin{subfigure}{0.136\linewidth}
    \includegraphics[width=\linewidth]{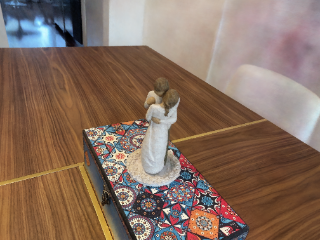}
    \caption*{$\text{Quantile}=0.9$}
  \end{subfigure}
  \hfill
  \begin{subfigure}{0.136\linewidth}
    \includegraphics[width=\linewidth]{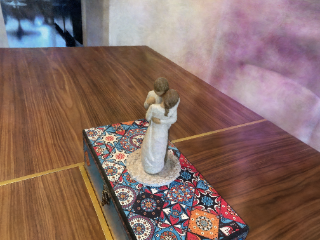}
    \caption*{$\text{Quantile}=1.0$}
  \end{subfigure}
  \caption{\textbf{Quantitative and qualitative results of RobustNeRF with different thresholds on the Distractor dataset.} The \textbf{best results} are highlighted. 
  \vspace{-4mm}
  }
  \label{fig:compare_distractor_robustnerf}
\end{figure*}

\subsection{Model Architecture, Training and Evaluation}

\vspace{1mm}
\noindent \textbf{Nerfacto~\cite{tancik2023nerfstudio}.} 
Nerfacto-related models in our experiments have the same architecture as Nerfacto-huge from the official Nerfstudio codebase.
The model consists of two density fields for proposal sampling and one nerfacto field for final sampling, and each field consists of a hash encoding~\cite{muller2022instant} with a base resolution of 16 and a tiny MLP with a width of 256.
The maximum hash encoding resolution is 512/2048 for the first/second density field, and 8192 for the nerfacto field.
We sample 512/256/128 points per ray in the first/second/third sampling.
We train these models for $25k$ iterations ($\sim$1 hour on one NVIDIA RTX 4090) through the Adam optimizer with a batch size of $16,384$ rays.
The learning rate is exponentially decayed from $10^{-2}$ to $10^{-3}$.
When experimenting on the Kubric dataset, we remove the first density field because we find that multiple proposal sampling will result in biased geometry and degraded rendering quality.
For the partially trained models used in HuGS, We reduce the number of training iterations to at most $5k$. 

\vspace{1mm}
\noindent \textbf{Mip-NeRF 360~\cite{barron2022mip2}.} 
For all models related to Mip-NeRF 360, we use the reference implementation of the MultiNeRF codebase~\cite{multinerf2022}.
The model architecture comprises a proposal MLP with 4 hidden layers and 256 units, and a NeRF MLP with 8 hidden layers and 1,024 units.
Following the default settings, we train Mip-NeRF 360 for $250k$ iterations ($\sim$1 day on four NVIDIA RTX 3090) through the Adam optimizer with a batch size of 16,384 rays.
The learning rate is exponentially decayed from $2\times 10^{-3}$ to $2\times 10^{-5}$.

\vspace{1mm}
\noindent \textbf{NeRF-W~\cite{martin2021nerf}.}
Because no official NeRF-W code is available, we refer to another popular open-source work~\cite{queianchen_nerf}.
The model architecture and training settings are the same as \cite{queianchen_nerf}, except that the number of training iterations is unified to $250k$ (\vs $200k$-$900k$ in \cite{queianchen_nerf}) and the batch size is expanded to 4,096 (\vs 1,024 in \cite{queianchen_nerf}).
To obtain heuristics or binary masks related to transient distractors,
we follow \cite{wu2022d} and apply a threshold of 0.1 on the accumulated weights of the transient component.

\vspace{1mm}
\noindent \textbf{HA-NeRF~\cite{chen2022hallucinated}.}
The model architecture and training settings are the same as the official codebase.
When experimenting on the Kubric and Distractor datasets, we remove the appearance hallucination module and unify the number of training iterations to $250k$, because the shooting environment of these datasets is consistent and the image style does not need to be transferred. 
The way to obtain heuristics or binary masks is the same as that of NeRF-W.

\vspace{1mm}
\noindent \textbf{D$^{2}$NeRF~\cite{wu2022d}.} 
The model architecture, training settings and the way to obtain heuristics or binary masks are the same as the official codebase.

\vspace{1mm}
\noindent \textbf{RobustNeRF~\cite{sabour2023robustnerf}.} 
Since the authors only provide the loss function as the official code and claim that RobustNeRF is implemented based on Mip-NeRF 360, we apply the loss function to MultiNeRF codebase~\cite{multinerf2022}.
RobustNeRF shares the same model architecture and training settings as Mip-NeRF 360.
During the experiment, we find that the color residual threshold ($\text{Quantile}=0.5$) provided in \cite{sabour2023robustnerf} is too low to reconstruct most static objects, so we experiment with different thresholds on the Distractor dataset and find that $\text{Quantile}=0.8$ performs the best (\cref{fig:compare_distractor_robustnerf}).
We therefore apply this threshold uniformly in other experiments.
To obtain heuristics or binary masks, we follow the procedure in \cite{sabour2023robustnerf} to process the color residual maps.

\begin{table}
    \centering
    \resizebox{\linewidth}{!}{
    \begin{tabular}{@{}l|c|c@{}}
    \toprule
    Scene & Semantic Label & Text Prompt \\
    \midrule
    Car & \textit{car} & \textit{``Moving red car and its shadow. ''} \\
    Cars & \textit{car} & \makecell{\textit{``Moving red car and its shadow. Moving blue car}\\\textit{and its shadow. Moving green car and its shadow. ''}} \\
    Bag & \textit{bag} & \textit{``Moving bag and its shadow. ''} \\
    Chairs & \textit{chair} & \textit{``Moving chairs and their shadows. ''} \\
    Pillow & \textit{pillow, cushion} & \textit{``Moving pillow and its shadow. ''} \\
    \bottomrule
    \end{tabular}}
    \caption{\textbf{Prior knowledge that semantic and open-set segmentation models use on the Kubric dataset.}
    \vspace{-4mm}
    }
    \label{tab:model_prior}
\end{table}

\subsection{Settings on the Baseline Segmentation Models}
\vspace{1mm}
\noindent \textbf{DeepLabv3+~\cite{chen2018encoder} and Mask2Former~\cite{cheng2022masked}.} 
For the two semantic segmentation models, we use the implementation of MMSegmentation~\cite{mmseg2020} that has been pre-trained on the ADE20K dataset~\cite{zhou2019semantic}.
When evaluating on the Kubric dataset, we segment transient objects instead of static ones because the former have fewer types and are easier to identify.
The semantic labels used are shown in \cref{tab:model_prior}.

\vspace{1mm}
\noindent \textbf{Grounded-SAM~\cite{kirillov2023segment,liu2023grounding}.} 
This open-set segmentation model uses Grounding DINO~\cite{liu2023grounding} to locate target objects through natural language and then uses SAM~\cite{kirillov2023segment} to obtain related segmentation masks.
The text prompts used are shown in \cref{tab:model_prior}.

\vspace{1mm}
\noindent \textbf{DINO~\cite{caron2021emerging}.} 
This model is a pre-trained backbone model that can be used for video segmentation~\cite{mirzaei2023spin}.
To segment the target object, DINO requires the object mask in the first video frame as initialization, which we provide in the form of the first ground-truth transient mask.

\section{Additional Experiments}

\begin{figure}
  \centering
  \begin{subfigure}{0.95\linewidth}
    \includegraphics[width=\linewidth]{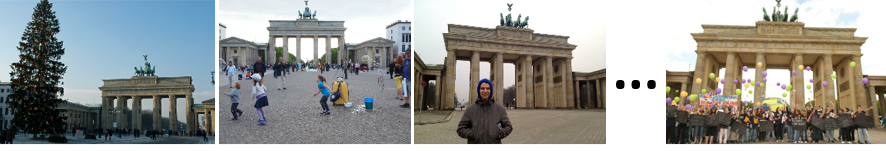}
    \caption{Training Images}
  \end{subfigure} \\
  \begin{subfigure}{0.95\linewidth}
    \includegraphics[width=\linewidth]{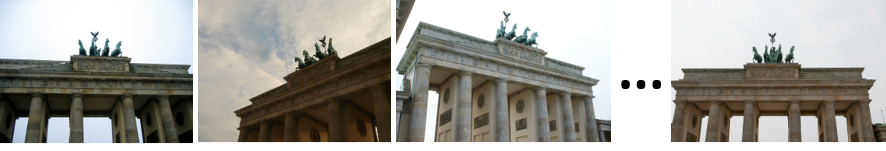}
    \caption{Test Images}
  \end{subfigure}
  \caption{\textbf{Examples of training and test images from the Phototourism dataset.} Note that most of the transient distractors (\eg, tourists and trees) in the training images do not block the landmark main bodies (\eg, the attic storey of the Brandenburg Gate and its top sculptural group), which are the focus of the test images. 
  \vspace{-4mm}
  }
  \label{fig:phototourism_examples}
\end{figure}

\subsection{Comparison on the Phototourism Dataset}
\label{sec:supple_comparison_on_photourism}
As shown in \cref{fig:compare_phototourism} of the main paper, our method brings less improvement on Phototourism compared to other datasets.
Similar to \cite{lee2023semantic}, we ascribe this to: i) most of the transient distractors in its training set do not disturb the landmark main bodies and are located at the boundary of the landmarks or images (\cref{fig:phototourism_examples}); ii) the content of almost all test images concentrates on the landmark main bodies, meaning that models without any processing of transient distractors can also generate acceptable results when evaluating on the test set (see the rendered images of Nerfacto and Mip-NeRF 360 in \cref{fig:compare_phototourism,fig:supplement_visual_render_phototourism}).
Therefore, the improvements achieved on this dataset are mostly from addressing the variable appearances to improve the model's rendering quality~\cite{chen2022hallucinated,rudnev2022nerf,li2023climatenerf,lin2023neural,wang2023neural}, which is orthogonal to our work.

\begin{figure}
  \centering
  \begin{subfigure}{\linewidth}
    \includegraphics[width=\linewidth]{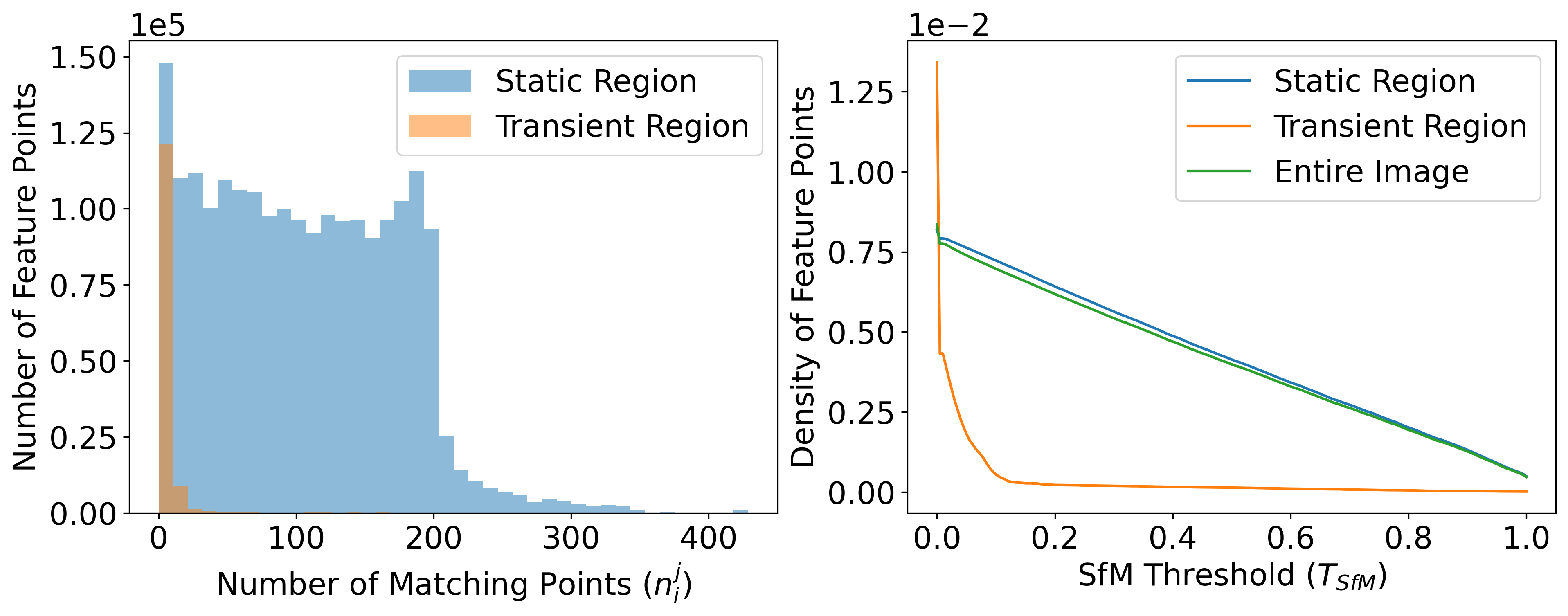}
  \end{subfigure}\\
  \vspace{0.5mm}
  \begin{subfigure}{\linewidth}
    \includegraphics[width=\linewidth]{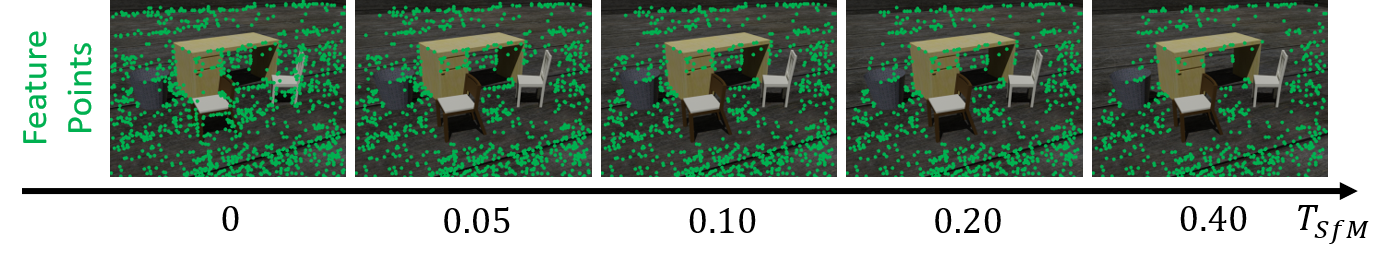}
  \end{subfigure}
  \caption{\textbf{Verification of \cref{ob:feat} on the Kubric dataset. } 
  (top left) The histogram of $n_i^j$ for static and transient regions.
  (top right) As the threshold $\mathcal{T}_{SfM}$ increases, the density of feature points from transient regions decreases much faster than that in static regions.
  (bottom) As a result, feature points in transient regions (\eg, chairs) are gradually removed and the static regions are identified at the same time.
  \vspace{-4mm}
  }
  \label{fig:verify_observation}
\end{figure}

\subsection{Verification of \cref{ob:feat}}
\label{sec:verification_obser1}

\cref{ob:feat} is the basis of our SfM-based heuristics.
To verify its correctness, we conduct experiments on the Kubric dataset since it has ground-truth masks that indicate which image regions are transient.

\vspace{1mm}
\noindent \textbf{Distribution of $\mathbf{n_i^j}$.}
As \cref{fig:verify_observation} (top left) shows, the $n_i^j$ distributions of static and transient regions are significantly different.
Specifically, the $n_i^j$ distribution of the static region is relatively uniform and concentrated in $[0,200]$; in contrast, the $n_i^j$ distribution of the transient region is significantly biased towards 0 and concentrated in $[0,10]$.
These validate our \cref{ob:feat}: most SfM features of static objects have much larger $n_i^j$ than those of transient ones.

\vspace{1mm}
\noindent \textbf{Effectiveness of $\mathbf{\mathcal{T}_{SfM}}$.}
Based on \cref{ob:feat}, we applied a simple yet effective threshold $\mathcal{T}_{SfM}$ to distinguish between static and transient regions.
As shown in \cref{fig:verify_observation} (top right), the density of feature points decreases {\it linearly} as $\mathcal{T}_{SfM}$ increases for both the entire image and static regions, while it decreases {\it exponentially} for transient regions.
When $\mathcal{T}_{SfM}\geq 0.2$, the density of feature points in transient regions is almost $0$, meaning that there are almost no remaining feature points in transient regions (\cref{fig:verify_observation}, bottom), \ie, the static regions can be well-segmented from the remaining feature points.

\subsection{Sensitivity to Hyperparameters}
\label{sec:sensitivity_hyperparameter}
Our method has two hyperparameters: i) $\mathcal{T}_{SfM}$ which is used to filter transient feature points and ii) $\mathcal{T}_{CR}$ which is used as additional insurance to our combined heuristics (\cref{sec:heuristics_development} of the main paper).
Following the main paper, we conduct experiments based on Nerfacto and investigate how the values of $\mathcal{T}_{SfM}$ and $\mathcal{T}_{CR}$ affect the performance of our method.

\begin{figure}
  \centering
  \includegraphics[width=\linewidth]{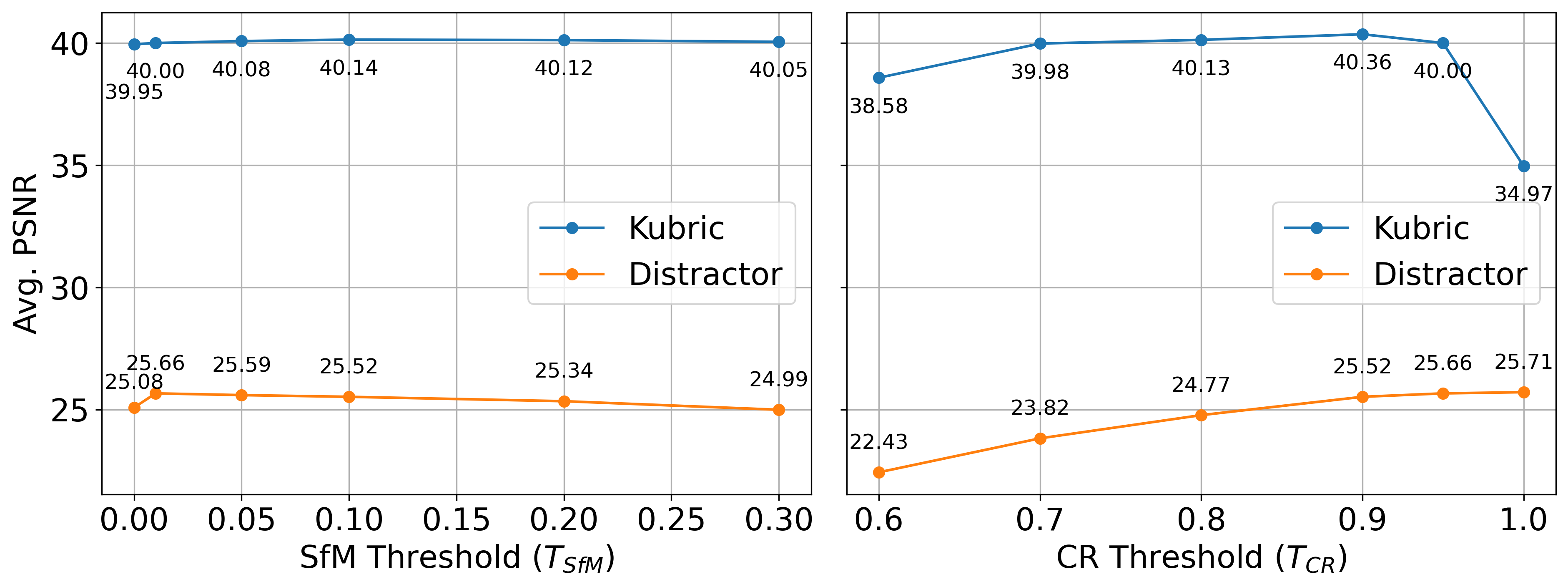}
  \caption{\textbf{Sensitivity to $\mathcal{T}_{SfM}$ and $\mathcal{T}_{CR}$ on different datasets.}  (left) $\mathcal{T}_{CR}$ is fixed at $0.95$ and (right) $\mathcal{T}_{SfM}$ is fixed at $0.01$. 
  }
  \label{fig:sensitivity}
\end{figure}

\vspace{1mm}
\noindent \textbf{Sensitivity to $\mathbf{\mathcal{T}_{SfM}}$.} 
As shown in \cref{fig:sensitivity} (left), the average PSNRs first slightly increase and then slightly decrease as $\mathcal{T}_{SfM}$ gradually increases on both the Kubric and Distractor datasets.
This indicates that the best $\mathcal{T}_{SfM}$ is achieved when striking a balance between removing transient features and retaining static features.
Nevertheless, the improvement brought about by the choice of $\mathcal{T}_{SfM}$ is up to $0.19$dB on the Kubric and $0.58$dB on the Distractor dataset, which suggests that even without careful choice of $\mathcal{T}_{SfM}$, our method achieves a performance superior to the state-of-the-art methods.


\vspace{1mm}
\noindent \textbf{Sensitivity to $\mathbf{\mathcal{T}_{CR}}$. }
As shown in \cref{fig:sensitivity} (right), consistent with the results of RobustNeRF (\cref{fig:robustnerf_with_diff_threshold} of the main paper and \cref{fig:compare_distractor_robustnerf}), using a smaller $\mathcal{T}_{CR}$ usually causes the performance to decline as it unnecessarily removes some static objects while using a overly large $\mathcal{T}_{CR}$ (\eg, $\mathcal{T}_{CR}=1$) makes it ineffective by losing the capability of removing ``obvious'' transient objects as insurance.
These validate our choice of using a relatively large $\mathcal{T}_{CR}=0.95$ as additional insurance.  


\begin{figure}
  \centering
  \includegraphics[width=\linewidth]{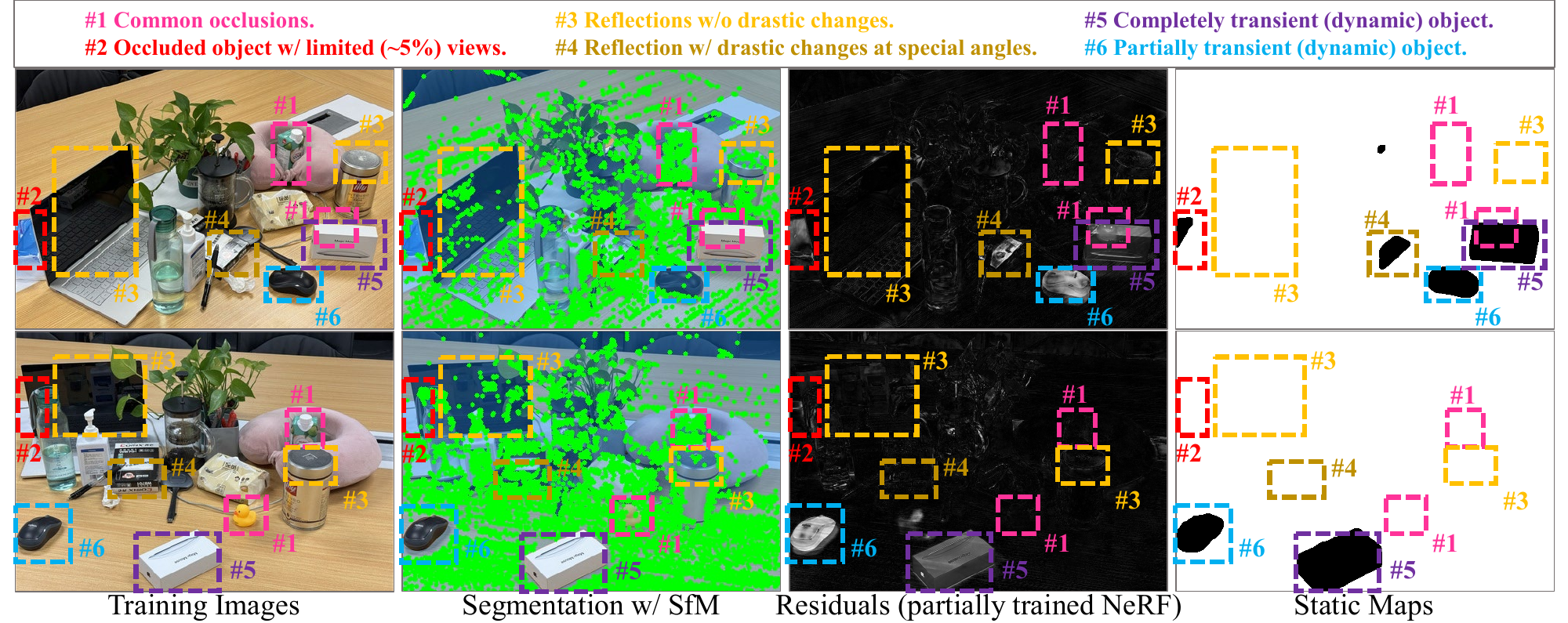}
  \caption{\textbf{Visualizations of HuGS in the real-world scene \textit{Room}}. Zoom in for a better view.}
  \label{fig:room_segment}
  \vspace{-4mm}
\end{figure}

\subsection{Performance in Challenging Cases}
To enable a more in-depth analysis of our method's performance, we introduce a novel real-world scene \textit{Room} with a variety of different materials (\eg, glass, liquid and metal), featuring challenging cases of occlusion, reflection and movement.
It contains 64 images.
A white box acts as transient distractors that changes its position in every image, and
a black mouse is partially dynamic that keep moving in half of the images and static in the other half.
As \cref{fig:room_segment} shows:
\begin{enumerate}
    \item \textbf{Occlusion}. Our method excels in dealing with common occlusions (\#1) but struggles with extreme cases (\#2) that are rarely visible as they provide few features or supervision signals for SfM matching or NeRF fitting.
    \item \textbf{Reflection.} Our method handles reflections well (\#3) if there are no drastic appearance changes. However, limited views of highly reflective objects may induce such changes (\#4), leading to misclassification as transient/static due to larger residuals from unfitted NeRF even with successful SfM matching.
    \item \textbf{Movement.} Partially dynamic/transient objects are inherently ambiguous, leading to relatively large residuals from unfitted NeRF. For half-dynamic objects (static in 50\% views), our method tends to treat them (\#6) as transient distractors. The specific outcome, however, depends on the degree of ``partially dynamic''.
\end{enumerate}

\begin{figure}
  \centering
  \begin{subfigure}{0.24\linewidth}
    \includegraphics[width=\linewidth]{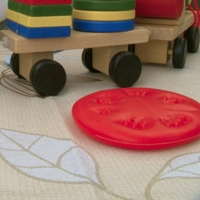}
    \caption*{Training Image}
  \end{subfigure}
  \hfill
  \begin{subfigure}{0.24\linewidth}
    \includegraphics[width=\linewidth]{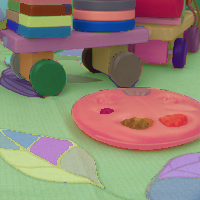}
    \caption*{SAM Result}
  \end{subfigure}
  \hfill
  \begin{subfigure}{0.24\linewidth}
    \includegraphics[width=\linewidth]{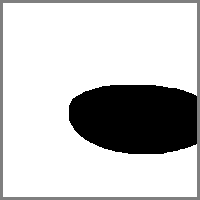}
    \caption*{Static Map}
  \end{subfigure}
  \hfill
  \begin{subfigure}{0.24\linewidth}
    \includegraphics[width=\linewidth]{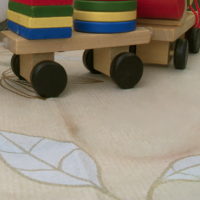}
    \caption*{Rendered Image}
  \end{subfigure}
  \caption{\textbf{A failure case.}
  SAM sometimes fails to segment the shadow around transient distractors (\eg, the red plate in the image) and incorrectly counts it in the static map, resulting in artifacts in the rendered image. 
  }
  \label{fig:failure_case}
\end{figure}

\begin{figure}
  \centering
  \resizebox{\linewidth}{!}{
  \begin{tabular}{ll|ccc}
    \toprule
    & \multirow{2}*{Training Data} & \multicolumn{3}{c}{Kubric (Avg.)} \\
    & & PSNR$\uparrow$ & SSIM$\uparrow$ & LPIPS$\downarrow$ \\
    \midrule
    (a) & Noisy Images w/ Our Static Maps & 40.40 & .969 & .033 \\
    (b) & Noisy Images w/ GT static Maps & 40.49 & .969 & .032 \\
    (c) & Clean Images & \textbf{40.95} & \textbf{.971} & \textbf{.031} \\
    \bottomrule
  \end{tabular}}\\
  \vspace{1mm}
  \begin{subfigure}{0.185\linewidth}
    \includegraphics[width=\linewidth]{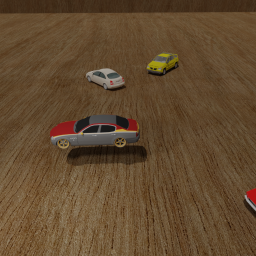}
    \caption*{Noisy Image}
  \end{subfigure}
  \rotatebox{90}{\scriptsize ~~~~~~~~~~Color Image}
  \begin{subfigure}{0.185\linewidth}
    \includegraphics[width=\linewidth]{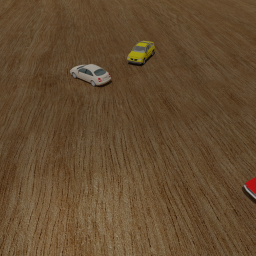}
    \caption*{Test Image}
  \end{subfigure}
  \hfill
  \begin{subfigure}{0.185\linewidth}
    \includegraphics[width=\linewidth]{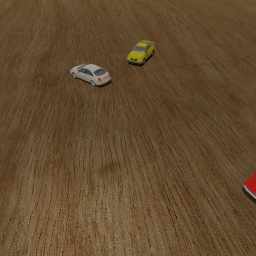}
    \caption*{(a)}
  \end{subfigure}
  \hfill
  \begin{subfigure}{0.185\linewidth}
    \includegraphics[width=\linewidth]{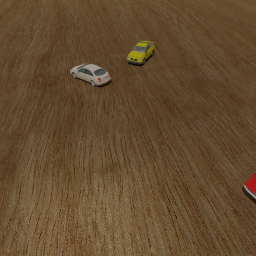}
    \caption*{(b)}
  \end{subfigure}
  \hfill
  \begin{subfigure}{0.185\linewidth}
    \includegraphics[width=\linewidth]{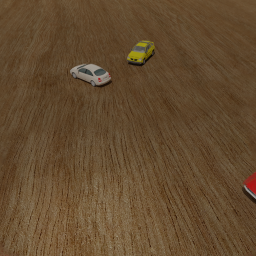}
    \caption*{(c)}
  \end{subfigure}\\
  \begin{subfigure}{0.185\linewidth}
    \includegraphics[width=\linewidth]{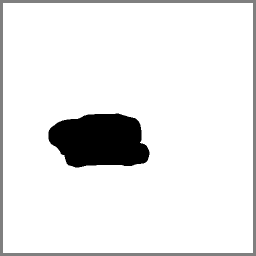}
    \caption*{GT Map}
  \end{subfigure}
  \rotatebox{90}{\scriptsize ~~~~~~~~~~Depth Map}
  \begin{subfigure}{0.185\linewidth}
    \includegraphics[width=\linewidth]{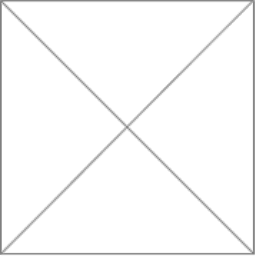}
    \caption*{}
  \end{subfigure}
  \hfill
  \begin{subfigure}{0.185\linewidth}
    \includegraphics[width=\linewidth]{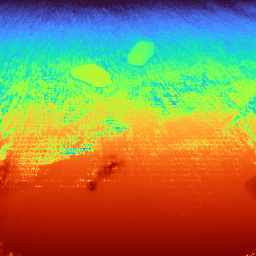}
    \caption*{(a)}
  \end{subfigure}
  \hfill
  \begin{subfigure}{0.185\linewidth}
    \includegraphics[width=\linewidth]{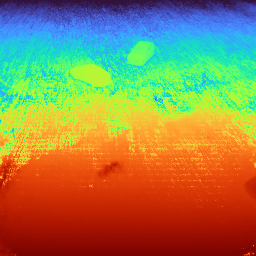}
    \caption*{(b)}
  \end{subfigure}
  \hfill
  \begin{subfigure}{0.185\linewidth}
    \includegraphics[width=\linewidth]{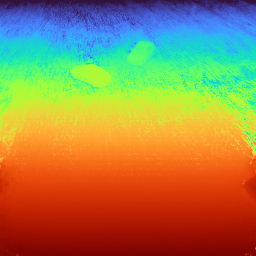}
    \caption*{(c)}
  \end{subfigure}
  \caption{\textbf{Performance of Nerfacto with different kinds of training data. } The \textbf{best results} are highlighted. Even if ground-truth static maps are provided, simply removing large areas of transient pixels during training may lead to a lack of supervision in some spatial regions and viewing directions, resulting in incorrect densities or biased colors (\eg, floater in the middle).
  \vspace{-4mm}
  }
  \label{fig:additional_failure_case}
\end{figure}

\section{Limitations and Future Work}

Our method relies on the effective design of heuristics and the power of the segmentation model used.
Despite this, our method is still limited when facing extreme cases in real-world scenes, such as rarely visible static objects and highly reflective materials (\cref{fig:room_segment}).
In our work, we used a naive SfM algorithm, which could be further improved to handle cases of sparse views or scenes dominated by transient distractors.
Additionally, although SAM is one of the most advanced segmentation models, it still cannot handle very slight shadows. This limitation results in reduced rendering quality in some cases (\cref{fig:failure_case}).
Therefore, utilizing more robust SfM algorithms and/or more capable segmentation models represents a potential path to further improve our method.
Furthermore, our method removes the transient pixels using Eq.~\ref{eq:nerf_static_scenes} without additional processing of them. Excessive pixel removal can result in insufficient supervisory signals, potentially degrading the model's performance relative to training on clean images (\cref{fig:additional_failure_case}).  
Therefore, another possible improvement is to combine our method with 2D/3D inpainting~\cite{mirzaei2023spin,weder2023removing,yin2023or,suvorov2022resolution}, which can predict the missing content caused by removing transient pixels to assist NeRF training.

\section{Additional Qualitative Results}

\begin{figure*}
  \centering
  \rotatebox{90}{\scriptsize $<$--------------- Pillow --------------$>$$<$--------------- Chairs ---------------$>$$<$----------------- Bag -----------------$>$$<$----------------- Cars ----------------$>$$<$----------------- Car -----------------$>$}
  \hfill
  \begin{subfigure}{0.1364\linewidth}
    \caption*{Training Image}
    \includegraphics[width=\linewidth]{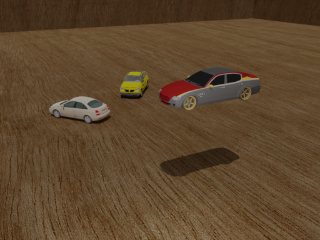}\\
    \includegraphics[width=\linewidth]{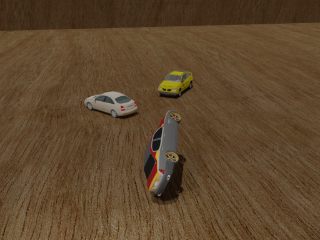}\\
    \includegraphics[width=\linewidth]{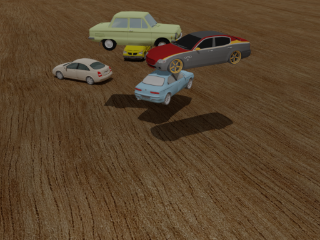}\\
    \includegraphics[width=\linewidth]{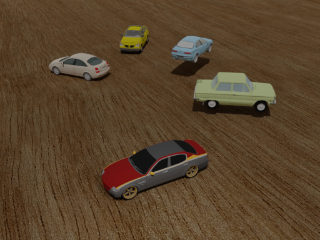}\\
    \includegraphics[width=\linewidth]{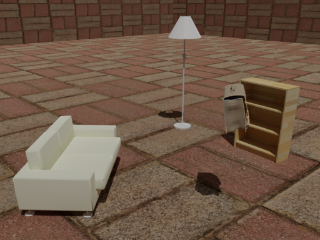}\\
    \includegraphics[width=\linewidth]{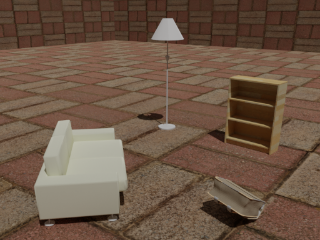}\\
    \includegraphics[width=\linewidth]{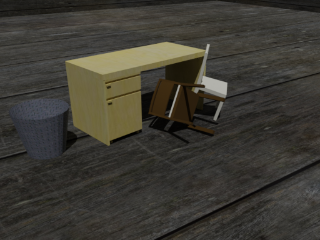}\\
    \includegraphics[width=\linewidth]{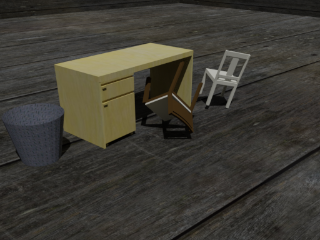}\\
    \includegraphics[width=\linewidth]{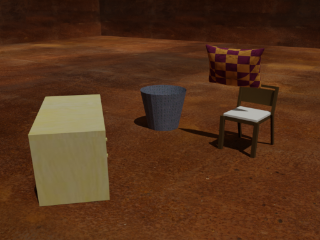}\\
    \includegraphics[width=\linewidth]{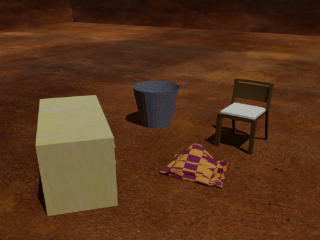}
  \end{subfigure}
  \hfill
  \begin{subfigure}{0.1364\linewidth}
    \caption*{Seg. w/ SfM}
    \includegraphics[width=\linewidth]{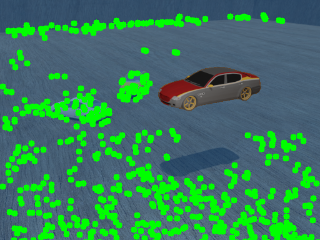}\\
    \includegraphics[width=\linewidth]{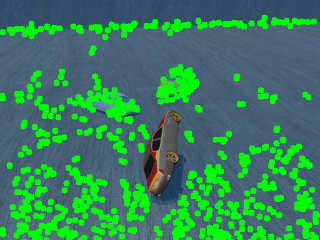}\\
    \includegraphics[width=\linewidth]{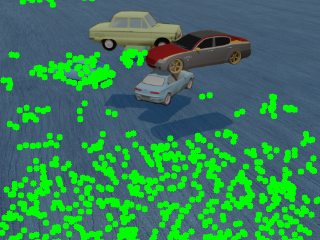}\\
    \includegraphics[width=\linewidth]{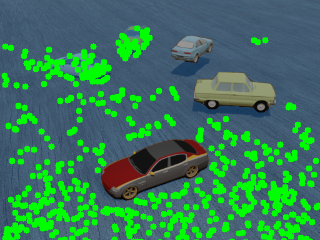}\\
    \includegraphics[width=\linewidth]{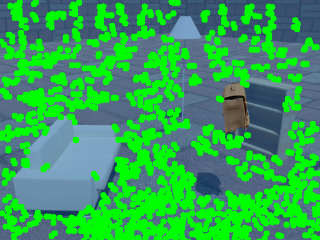}\\
    \includegraphics[width=\linewidth]{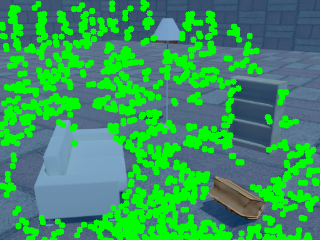}\\
    \includegraphics[width=\linewidth]{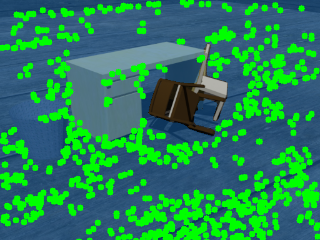}\\
    \includegraphics[width=\linewidth]{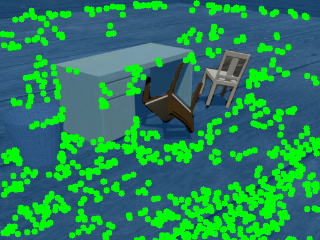}\\
    \includegraphics[width=\linewidth]{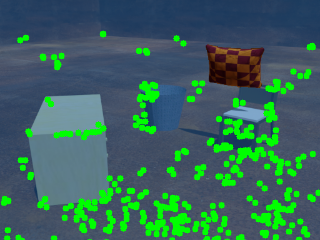}\\
    \includegraphics[width=\linewidth]{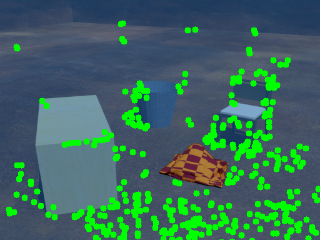}
  \end{subfigure}
  \hfill
  \begin{subfigure}{0.1364\linewidth}
    \caption*{$\mathcal{H}^{SfM}$}
    \includegraphics[width=\linewidth]{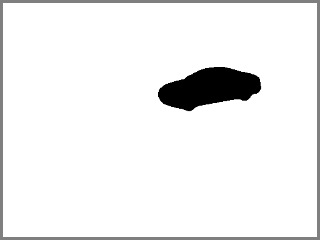}\\
    \includegraphics[width=\linewidth]{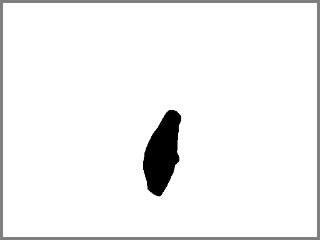}\\
    \includegraphics[width=\linewidth]{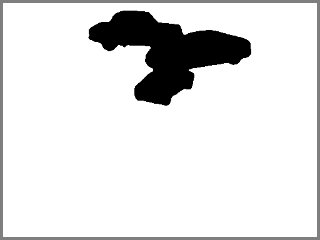}\\
    \includegraphics[width=\linewidth]{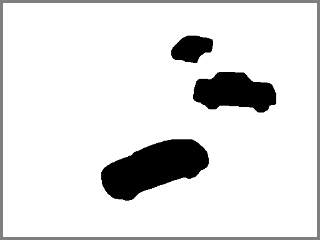}\\
    \includegraphics[width=\linewidth]{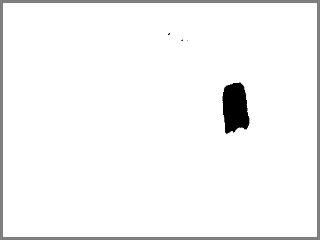}\\
    \includegraphics[width=\linewidth]{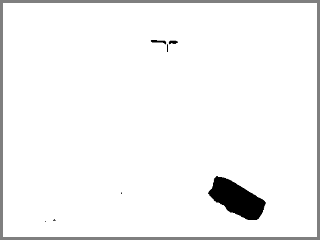}\\
    \includegraphics[width=\linewidth]{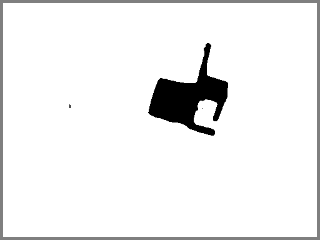}\\
    \includegraphics[width=\linewidth]{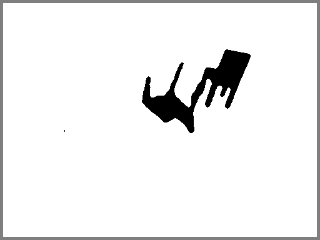}\\
    \includegraphics[width=\linewidth]{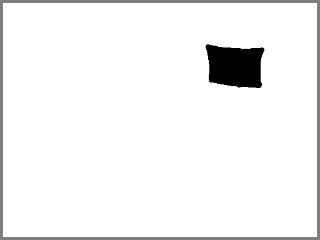}\\
    \includegraphics[width=\linewidth]{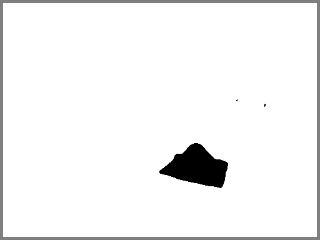}
  \end{subfigure}
  \hfill
  \begin{subfigure}{0.1364\linewidth}
    \caption*{Color Residual}
    \includegraphics[width=\linewidth]{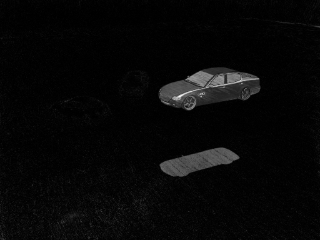}\\
    \includegraphics[width=\linewidth]{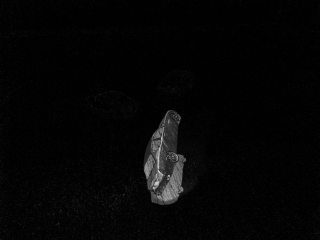}\\
    \includegraphics[width=\linewidth]{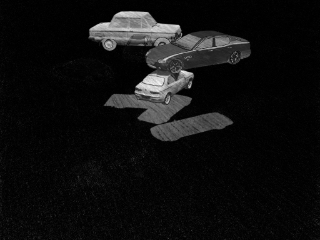}\\
    \includegraphics[width=\linewidth]{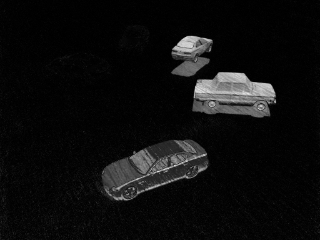}\\
    \includegraphics[width=\linewidth]{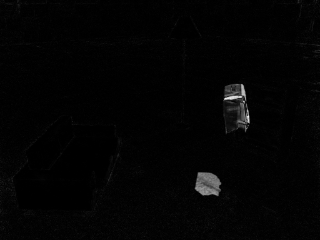}\\
    \includegraphics[width=\linewidth]{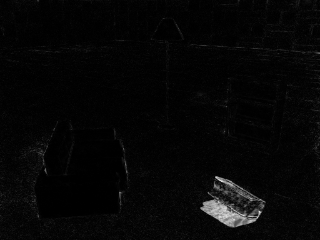}\\
    \includegraphics[width=\linewidth]{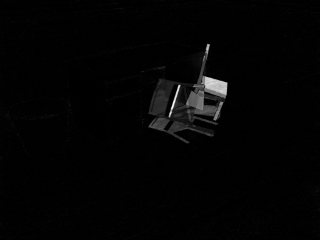}\\
    \includegraphics[width=\linewidth]{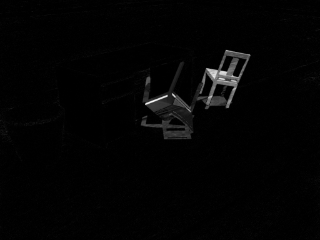}\\
    \includegraphics[width=\linewidth]{images/visual_segmentation/kubric_pillow_rgba_00021_residual.png}\\
    \includegraphics[width=\linewidth]{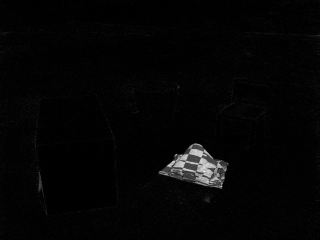}
  \end{subfigure}
  \hfill
  \begin{subfigure}{0.1364\linewidth}
    \caption*{$\mathcal{H}^{CR}$}
    \includegraphics[width=\linewidth]{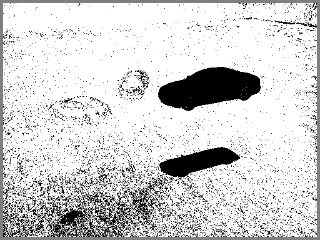}\\
    \includegraphics[width=\linewidth]{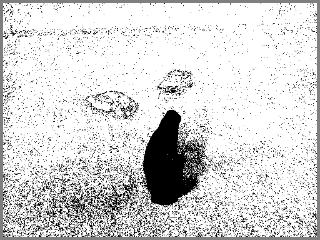}\\
    \includegraphics[width=\linewidth]{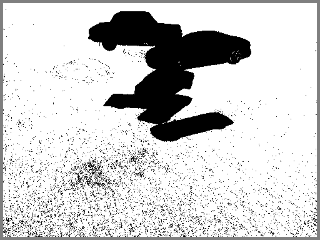}\\
    \includegraphics[width=\linewidth]{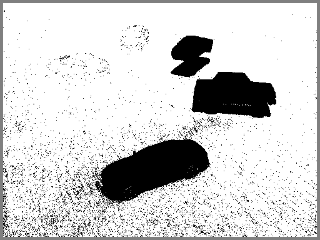}\\
    \includegraphics[width=\linewidth]{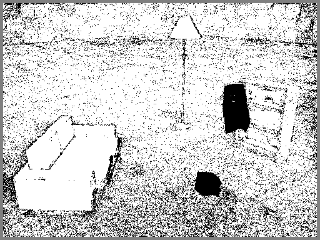}\\
    \includegraphics[width=\linewidth]{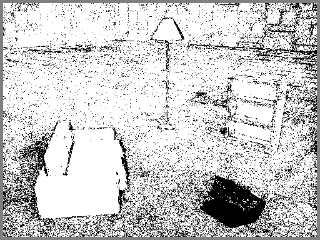}\\
    \includegraphics[width=\linewidth]{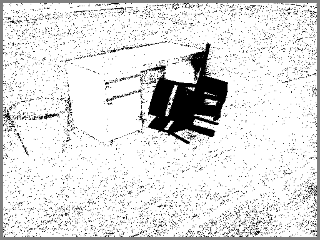}\\
    \includegraphics[width=\linewidth]{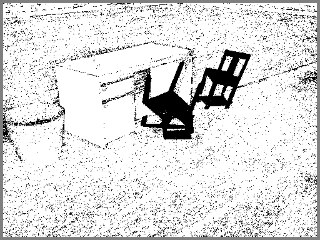}\\
    \includegraphics[width=\linewidth]{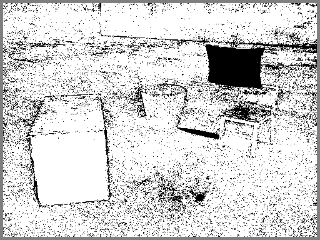}\\
    \includegraphics[width=\linewidth]{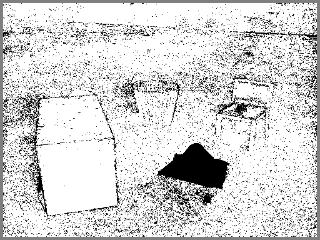}
  \end{subfigure}
  \hfill
  \begin{subfigure}{0.1364\linewidth}
    \caption*{Combined $\mathcal{H}$}
    \includegraphics[width=\linewidth]{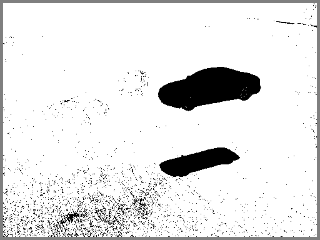}\\
    \includegraphics[width=\linewidth]{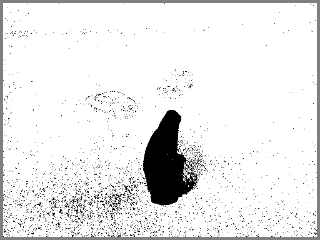}\\
    \includegraphics[width=\linewidth]{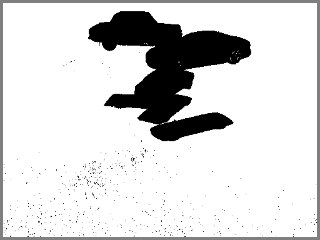}\\
    \includegraphics[width=\linewidth]{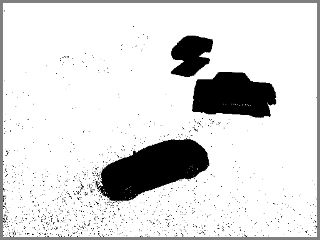}\\
    \includegraphics[width=\linewidth]{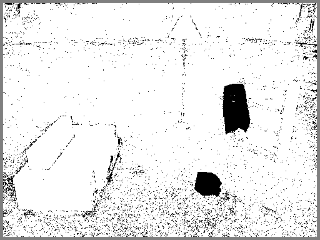}\\
    \includegraphics[width=\linewidth]{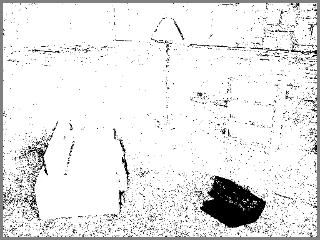}\\
    \includegraphics[width=\linewidth]{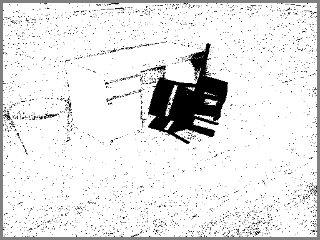}\\
    \includegraphics[width=\linewidth]{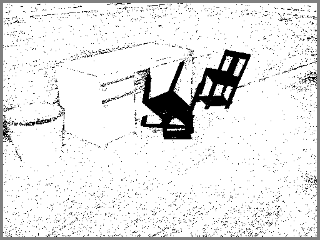}\\
    \includegraphics[width=\linewidth]{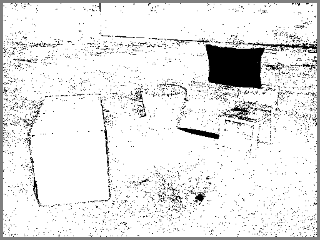}\\
    \includegraphics[width=\linewidth]{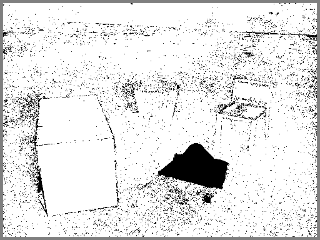}
  \end{subfigure}
  \hfill
  \begin{subfigure}{0.1364\linewidth}
    \caption*{Static Map}
    \includegraphics[width=\linewidth]{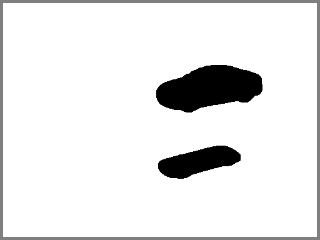}\\
    \includegraphics[width=\linewidth]{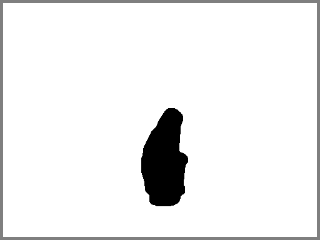}\\
    \includegraphics[width=\linewidth]{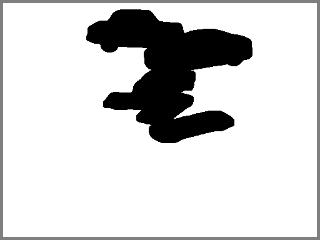}\\
    \includegraphics[width=\linewidth]{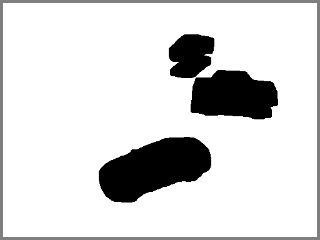}\\
    \includegraphics[width=\linewidth]{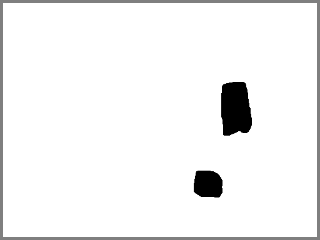}\\
    \includegraphics[width=\linewidth]{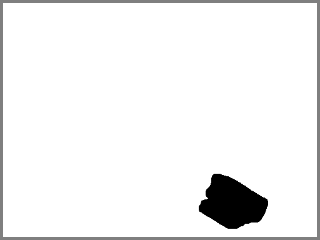}\\
    \includegraphics[width=\linewidth]{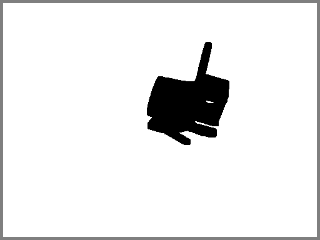}\\
    \includegraphics[width=\linewidth]{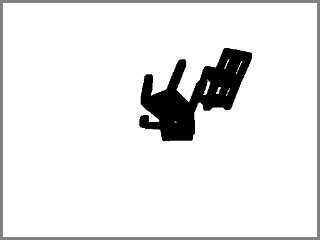}\\
    \includegraphics[width=\linewidth]{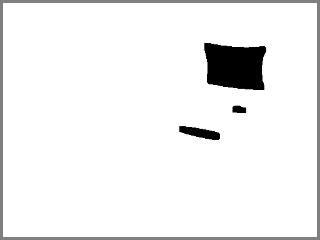}\\
    \includegraphics[width=\linewidth]{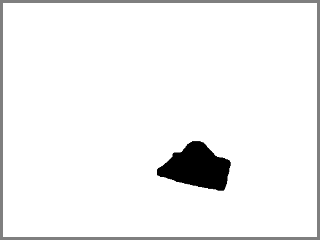}
  \end{subfigure}
  \caption{\textbf{Visualization of HuGS on the Kubric dataset. }
  \vspace{-4mm}
  }
  \label{fig:supplement_visual_segment_kubric}
\end{figure*}

\begin{figure*}
  \centering
  \rotatebox{90}{\scriptsize $<$----------------------- BabyYoda ------------------------$>$$<$--------------------------- Crab ----------------------------$>$$<$------------------------- Android -------------------------$>$$<$-------------------------- Statue --------------------------$>$}
  \hfill
  \begin{subfigure}{0.1364\linewidth}
    \caption*{Training Image}
    \includegraphics[width=\linewidth]{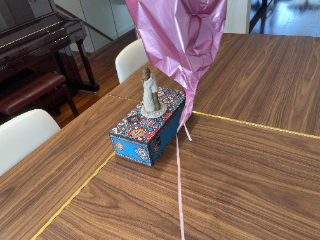}\\
    \includegraphics[width=\linewidth]{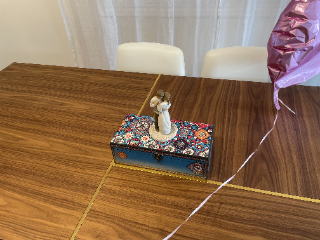}\\
    \includegraphics[width=\linewidth]{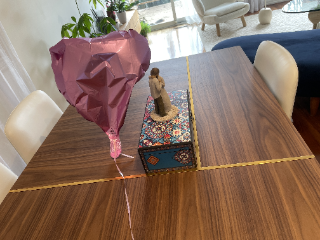}\\
    \includegraphics[width=\linewidth]{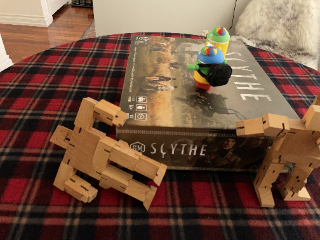}\\
    \includegraphics[width=\linewidth]{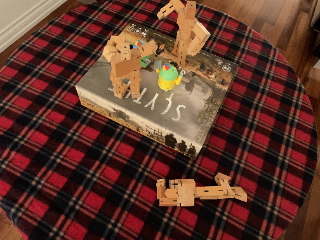}\\
    \includegraphics[width=\linewidth]{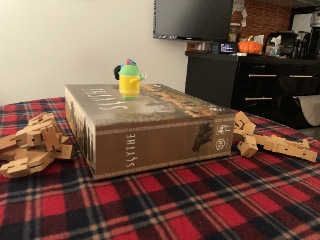}\\
    \includegraphics[width=\linewidth]{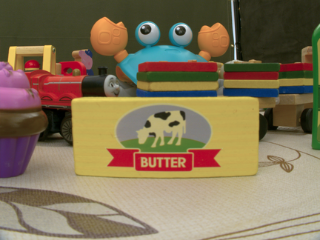}\\
    \includegraphics[width=\linewidth]{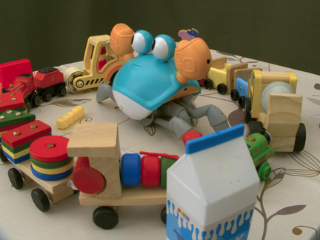}\\
    \includegraphics[width=\linewidth]{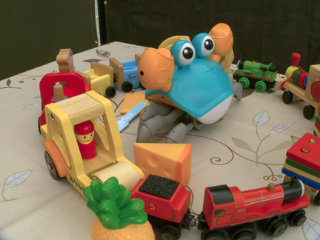}\\
    \includegraphics[width=\linewidth]{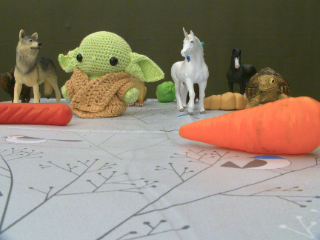}\\
    \includegraphics[width=\linewidth]{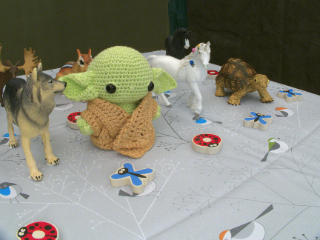}\\
    \includegraphics[width=\linewidth]{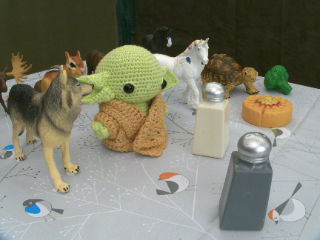}
  \end{subfigure}
  \hfill
  \begin{subfigure}{0.1364\linewidth}
    \caption*{Seg. w/ SfM}
    \includegraphics[width=\linewidth]{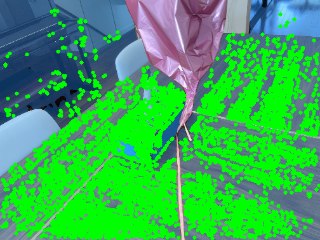}\\
    \includegraphics[width=\linewidth]{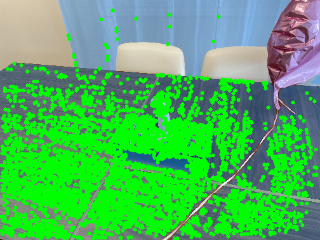}\\
    \includegraphics[width=\linewidth]{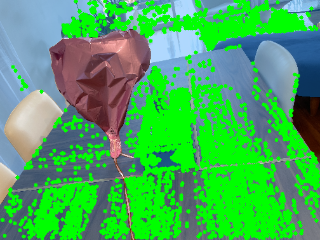}\\
    \includegraphics[width=\linewidth]{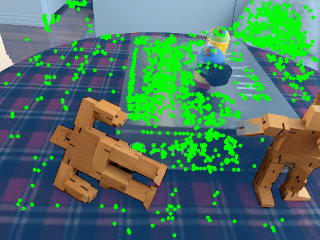}\\
    \includegraphics[width=\linewidth]{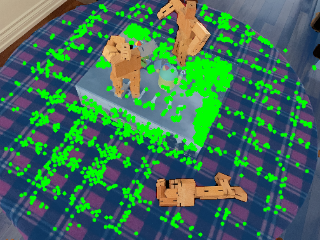}\\
    \includegraphics[width=\linewidth]{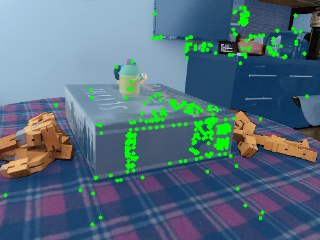}\\
    \includegraphics[width=\linewidth]{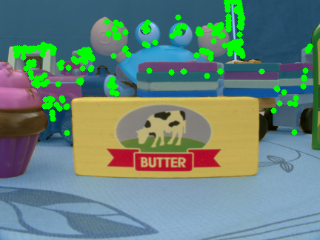}\\
    \includegraphics[width=\linewidth]{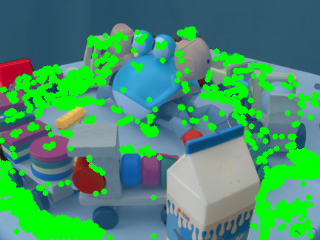}\\
    \includegraphics[width=\linewidth]{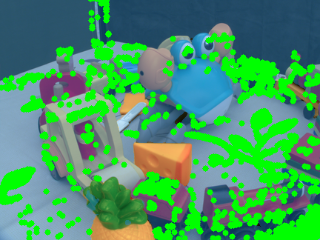}\\
    \includegraphics[width=\linewidth]{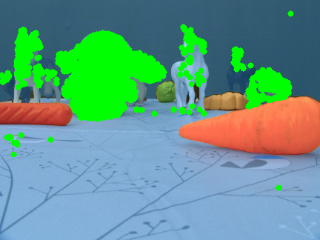}\\
    \includegraphics[width=\linewidth]{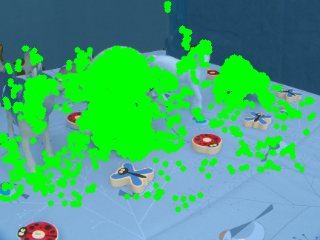}\\
    \includegraphics[width=\linewidth]{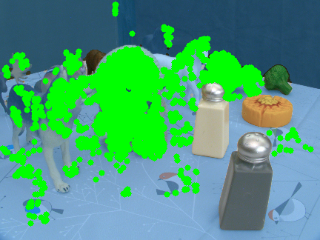}
  \end{subfigure}
  \hfill
  \begin{subfigure}{0.1364\linewidth}
    \caption*{$\mathcal{H}^{SfM}$}
    \includegraphics[width=\linewidth]{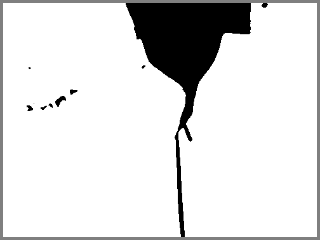}\\
    \includegraphics[width=\linewidth]{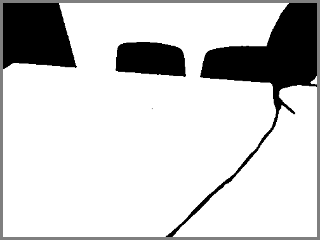}\\
    \includegraphics[width=\linewidth]{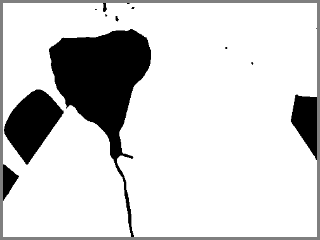}\\
    \includegraphics[width=\linewidth]{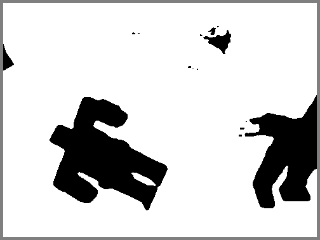}\\
    \includegraphics[width=\linewidth]{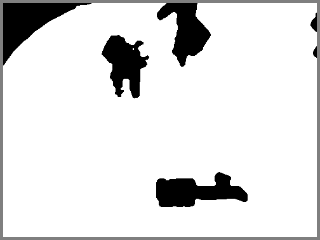}\\
    \includegraphics[width=\linewidth]{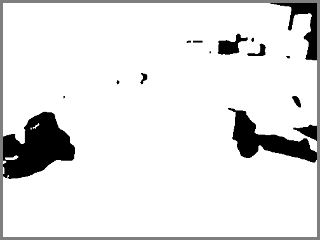}\\
    \includegraphics[width=\linewidth]{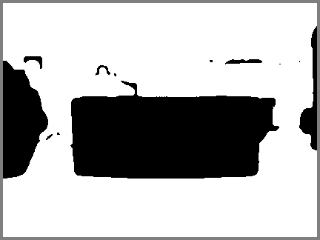}\\
    \includegraphics[width=\linewidth]{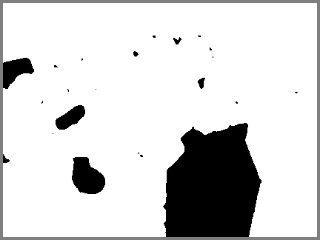}\\
    \includegraphics[width=\linewidth]{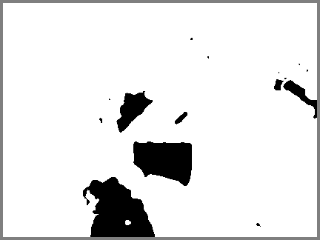}\\
    \includegraphics[width=\linewidth]{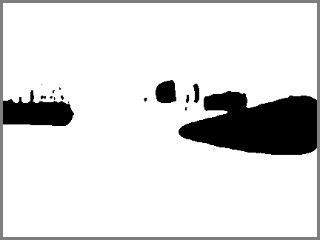}\\
    \includegraphics[width=\linewidth]{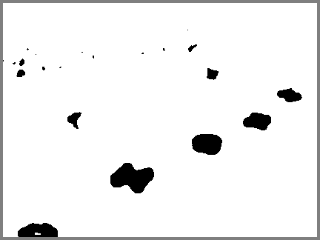}\\
    \includegraphics[width=\linewidth]{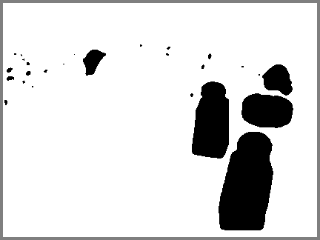}
  \end{subfigure}
  \hfill
  \begin{subfigure}{0.1364\linewidth}
    \caption*{Color Residual}
    \includegraphics[width=\linewidth]{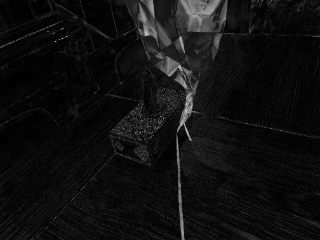}\\
    \includegraphics[width=\linewidth]{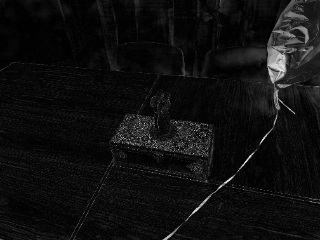}\\
    \includegraphics[width=\linewidth]{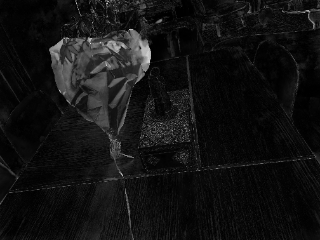}\\
    \includegraphics[width=\linewidth]{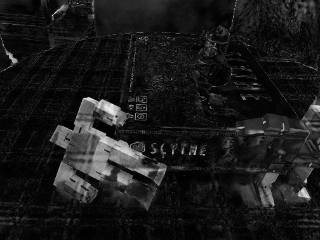}\\
    \includegraphics[width=\linewidth]{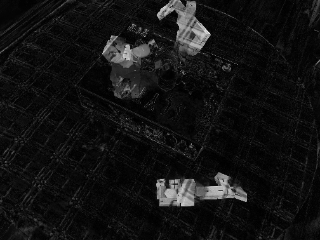}\\
    \includegraphics[width=\linewidth]{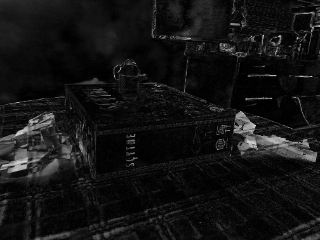}\\
    \includegraphics[width=\linewidth]{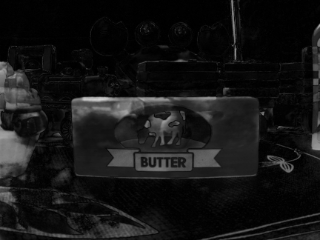}\\
    \includegraphics[width=\linewidth]{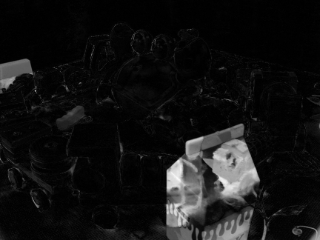}\\
    \includegraphics[width=\linewidth]{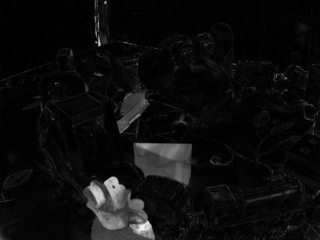}\\
    \includegraphics[width=\linewidth]{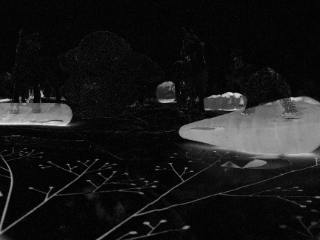}\\
    \includegraphics[width=\linewidth]{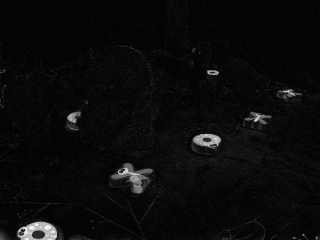}\\
    \includegraphics[width=\linewidth]{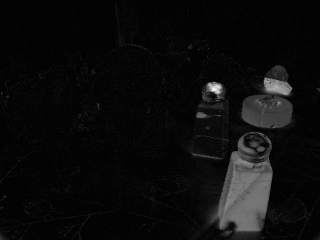}
  \end{subfigure}
  \hfill
  \begin{subfigure}{0.1364\linewidth}
    \caption*{$\mathcal{H}^{CR}$}
    \includegraphics[width=\linewidth]{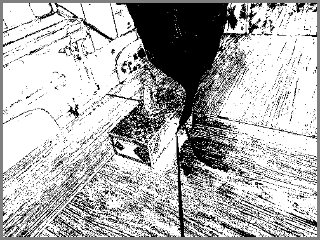}\\
    \includegraphics[width=\linewidth]{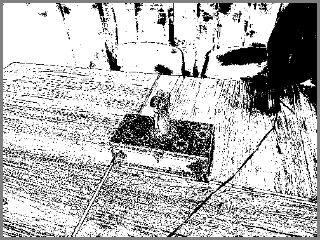}\\
    \includegraphics[width=\linewidth]{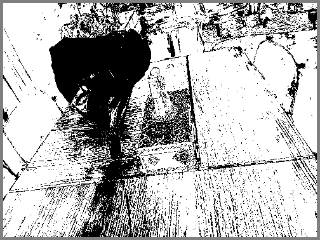}\\
    \includegraphics[width=\linewidth]{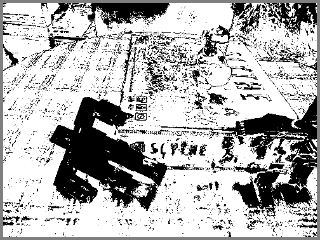}\\
    \includegraphics[width=\linewidth]{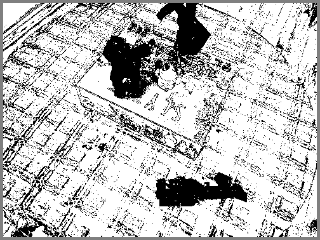}\\
    \includegraphics[width=\linewidth]{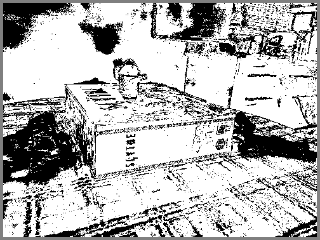}\\
    \includegraphics[width=\linewidth]{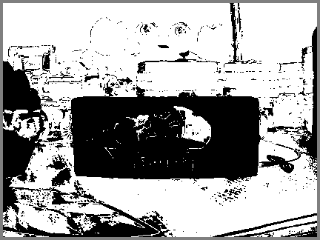}\\
    \includegraphics[width=\linewidth]{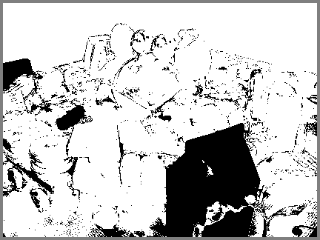}\\
    \includegraphics[width=\linewidth]{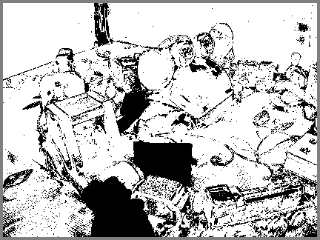}\\
    \includegraphics[width=\linewidth]{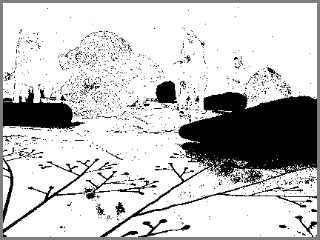}\\
    \includegraphics[width=\linewidth]{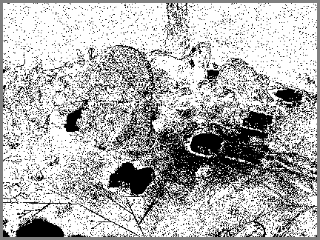}\\
    \includegraphics[width=\linewidth]{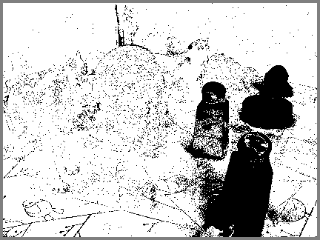}
  \end{subfigure}
  \hfill
  \begin{subfigure}{0.1364\linewidth}
    \caption*{Combined $\mathcal{H}$}
    \includegraphics[width=\linewidth]{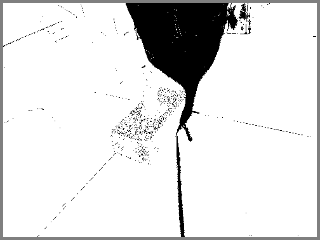}\\
    \includegraphics[width=\linewidth]{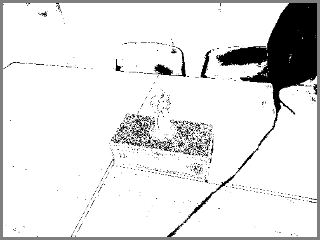}\\
    \includegraphics[width=\linewidth]{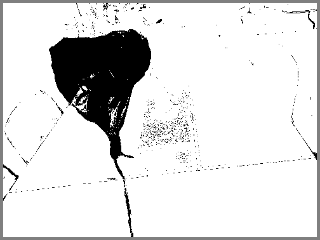}\\
    \includegraphics[width=\linewidth]{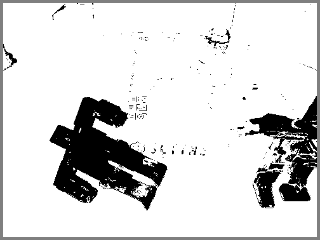}\\
    \includegraphics[width=\linewidth]{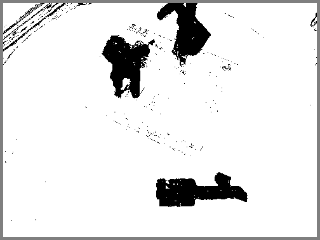}\\
    \includegraphics[width=\linewidth]{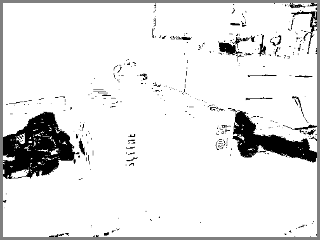}\\
    \includegraphics[width=\linewidth]{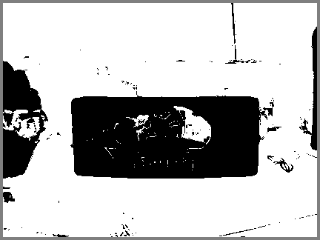}\\
    \includegraphics[width=\linewidth]{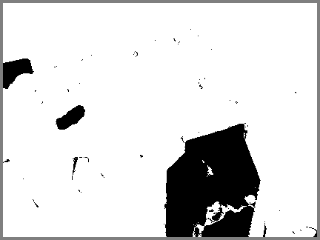}\\
    \includegraphics[width=\linewidth]{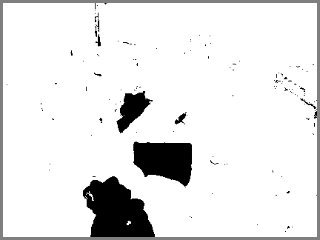}\\
    \includegraphics[width=\linewidth]{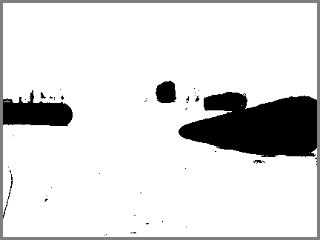}\\
    \includegraphics[width=\linewidth]{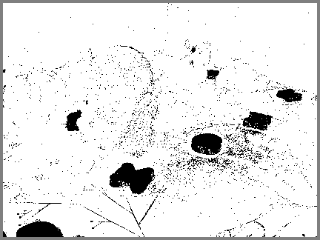}\\
    \includegraphics[width=\linewidth]{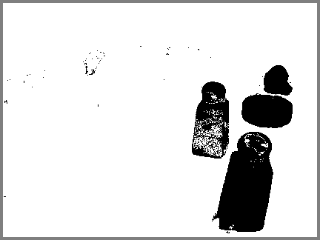}
  \end{subfigure}
  \hfill
  \begin{subfigure}{0.1364\linewidth}
    \caption*{Static Map}
    \includegraphics[width=\linewidth]{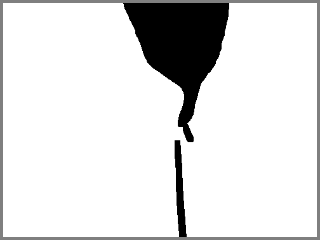}\\
    \includegraphics[width=\linewidth]{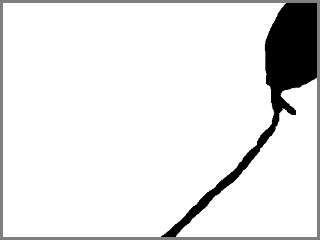}\\
    \includegraphics[width=\linewidth]{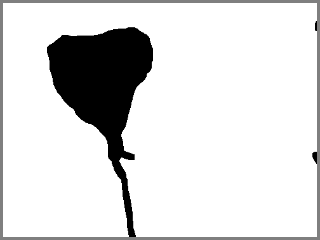}\\
    \includegraphics[width=\linewidth]{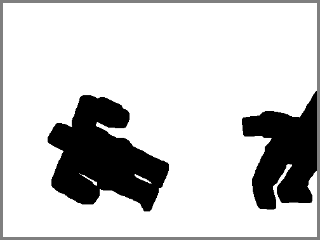}\\
    \includegraphics[width=\linewidth]{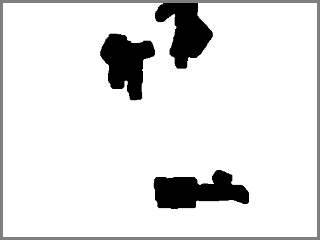}\\
    \includegraphics[width=\linewidth]{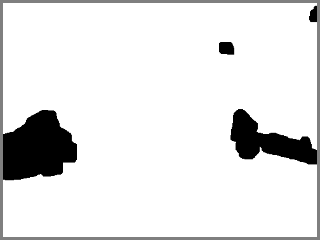}\\
    \includegraphics[width=\linewidth]{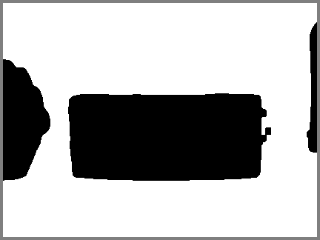}\\
    \includegraphics[width=\linewidth]{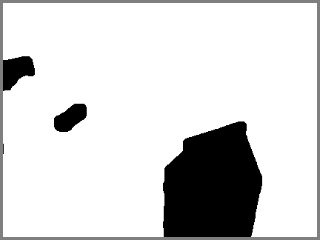}\\
    \includegraphics[width=\linewidth]{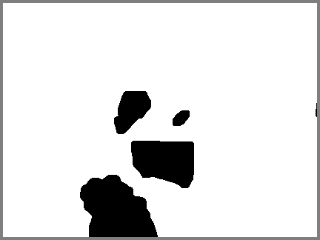}\\
    \includegraphics[width=\linewidth]{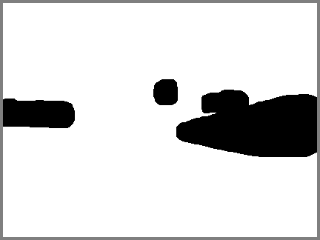}\\
    \includegraphics[width=\linewidth]{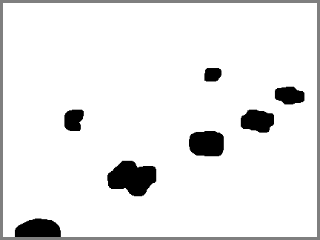}\\
    \includegraphics[width=\linewidth]{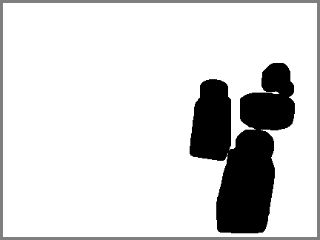}
  \end{subfigure}
  \caption{\textbf{Visualization of HuGS on the Distractor dataset. }
  \vspace{-4mm}
  }
  \label{fig:supplement_visual_segment_distractor}
\end{figure*}

\begin{figure*}
  \centering
  \rotatebox{90}{\scriptsize $<$-------------------- Trevi Fountain ---------------------$>$$<$-------------------- Taj Mahal ---------------------$>$$<$------------------ Sacre Coeur -------------------$>$$<$--------------- Brandenburg Gate ----------------$>$}
  \begin{subfigure}{0.1364\linewidth}
    \caption*{Training Image}
    \includegraphics[width=\linewidth]{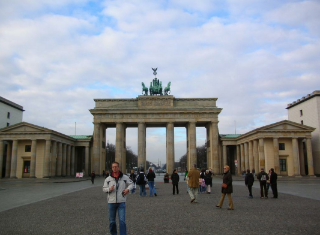}\\
    \includegraphics[width=\linewidth]{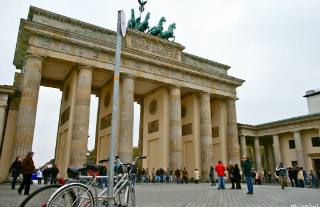}\\
    \includegraphics[width=\linewidth]{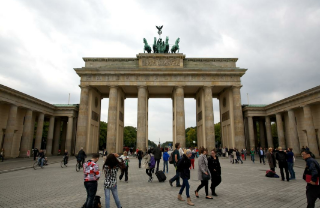}\\
    \includegraphics[width=\linewidth]{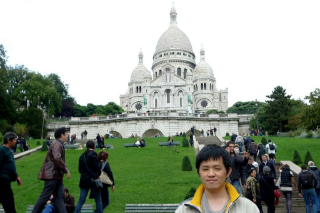}\\
    \includegraphics[width=\linewidth]{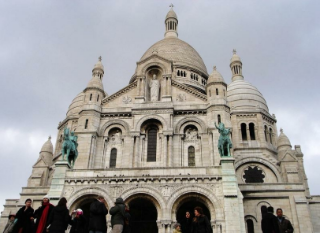}\\
    \includegraphics[width=\linewidth]{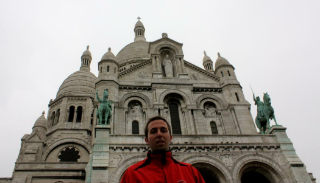}\\
    \includegraphics[width=\linewidth]{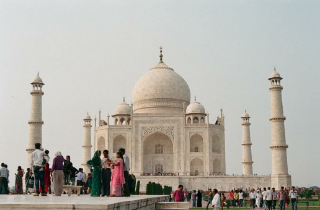}\\
    \includegraphics[width=\linewidth]{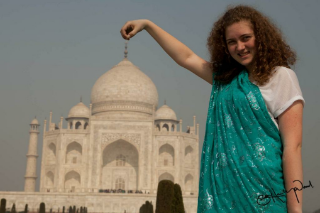}\\
    \includegraphics[width=\linewidth]{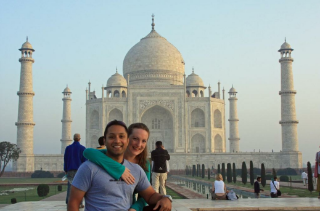}\\
    \includegraphics[width=\linewidth]{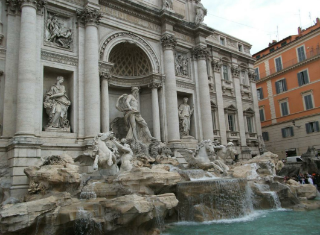}\\
    \includegraphics[width=\linewidth]{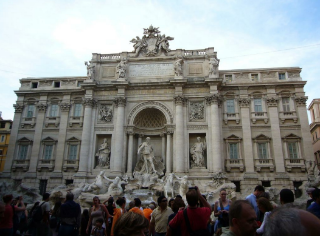}\\
    \includegraphics[width=\linewidth]{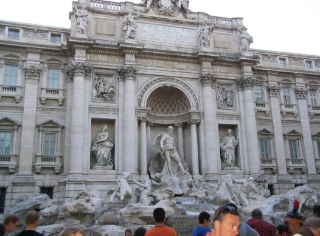}
  \end{subfigure}
  \hfill
  \begin{subfigure}{0.1364\linewidth}
    \caption*{Seg. w/ SfM}
    \includegraphics[width=\linewidth]{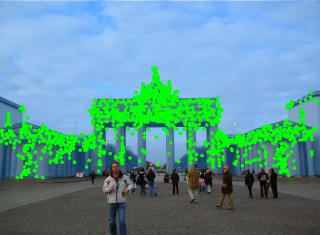}\\
    \includegraphics[width=\linewidth]{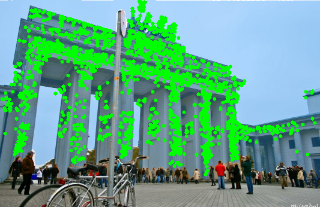}\\
    \includegraphics[width=\linewidth]{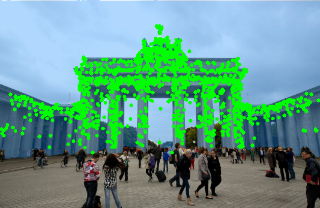}\\
    \includegraphics[width=\linewidth]{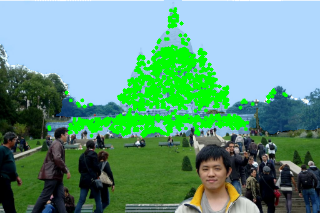}\\
    \includegraphics[width=\linewidth]{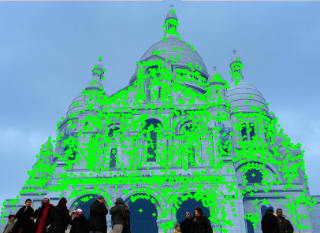}\\
    \includegraphics[width=\linewidth]{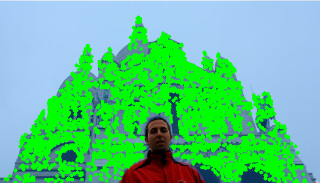}\\
    \includegraphics[width=\linewidth]{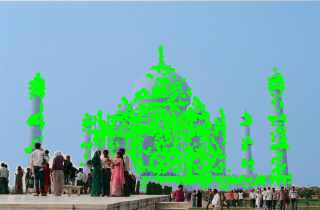}\\
    \includegraphics[width=\linewidth]{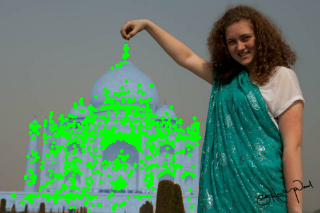}\\
    \includegraphics[width=\linewidth]{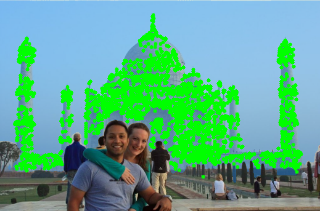}\\
    \includegraphics[width=\linewidth]{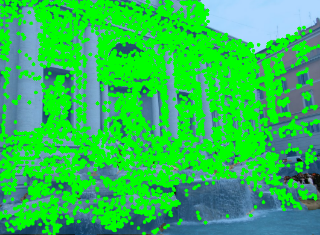}\\
    \includegraphics[width=\linewidth]{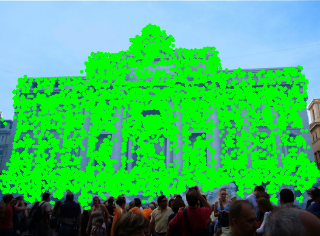}\\
    \includegraphics[width=\linewidth]{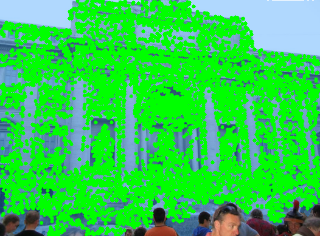}
  \end{subfigure}
  \hfill
  \begin{subfigure}{0.1364\linewidth}
    \caption*{$\mathcal{H}^{SfM}$}
    \includegraphics[width=\linewidth]{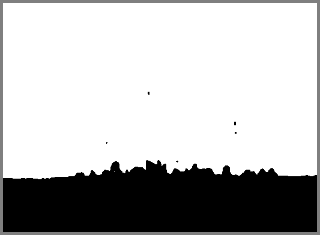}\\
    \includegraphics[width=\linewidth]{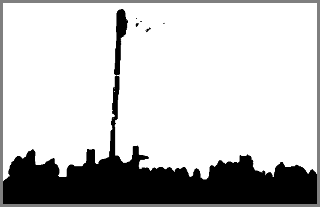}\\
    \includegraphics[width=\linewidth]{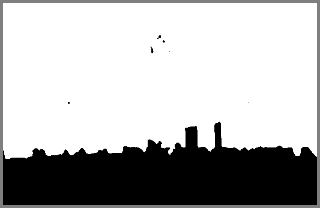}\\
    \includegraphics[width=\linewidth]{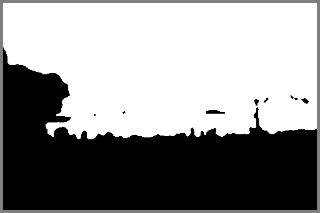}\\
    \includegraphics[width=\linewidth]{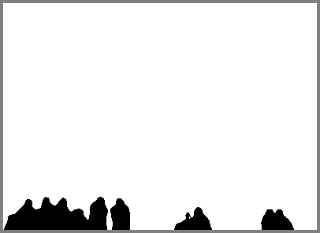}\\
    \includegraphics[width=\linewidth]{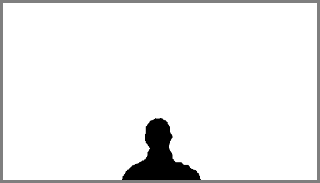}\\
    \includegraphics[width=\linewidth]{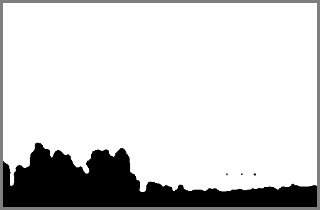}\\
    \includegraphics[width=\linewidth]{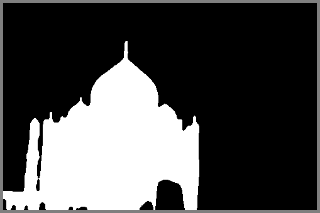}\\
    \includegraphics[width=\linewidth]{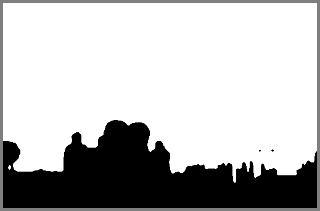}\\
    \includegraphics[width=\linewidth]{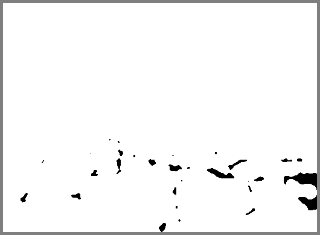}\\
    \includegraphics[width=\linewidth]{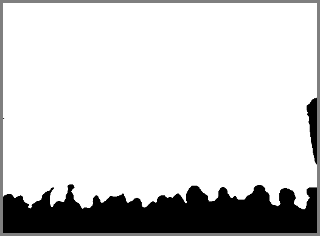}\\
    \includegraphics[width=\linewidth]{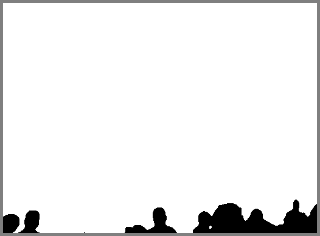}
  \end{subfigure}
  \hfill
  \begin{subfigure}{0.1364\linewidth}
    \caption*{Color Residual}
    \includegraphics[width=\linewidth]{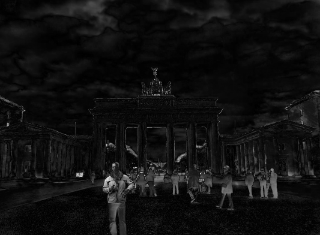}\\
    \includegraphics[width=\linewidth]{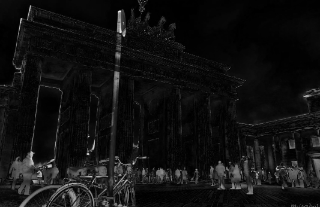}\\
    \includegraphics[width=\linewidth]{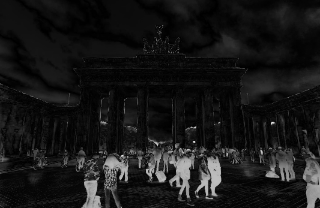}\\
    \includegraphics[width=\linewidth]{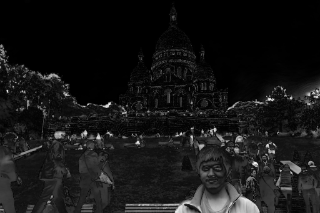}\\
    \includegraphics[width=\linewidth]{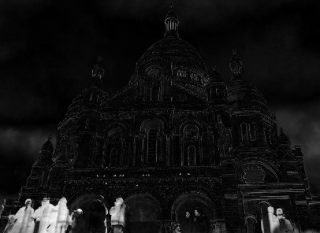}\\
    \includegraphics[width=\linewidth]{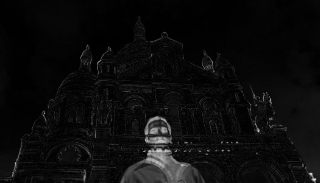}\\
    \includegraphics[width=\linewidth]{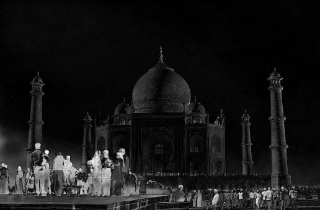}\\
    \includegraphics[width=\linewidth]{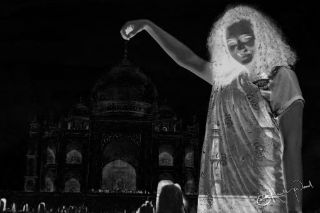}\\
    \includegraphics[width=\linewidth]{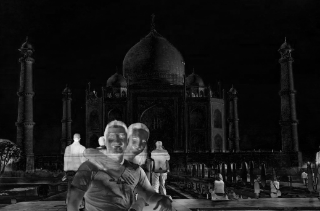}\\
    \includegraphics[width=\linewidth]{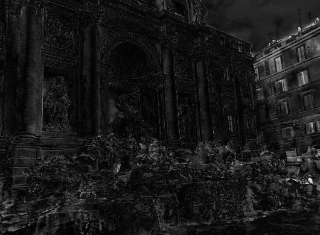}\\
    \includegraphics[width=\linewidth]{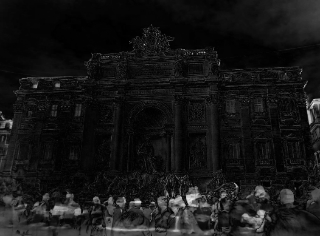}\\
    \includegraphics[width=\linewidth]{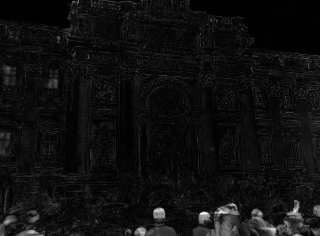}
  \end{subfigure}
  \hfill
  \begin{subfigure}{0.1364\linewidth}
    \caption*{$\mathcal{H}^{CR}$}
    \includegraphics[width=\linewidth]{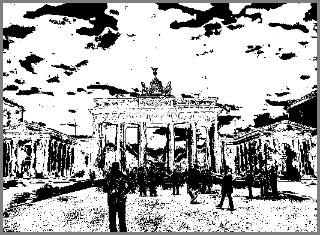}\\
    \includegraphics[width=\linewidth]{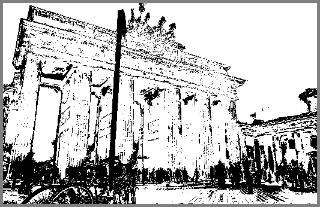}\\
    \includegraphics[width=\linewidth]{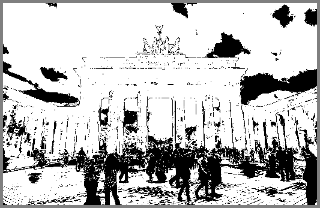}\\
    \includegraphics[width=\linewidth]{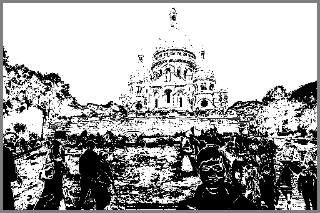}\\
    \includegraphics[width=\linewidth]{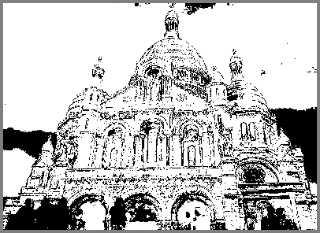}\\
    \includegraphics[width=\linewidth]{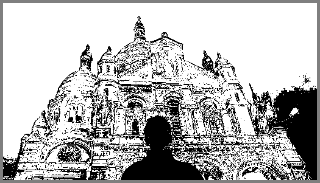}\\
    \includegraphics[width=\linewidth]{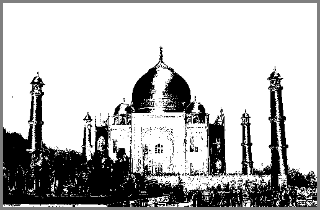}\\
    \includegraphics[width=\linewidth]{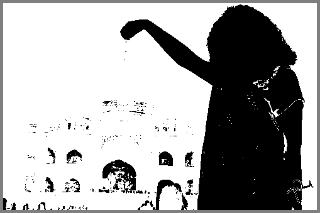}\\
    \includegraphics[width=\linewidth]{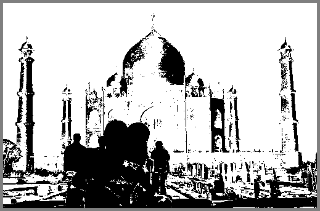}\\
    \includegraphics[width=\linewidth]{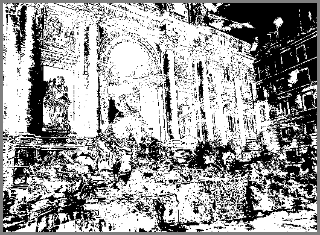}\\
    \includegraphics[width=\linewidth]{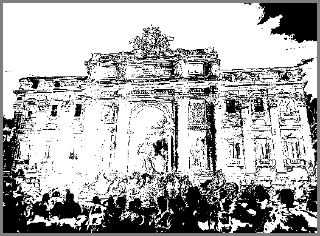}\\
    \includegraphics[width=\linewidth]{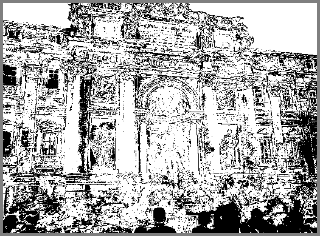}
  \end{subfigure}
  \hfill
  \begin{subfigure}{0.1364\linewidth}
    \caption*{Combined $\mathcal{H}$}
    \includegraphics[width=\linewidth]{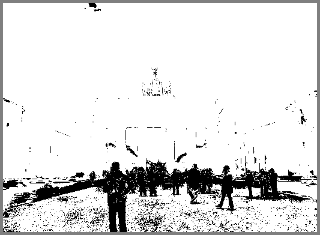}\\
    \includegraphics[width=\linewidth]{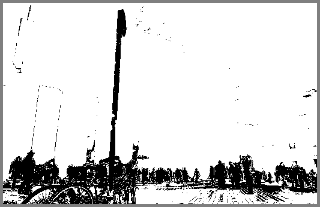}\\
    \includegraphics[width=\linewidth]{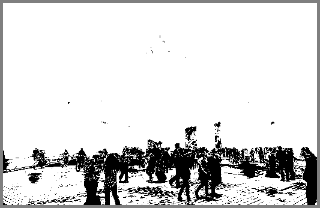}\\
    \includegraphics[width=\linewidth]{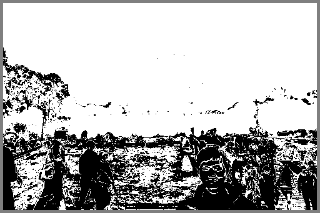}\\
    \includegraphics[width=\linewidth]{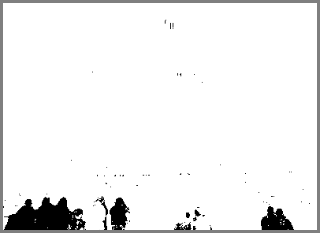}\\
    \includegraphics[width=\linewidth]{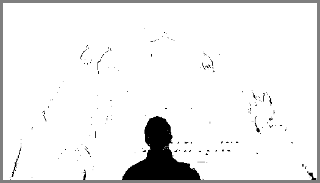}\\
    \includegraphics[width=\linewidth]{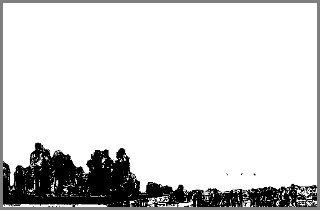}\\
    \includegraphics[width=\linewidth]{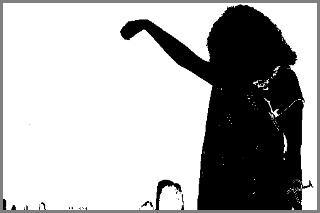}\\
    \includegraphics[width=\linewidth]{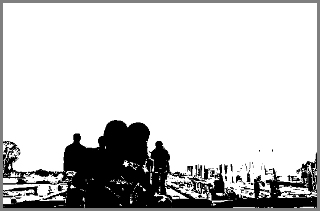}\\
    \includegraphics[width=\linewidth]{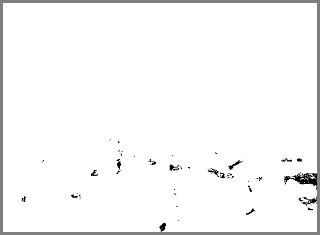}\\
    \includegraphics[width=\linewidth]{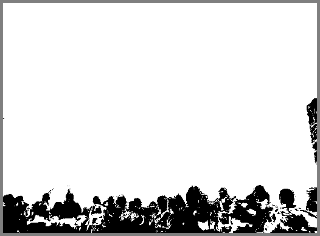}\\
    \includegraphics[width=\linewidth]{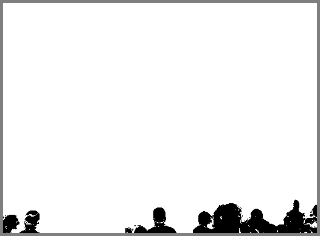}
  \end{subfigure}
  \hfill
  \begin{subfigure}{0.1364\linewidth}
    \caption*{Static Map}
    \includegraphics[width=\linewidth]{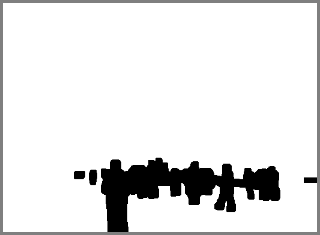}\\
    \includegraphics[width=\linewidth]{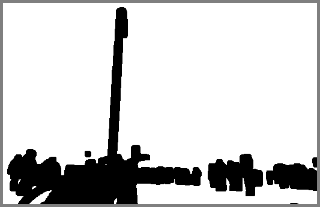}\\
    \includegraphics[width=\linewidth]{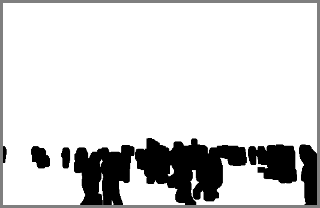}\\
    \includegraphics[width=\linewidth]{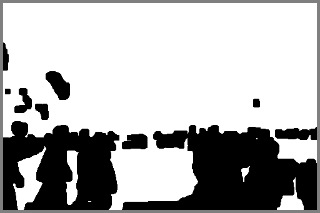}\\
    \includegraphics[width=\linewidth]{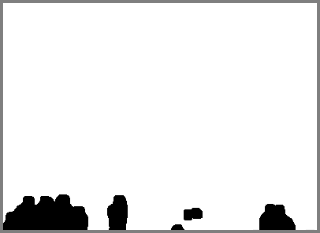}\\
    \includegraphics[width=\linewidth]{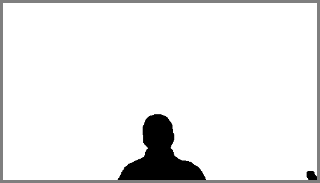}\\
    \includegraphics[width=\linewidth]{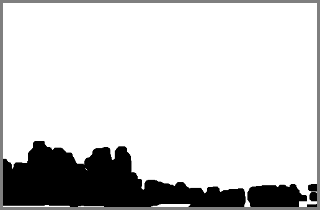}\\
    \includegraphics[width=\linewidth]{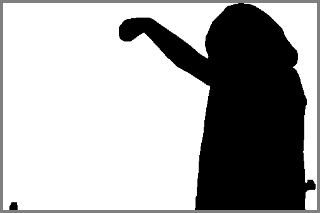}\\
    \includegraphics[width=\linewidth]{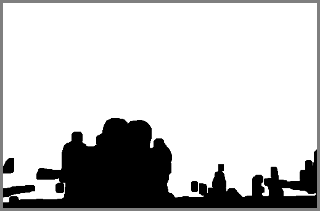}\\
    \includegraphics[width=\linewidth]{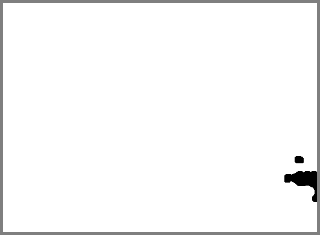}\\
    \includegraphics[width=\linewidth]{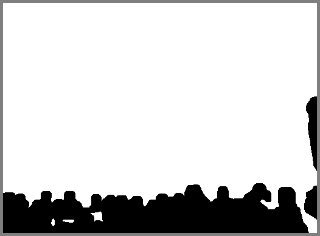}\\
    \includegraphics[width=\linewidth]{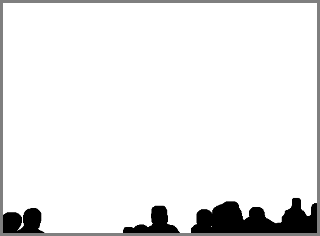}
  \end{subfigure}
  \caption{\textbf{Visualization of HuGS on the Phototourism dataset. }
  \vspace{-4mm}
  }
  \label{fig:supplement_visual_segment_phototourism}
\end{figure*}

\vspace{1mm}
\noindent \textbf{Visualization of HuGS.}
We additionally show the results of our HuGS, including intermediate results and final static maps.
\begin{itemize}
    \item {\it Kubric dataset} (\cref{fig:supplement_visual_segment_kubric}).
    Although there are no SfM features remaining on the shadows of moving objects after filtering, other static features on the ground still guide SAM to segment the entire ground, causing the SfM-based heuristics $\mathcal{H}^{SfM}$ to incorrectly include shadows. 
    As discussed in \cref{sec:heuristics_development} of the main paper and \cref{sec:sensitivity_hyperparameter}, the presence of $\mathcal{T}_{CR}$ can be additional insurance to remove shadows from combined $\mathcal{H}$ and the final static map.
    \item {\it Distractor dataset} (\cref{fig:supplement_visual_segment_distractor}).
    Consistent with \cref{sec:heuristics_development,fig:combine_example} of the main paper, the SfM-based heuristics $\mathcal{H}^{SfM}$ alone may miss static objects that only have smooth textures (\eg, white chairs of the Statue scene), while the residual-based heuristics $\mathcal{H}^{CR}$ alone may struggle with high-frequency static details (\eg, box textures of the Statue scene).
    Their combination $\mathcal{H}$ can roughly identify transient objects from static scenes, and final static maps are generated by SAM to provide accurate static \vs transient separation results.
    \item {\it Phototourism dataset} (\cref{fig:supplement_visual_segment_phototourism}).
    The results and conclusions are similar to those on the Distractor dataset. 
\end{itemize}

\begin{figure*}
  \centering
  \rotatebox{90}{\scriptsize ~Ours (Mip-NeRF 360)~~~~~~~~Ours (Nerfacto)~~~~~~~~~~~~~~~~RobustNeRF~~~~~~~~~~~~~~~~~~~~~D$^2$NeRF~~~~~~~~~~~~~~~~~~~~~~~~HA-NeRF~~~~~~~~~~~~~~~~~~~~~~~NeRF-W~~~~~~~~~~~~~~~~~~~~Mip-NeRF 360~~~~~~~~~~~~~~~~~~Nerfacto~~~~~~~~~~~~~~~~~~~~~~~Test Image}
  \begin{subfigure}{0.18\linewidth}
    \caption*{Car}
    \includegraphics[width=\linewidth]{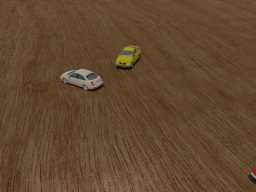}\\
    \includegraphics[width=\linewidth]{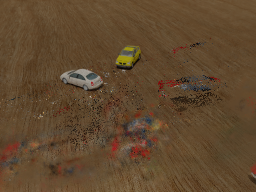}\\
    \includegraphics[width=\linewidth]{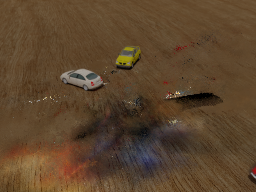}\\
    \includegraphics[width=\linewidth]{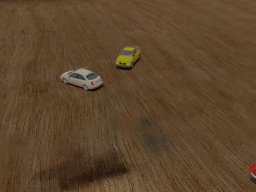}\\
    \includegraphics[width=\linewidth]{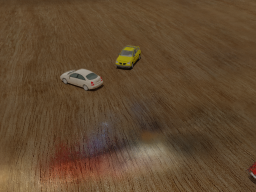}\\
    \includegraphics[width=\linewidth]{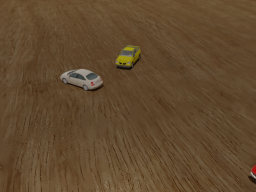}\\
    \includegraphics[width=\linewidth]{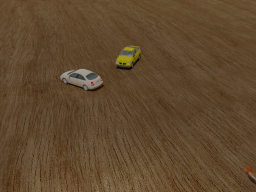}\\
    \includegraphics[width=\linewidth]{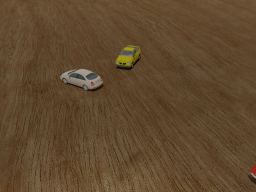}\\
    \includegraphics[width=\linewidth]{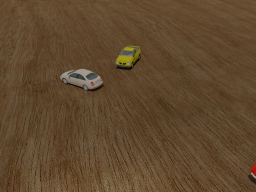}
  \end{subfigure}
  \begin{subfigure}{0.18\linewidth}
    \caption*{Cars}
    \includegraphics[width=\linewidth]{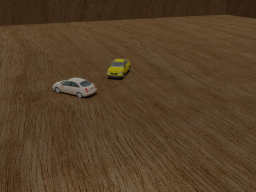}\\
    \includegraphics[width=\linewidth]{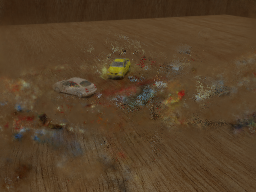}\\
    \includegraphics[width=\linewidth]{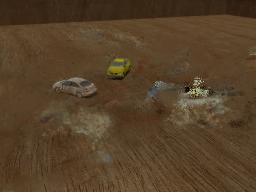}\\
    \includegraphics[width=\linewidth]{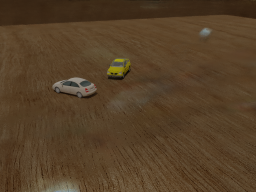}\\
    \includegraphics[width=\linewidth]{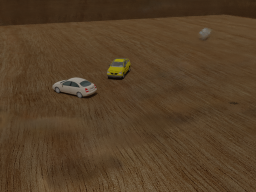}\\
    \includegraphics[width=\linewidth]{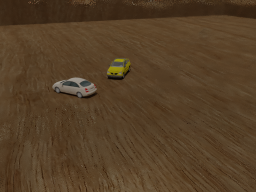}\\
    \includegraphics[width=\linewidth]{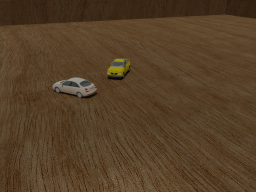}\\
    \includegraphics[width=\linewidth]{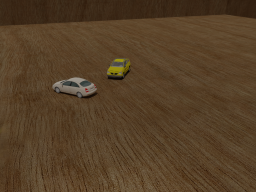}\\
    \includegraphics[width=\linewidth]{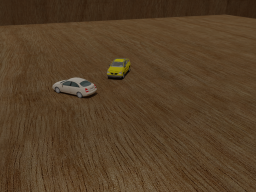}
  \end{subfigure}
  \begin{subfigure}{0.18\linewidth}
    \caption*{Bag}
    \includegraphics[width=\linewidth]{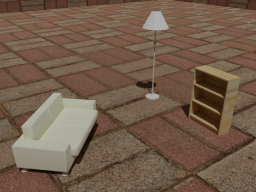}\\
    \includegraphics[width=\linewidth]{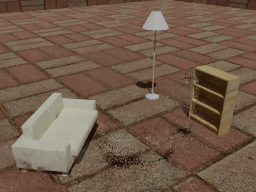}\\
    \includegraphics[width=\linewidth]{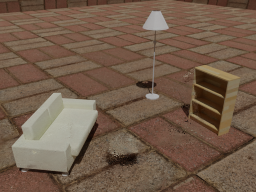}\\
    \includegraphics[width=\linewidth]{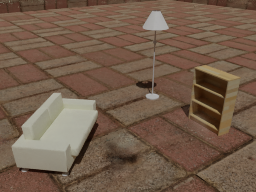}\\
    \includegraphics[width=\linewidth]{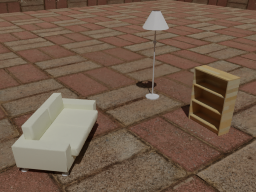}\\
    \includegraphics[width=\linewidth]{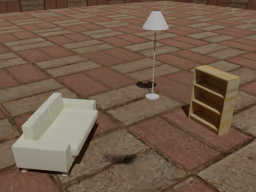}\\
    \includegraphics[width=\linewidth]{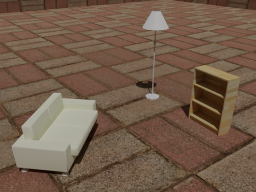}\\
    \includegraphics[width=\linewidth]{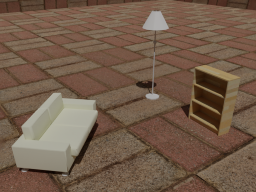}\\
    \includegraphics[width=\linewidth]{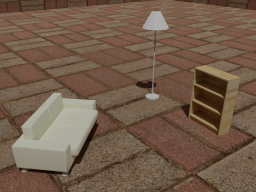}
  \end{subfigure}
  \begin{subfigure}{0.18\linewidth}
    \caption*{Chairs}
    \includegraphics[width=\linewidth]{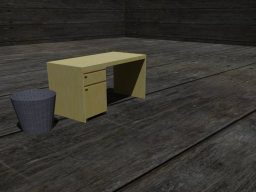}\\
    \includegraphics[width=\linewidth]{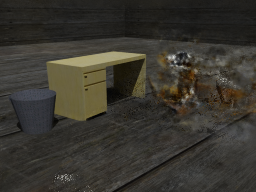}\\
    \includegraphics[width=\linewidth]{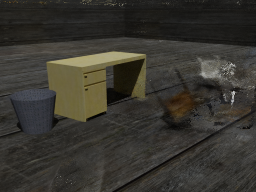}\\
    \includegraphics[width=\linewidth]{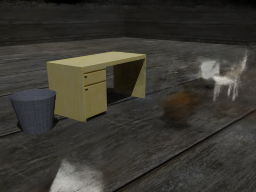}\\
    \includegraphics[width=\linewidth]{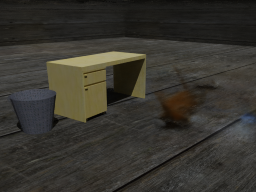}\\
    \includegraphics[width=\linewidth]{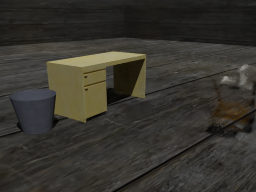}\\
    \includegraphics[width=\linewidth]{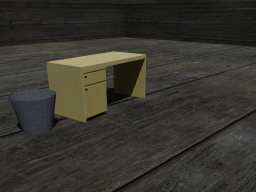}\\
    \includegraphics[width=\linewidth]{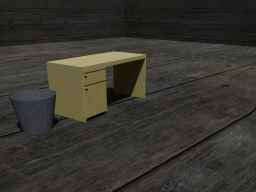}\\
    \includegraphics[width=\linewidth]{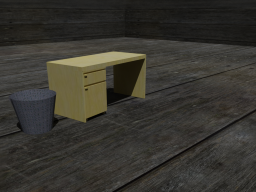}
  \end{subfigure}
  \begin{subfigure}{0.18\linewidth}
    \caption*{Pillow}
    \includegraphics[width=\linewidth]{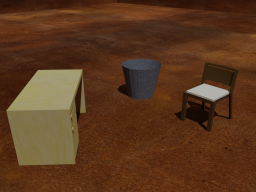}\\
    \includegraphics[width=\linewidth]{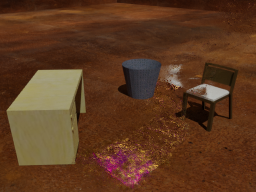}\\
    \includegraphics[width=\linewidth]{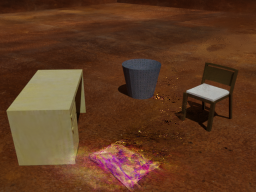}\\
    \includegraphics[width=\linewidth]{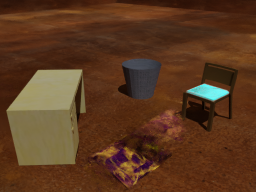}\\
    \includegraphics[width=\linewidth]{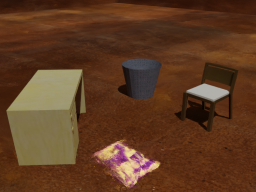}\\
    \includegraphics[width=\linewidth]{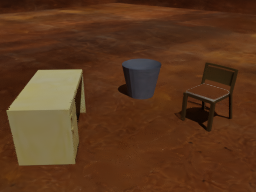}\\
    \includegraphics[width=\linewidth]{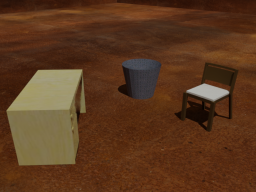}\\
    \includegraphics[width=\linewidth]{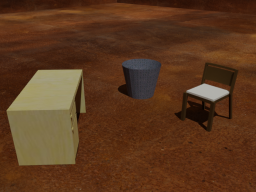}\\
    \includegraphics[width=\linewidth]{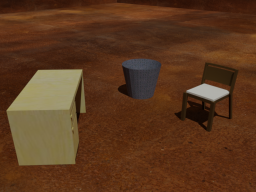}
  \end{subfigure}
  \caption{\textbf{Qualitative results of view synthesis on the Kubric dataset.} 
  Our method can eliminate artifacts from the native baseline models (Nerfacto and Mip-NeRF 360) and preserve more static details (\eg, the ground textures) than those of existing methods.
  \vspace{-4mm}
  }
  \label{fig:supplement_visual_render_kubric}
\end{figure*}

\begin{figure*}
  \centering
  \rotatebox{90}{\scriptsize ~~~Ours (Mip-NeRF 360)~~~~~~~~~~~~~Ours (Nerfacto)~~~~~~~~~~~~~~~~~~~~RobustNeRF~~~~~~~~~~~~~~~~~~~~~~~~~HA-NeRF~~~~~~~~~~~~~~~~~~~~~~~~~~~NeRF-W~~~~~~~~~~~~~~~~~~~~~~~~Mip-NeRF 360~~~~~~~~~~~~~~~~~~~~~~~Nerfacto~~~~~~~~~~~~~~~~~~~~~~~~~~~~Test Image}
  \begin{subfigure}{0.2\linewidth}
    \caption*{Statue}
    \includegraphics[width=\linewidth]{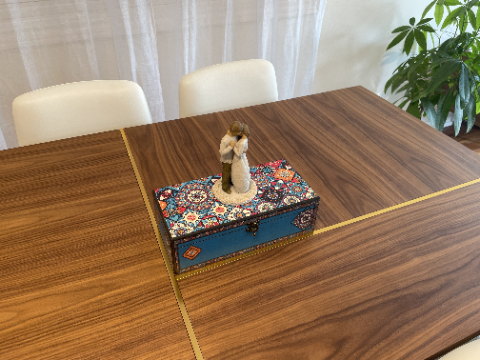}\\
    \includegraphics[width=\linewidth]{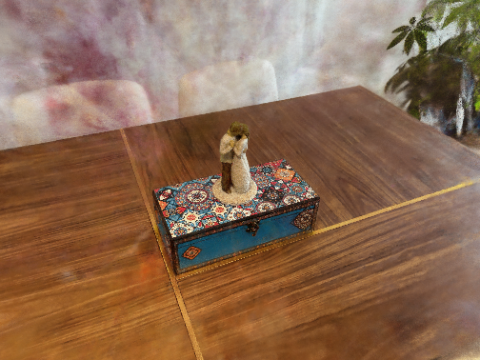}\\
    \includegraphics[width=\linewidth]{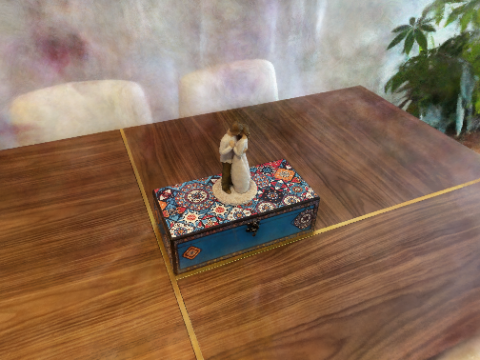}\\
    \includegraphics[width=\linewidth]{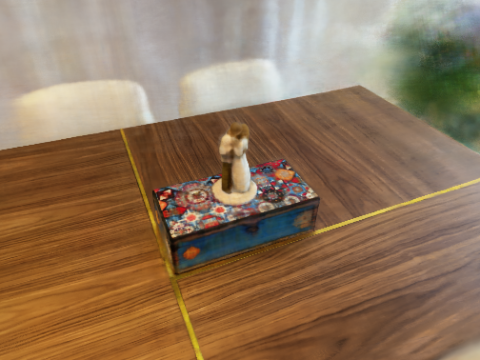}\\
    \includegraphics[width=\linewidth]{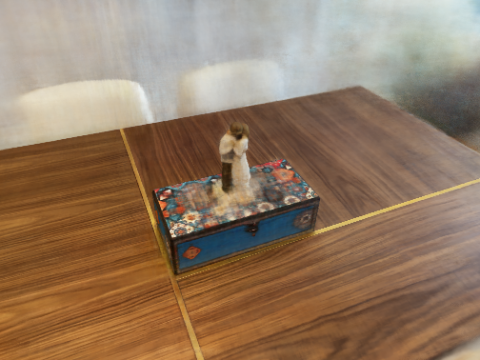}\\
    \includegraphics[width=\linewidth]{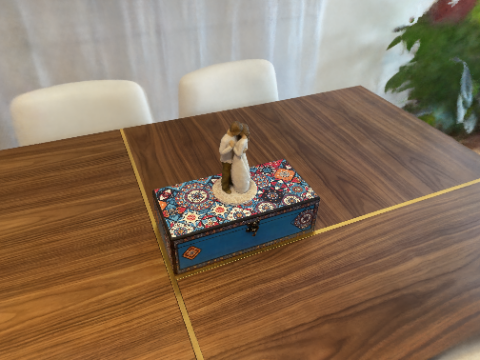}\\
    \includegraphics[width=\linewidth]{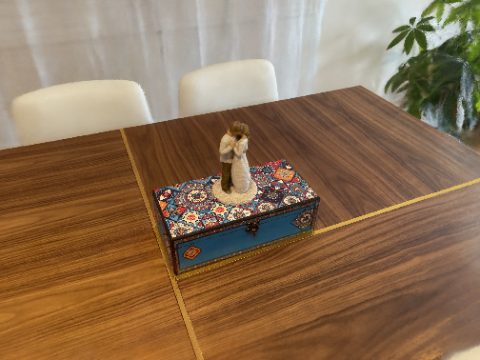}\\
    \includegraphics[width=\linewidth]{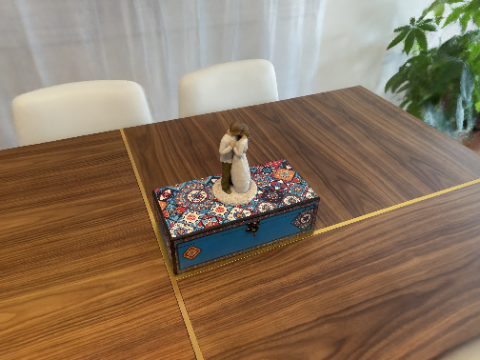}
  \end{subfigure}
  \begin{subfigure}{0.2\linewidth}
    \caption*{Android}
    \includegraphics[width=\linewidth]{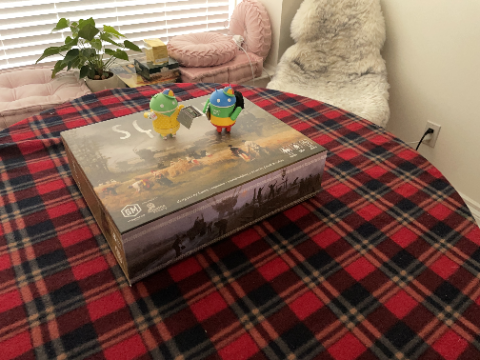}\\
    \includegraphics[width=\linewidth]{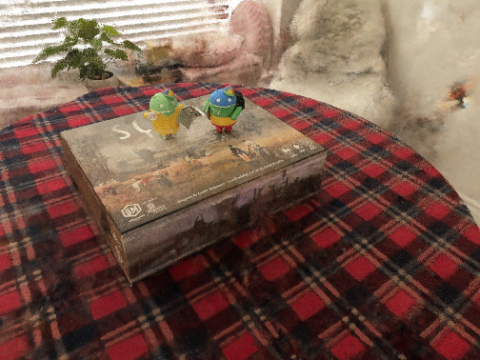}\\
    \includegraphics[width=\linewidth]{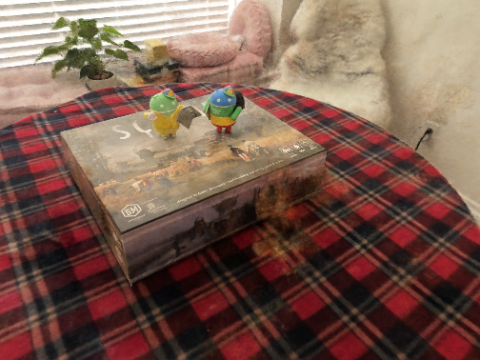}\\
    \includegraphics[width=\linewidth]{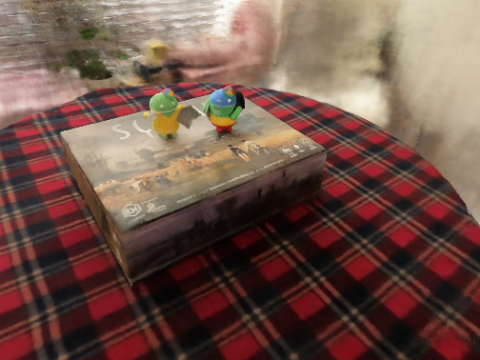}\\
    \includegraphics[width=\linewidth]{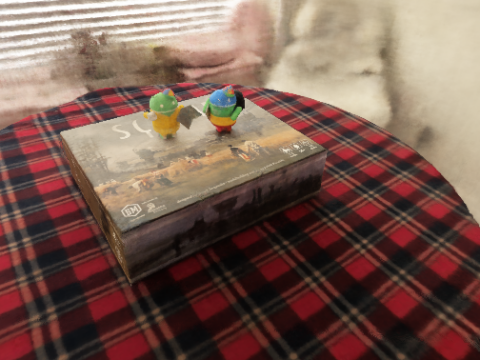}\\
    \includegraphics[width=\linewidth]{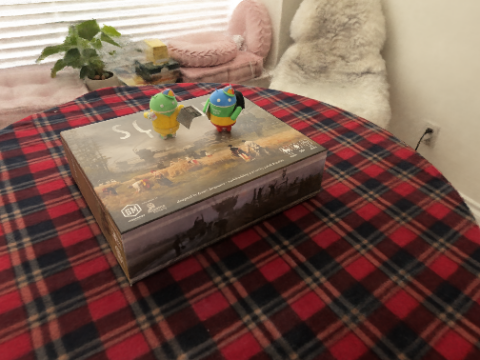}\\
    \includegraphics[width=\linewidth]{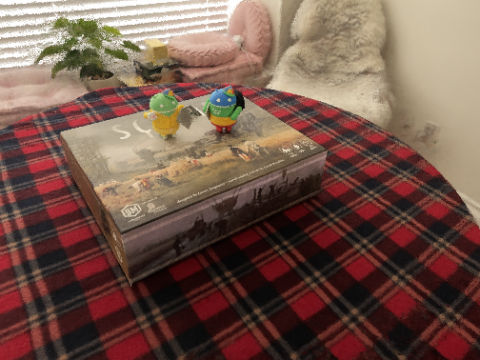}\\
    \includegraphics[width=\linewidth]{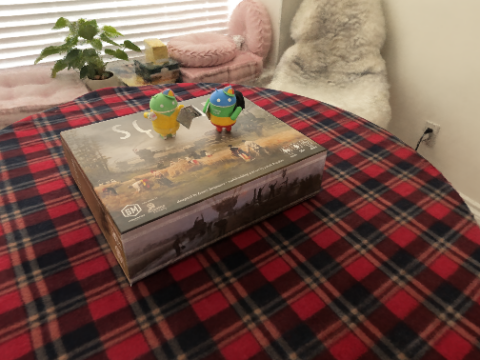}
  \end{subfigure}
  \begin{subfigure}{0.2\linewidth}
    \caption*{Crab}
    \includegraphics[width=\linewidth]{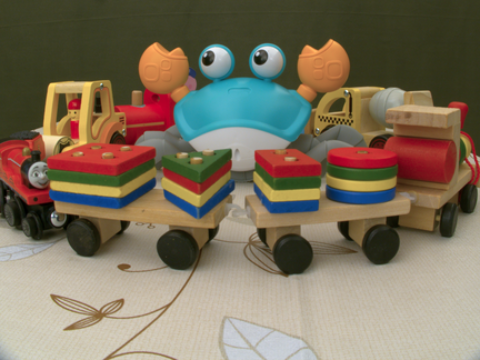}\\
    \includegraphics[width=\linewidth]{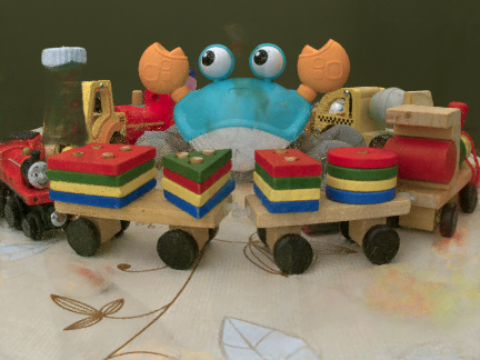}\\
    \includegraphics[width=\linewidth]{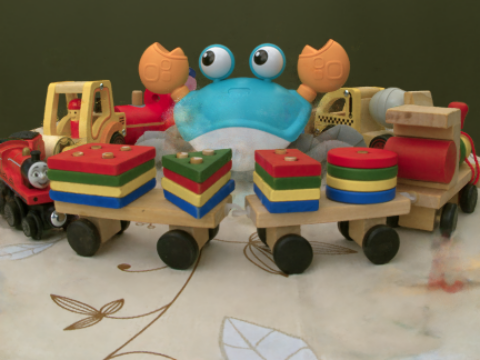}\\
    \includegraphics[width=\linewidth]{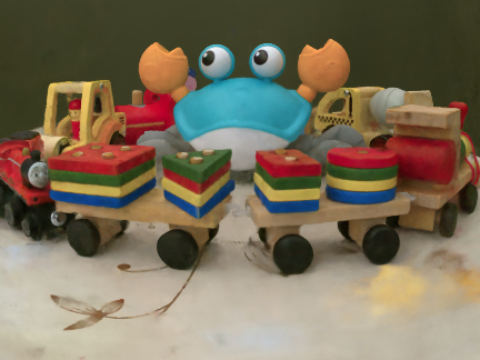}\\
    \includegraphics[width=\linewidth]{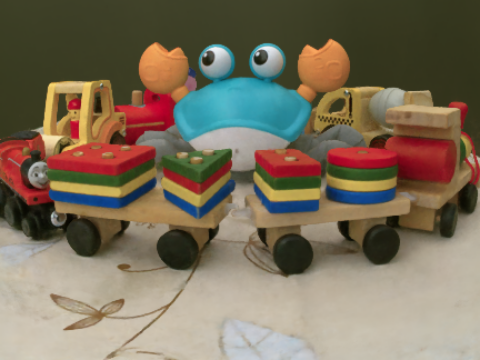}\\
    \includegraphics[width=\linewidth]{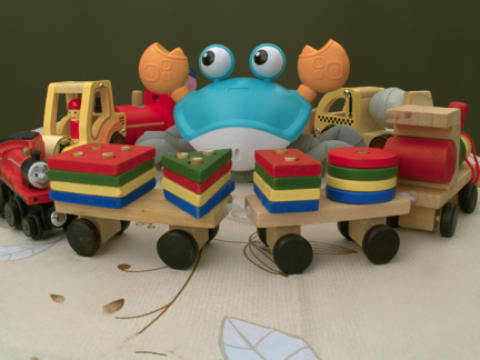}\\
    \includegraphics[width=\linewidth]{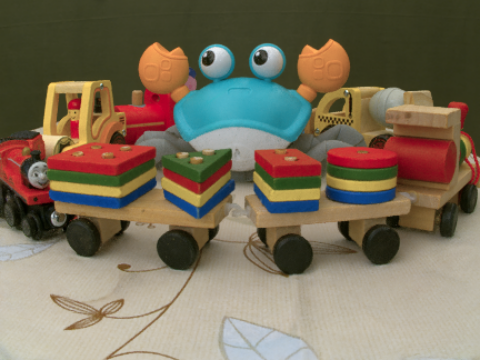}\\
    \includegraphics[width=\linewidth]{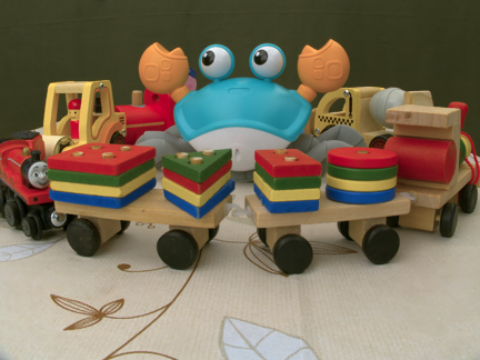}
  \end{subfigure}
  \begin{subfigure}{0.2\linewidth}
    \caption*{BabyYoda}
    \includegraphics[width=\linewidth]{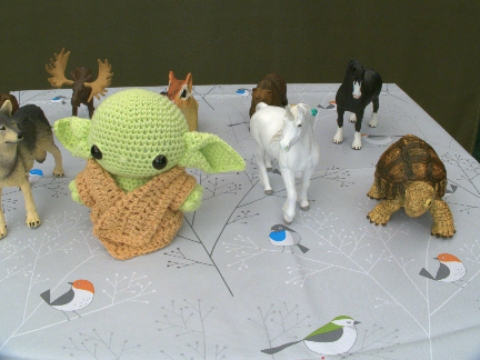}\\
    \includegraphics[width=\linewidth]{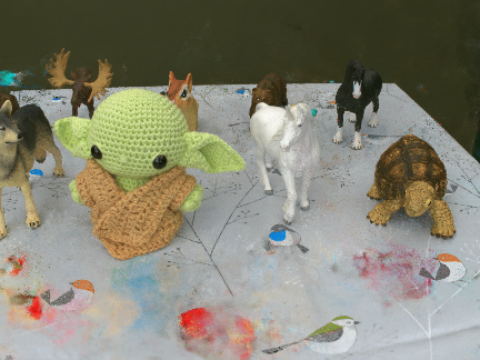}\\
    \includegraphics[width=\linewidth]{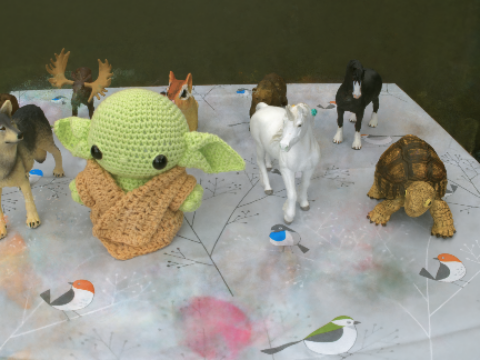}\\
    \includegraphics[width=\linewidth]{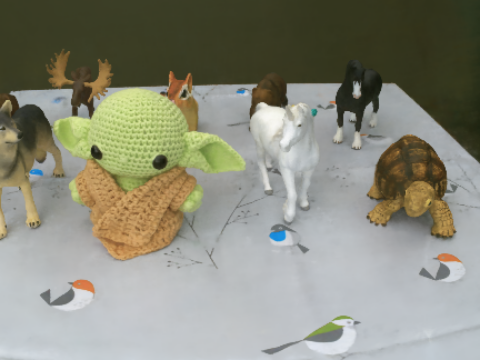}\\
    \includegraphics[width=\linewidth]{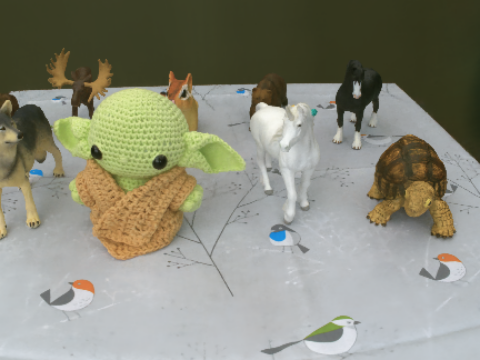}\\
    \includegraphics[width=\linewidth]{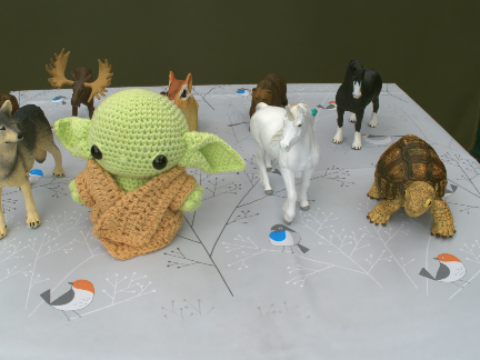}\\
    \includegraphics[width=\linewidth]{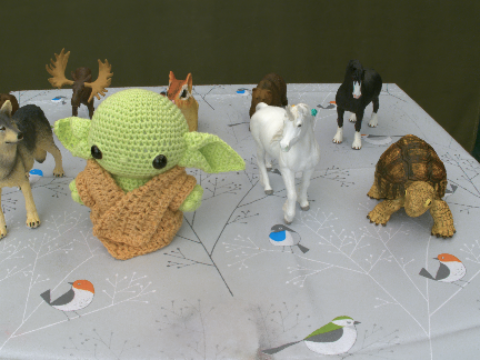}\\
    \includegraphics[width=\linewidth]{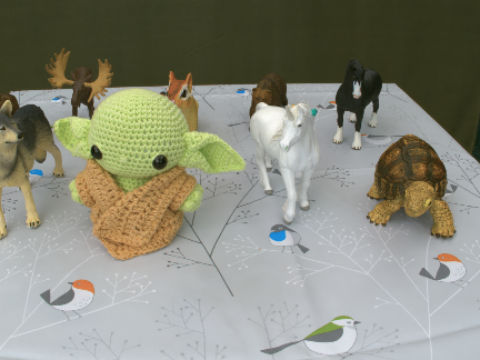}
  \end{subfigure}
  \caption{\textbf{Qualitative results of view synthesis on the Distractor dataset. } Our method can better preserve static details (\eg, plants of the Statue and Android scene, table-cloth textures of the Crab and BabyYoda scene) while ignoring transient distractors. 
  \vspace{-4mm}
  }
  \label{fig:supplement_visual_render_distractor}
\end{figure*}

\begin{figure*}
  \centering
  \rotatebox{90}{\scriptsize ~~~Ours (Mip-NeRF 360)~~~~~~~~~~~~~Ours (Nerfacto)~~~~~~~~~~~~~~~~~~~~RobustNeRF~~~~~~~~~~~~~~~~~~~~~~~~~HA-NeRF~~~~~~~~~~~~~~~~~~~~~~~~~~~NeRF-W~~~~~~~~~~~~~~~~~~~~~~~Mip-NeRF 360~~~~~~~~~~~~~~~~~~~~~~~~Nerfacto~~~~~~~~~~~~~~~~~~~~~~~~~~~~Test Image}
  \begin{subfigure}{0.2264\linewidth}
    \caption*{Brandenburg Gate}
    \includegraphics[width=\linewidth]{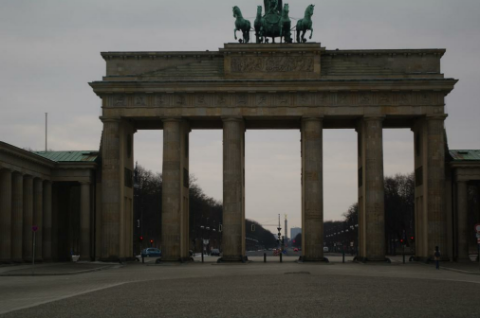}\\
    \includegraphics[width=\linewidth]{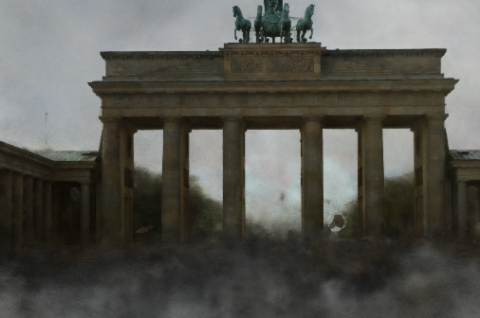}\\
    \includegraphics[width=\linewidth]{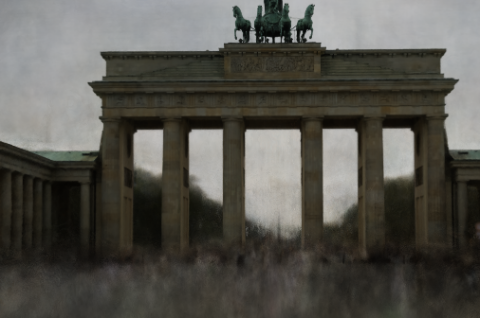}\\
    \includegraphics[width=\linewidth]{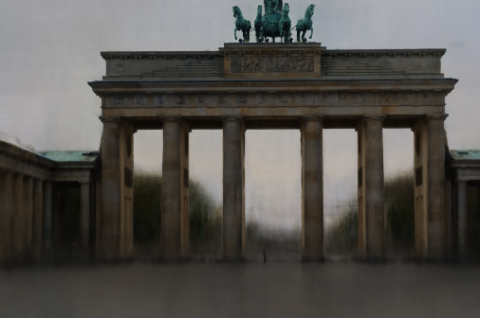}\\
    \includegraphics[width=\linewidth]{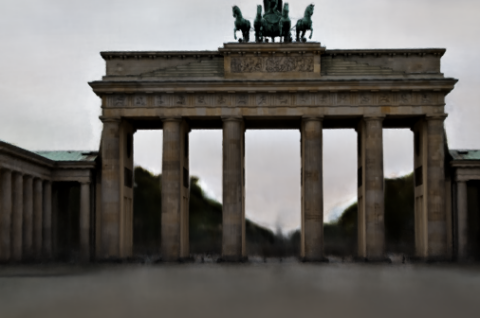}\\
    \includegraphics[width=\linewidth]{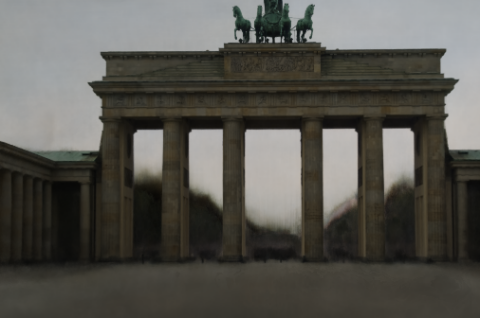}\\
    \includegraphics[width=\linewidth]{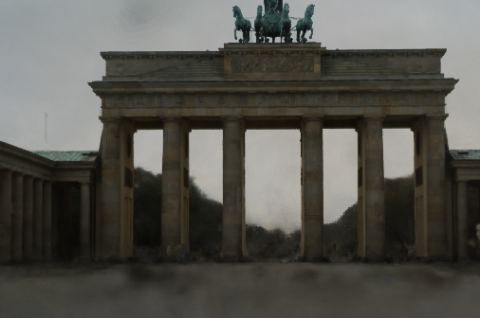}\\
    \includegraphics[width=\linewidth]{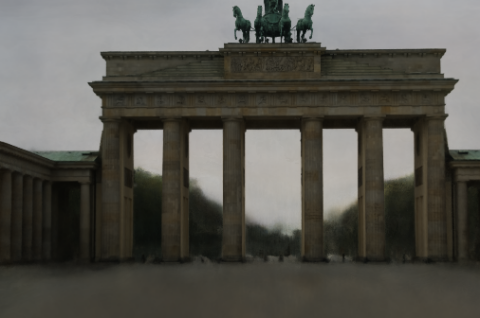}
  \end{subfigure}
  \begin{subfigure}{0.2006\linewidth}
    \caption*{Sacre Coeur}
    \includegraphics[width=\linewidth]{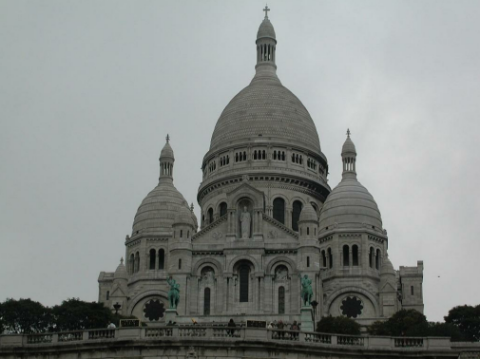}\\
    \includegraphics[width=\linewidth]{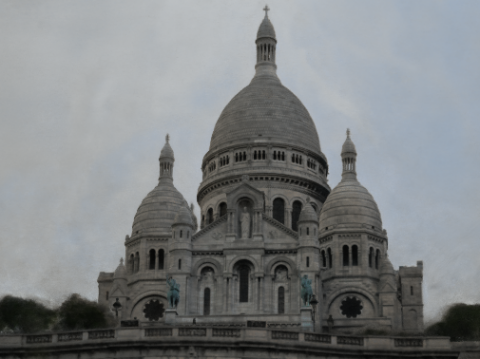}\\
    \includegraphics[width=\linewidth]{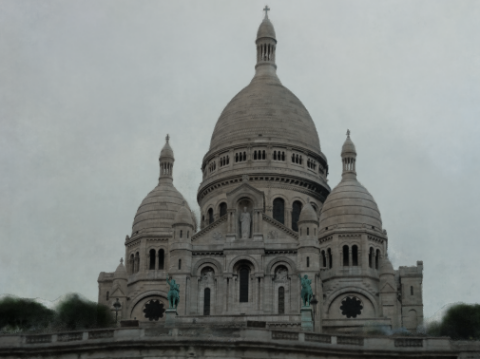}\\
    \includegraphics[width=\linewidth]{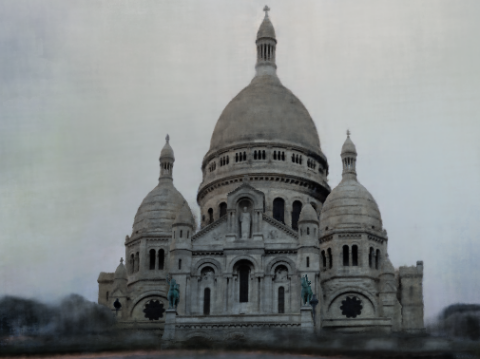}\\
    \includegraphics[width=\linewidth]{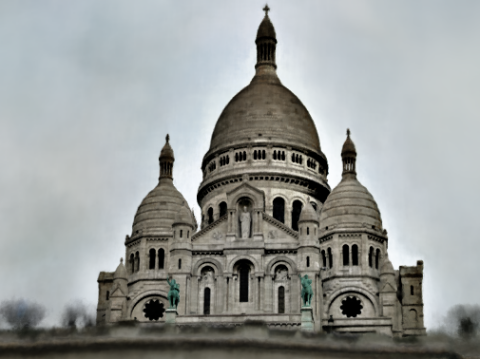}\\
    \includegraphics[width=\linewidth]{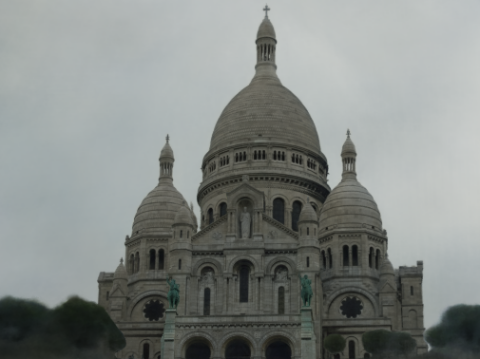}\\
    \includegraphics[width=\linewidth]{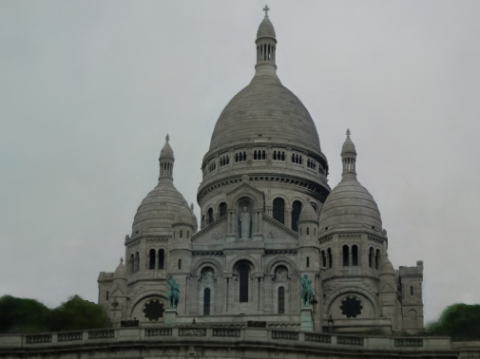}\\
    \includegraphics[width=\linewidth]{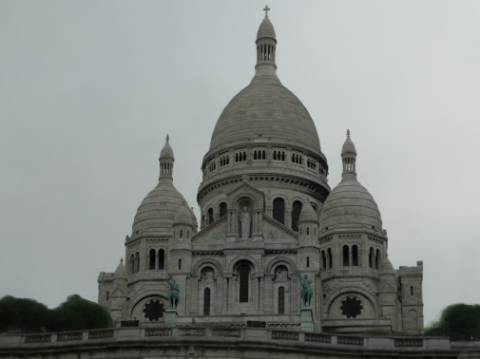}
  \end{subfigure}
  \begin{subfigure}{0.2271\linewidth}
    \caption*{Taj Mahal}
    \includegraphics[width=\linewidth]{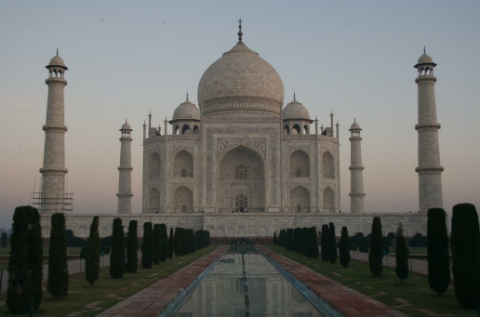}\\
    \includegraphics[width=\linewidth]{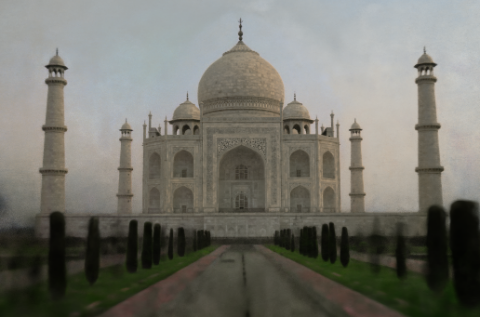}\\
    \includegraphics[width=\linewidth]{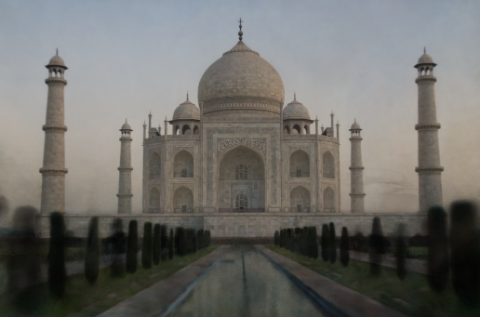}\\
    \includegraphics[width=\linewidth]{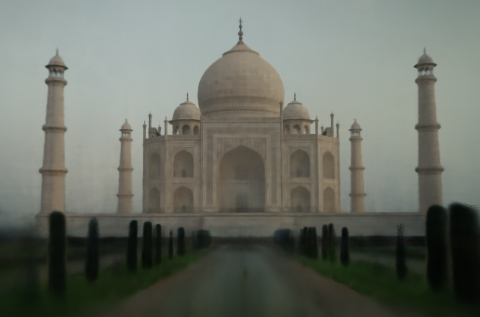}\\
    \includegraphics[width=\linewidth]{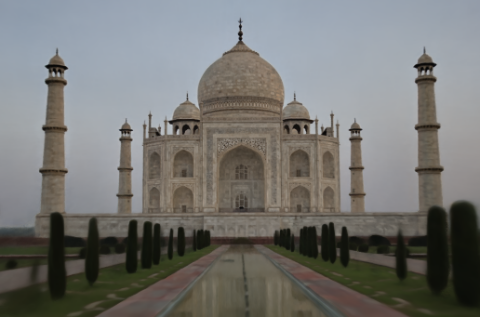}\\
    \includegraphics[width=\linewidth]{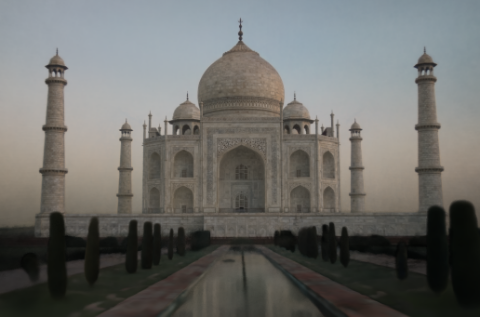}\\
    \includegraphics[width=\linewidth]{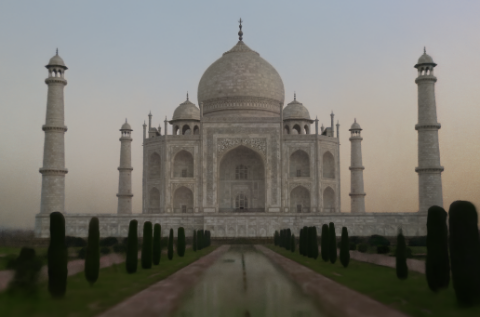}\\
    \includegraphics[width=\linewidth]{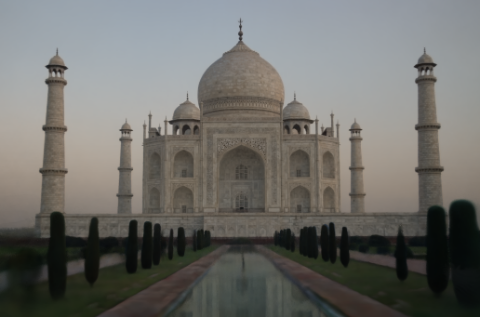}
  \end{subfigure}
  \begin{subfigure}{0.2308\linewidth}
    \caption*{Trevi Fountain}
    \includegraphics[width=\linewidth]{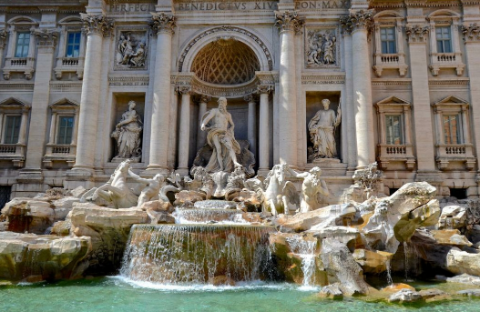}\\
    \includegraphics[width=\linewidth]{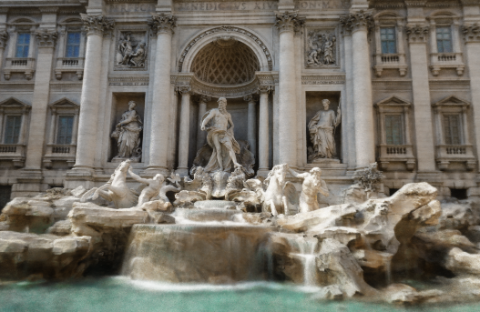}\\
    \includegraphics[width=\linewidth]{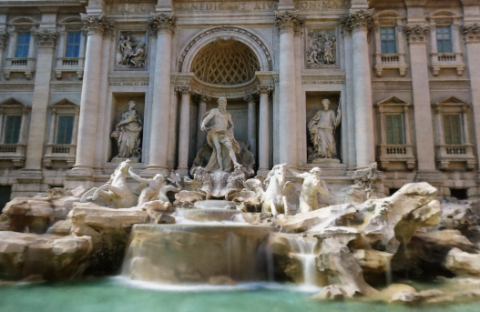}\\
    \includegraphics[width=\linewidth]{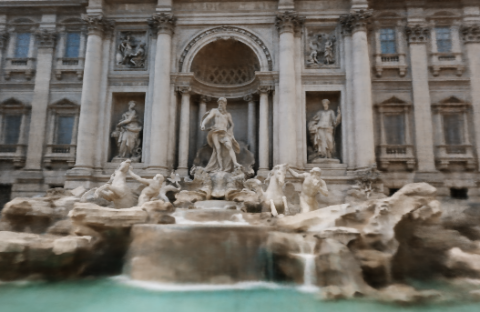}\\
    \includegraphics[width=\linewidth]{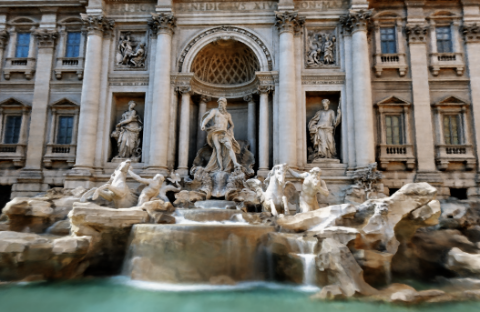}\\
    \includegraphics[width=\linewidth]{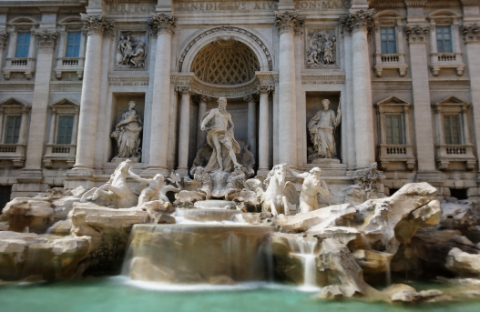}\\
    \includegraphics[width=\linewidth]{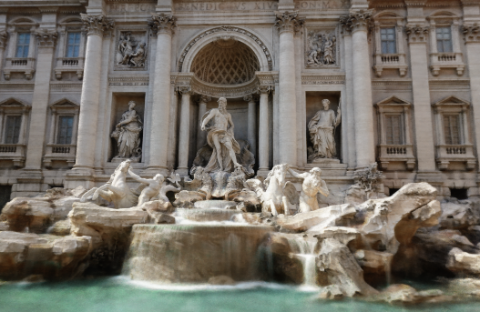}\\
    \includegraphics[width=\linewidth]{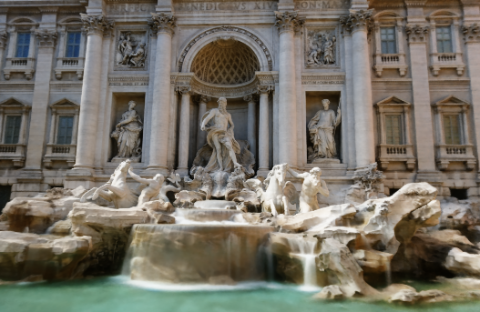}
  \end{subfigure}
  \caption{\textbf{Qualitative results of view synthesis on the Phototourism dataset. }
  The native baseline models (Nerfacto and Mip-NeRF 360) can generate acceptable results without any processing of transient distractors. Even so, our method achieves competitive results by eliminating artifacts (\eg, the bottom of the Brandenburg Gate) and recovering more static details (\eg, the fence of the Sacre Coeur).
  \vspace{-4mm}
  }
  \label{fig:supplement_visual_render_phototourism}
\end{figure*}

\vspace{1mm}
\noindent \textbf{Evaluation on View Synthesis.}
We show more qualitative results of view synthesis on different datasets.
\begin{itemize}
    \item {\it Kubric dataset} (\cref{fig:supplement_visual_render_kubric}). Existing baseline models, namely Nerfacto and Mip-NeRF 360, exhibit limitations in managing transient distractors, resulting in significant artifacts in their rendered images. Similarly, NeRF-W and HA-NeRF demonstrate inadequacies in segregating transient objects from static scenes, leading to artifact-ridden outputs. D$^2$NeRF shows improved accuracy in separating elements, yet some transient distractors persist within the static scenes. RobustNeRF achieves complete removal of transient distractors using a color residual threshold; however, this results in the inadvertent loss of certain static details. In contrast, our HuGS, applicable across various baseline models, effectively balances the exclusion of transient distractors and retention of static scene details, thereby enhancing the rendering quality.

\item {\it Distractor dataset} (\cref{fig:supplement_visual_render_distractor}).  The results and conclusions are similar to those on the Kubric dataset.

\item {\it Phototourism dataset} (\cref{fig:supplement_visual_render_phototourism}). As discussed in \cref{sec:supple_comparison_on_photourism}, transient distractor management is less critical in the test set evaluation of this dataset. Nevertheless, our approach demonstrates a notable reduction of artifacts (\eg, at the bottom of the Brandenburg Gate) and an enhanced recovery of architectural details (\eg, the fence of the Sacre Coeur), surpassing previous methodologies. This underscores the superiority of our method.
\end{itemize}

\begin{figure*}
  \centering
  \rotatebox{90}{\scriptsize ~~~~~HuGS (Ours)~~~~~~~~~~~RobustNeRF~~~~~~~~~~~~~D$^2$NeRF~~~~~~~~~~~~~~~~~HA-NeRF~~~~~~~~~~~~~~~~NeRF-W~~~~~~~~~~~~~~~~~~DINO~~~~~~~~~~~~~~Grounded-SAM~~~~~~~~Mask2Former~~~~~~~~~~DeepLabv3+~~~~~~~~~~~~~~~~~GT~~~~~~~~~~~~~~~Training Image}
  \begin{subfigure}{0.145\linewidth}
    \caption*{Car}
    \includegraphics[width=\linewidth]{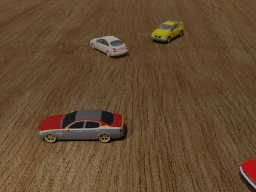}\\
    \includegraphics[width=\linewidth]{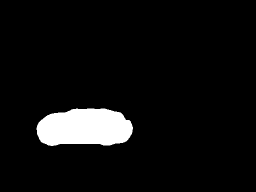}\\
    \includegraphics[width=\linewidth]{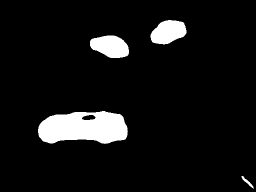}\\
    \includegraphics[width=\linewidth]{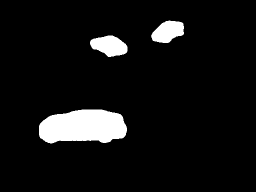}\\
    \includegraphics[width=\linewidth]{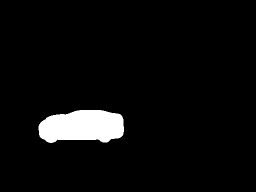}\\
    \includegraphics[width=\linewidth]{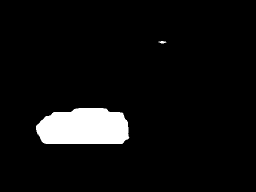}\\
    \includegraphics[width=\linewidth]{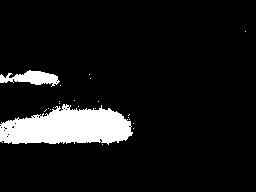}\\
    \includegraphics[width=\linewidth]{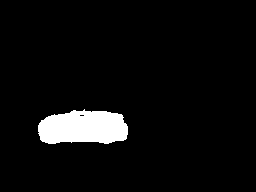}\\
    \includegraphics[width=\linewidth]{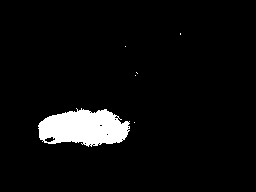}\\
    \includegraphics[width=\linewidth]{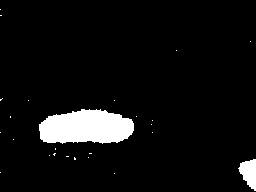}\\
    \includegraphics[width=\linewidth]{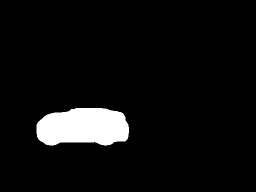}
  \end{subfigure}
  \begin{subfigure}{0.145\linewidth}
    \caption*{Cars}
    \includegraphics[width=\linewidth]{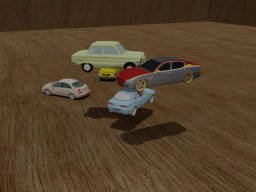}\\
    \includegraphics[width=\linewidth]{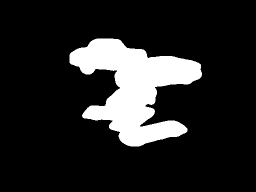}\\
    \includegraphics[width=\linewidth]{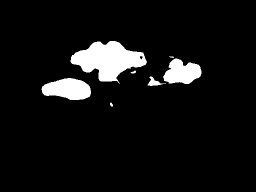}\\
    \includegraphics[width=\linewidth]{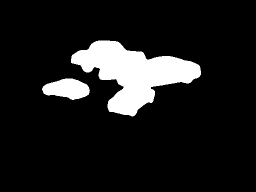}\\
    \includegraphics[width=\linewidth]{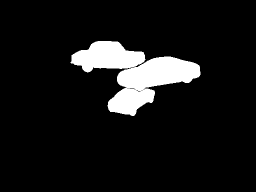}\\
    \includegraphics[width=\linewidth]{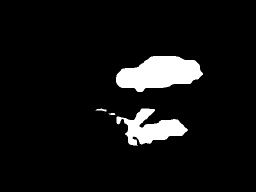}\\
    \includegraphics[width=\linewidth]{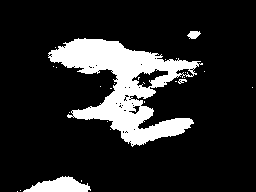}\\
    \includegraphics[width=\linewidth]{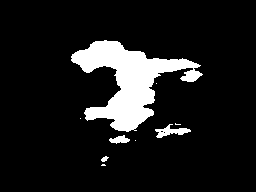}\\
    \includegraphics[width=\linewidth]{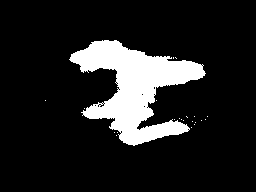}\\
    \includegraphics[width=\linewidth]{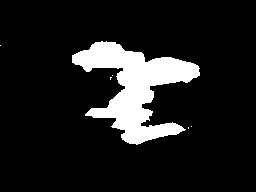}\\
    \includegraphics[width=\linewidth]{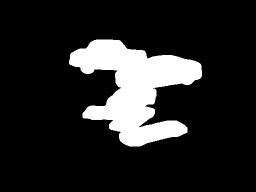}
  \end{subfigure}
  \begin{subfigure}{0.145\linewidth}
    \caption*{Bag}
    \includegraphics[width=\linewidth]{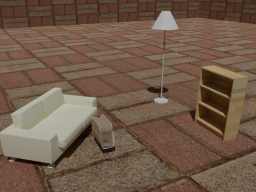}\\
    \includegraphics[width=\linewidth]{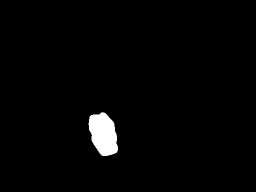}\\
    \includegraphics[width=\linewidth]{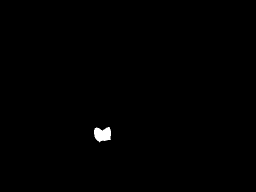}\\
    \includegraphics[width=\linewidth]{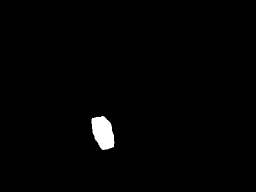}\\
    \includegraphics[width=\linewidth]{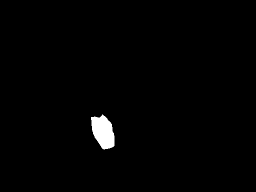}\\
    \includegraphics[width=\linewidth]{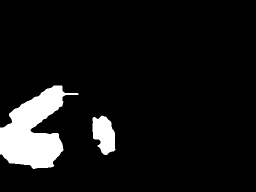}\\
    \includegraphics[width=\linewidth]{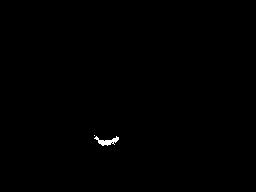}\\
    \includegraphics[width=\linewidth]{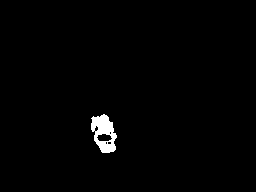}\\
    \includegraphics[width=\linewidth]{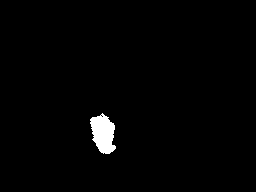}\\
    \includegraphics[width=\linewidth]{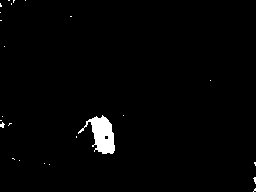}\\
    \includegraphics[width=\linewidth]{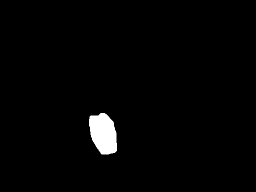}
  \end{subfigure}
  \begin{subfigure}{0.145\linewidth}
    \caption*{Chairs}
    \includegraphics[width=\linewidth]{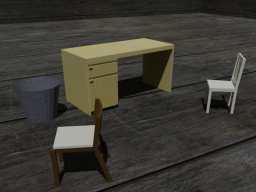}\\
    \includegraphics[width=\linewidth]{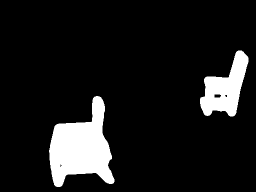}\\
    \includegraphics[width=\linewidth]{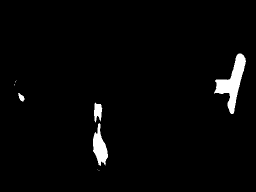}\\
    \includegraphics[width=\linewidth]{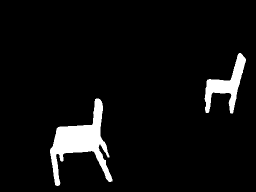}\\
    \includegraphics[width=\linewidth]{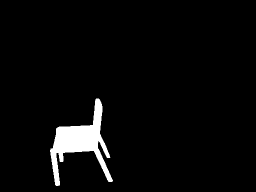}\\
    \includegraphics[width=\linewidth]{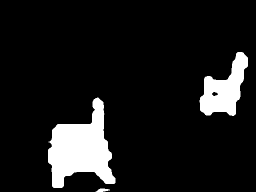}\\
    \includegraphics[width=\linewidth]{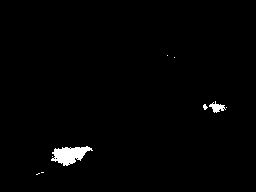}\\
    \includegraphics[width=\linewidth]{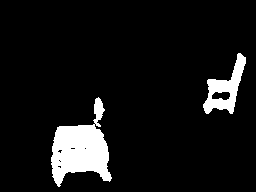}\\
    \includegraphics[width=\linewidth]{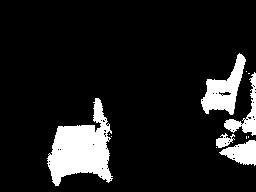}\\
    \includegraphics[width=\linewidth]{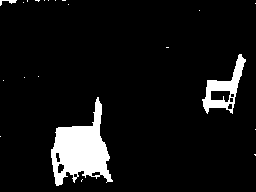}\\
    \includegraphics[width=\linewidth]{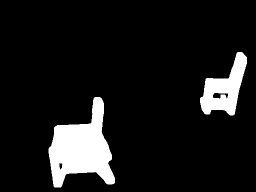}
  \end{subfigure}
  \begin{subfigure}{0.145\linewidth}
    \caption*{Pillow}
    \includegraphics[width=\linewidth]{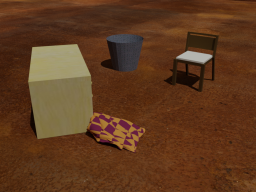}\\
    \includegraphics[width=\linewidth]{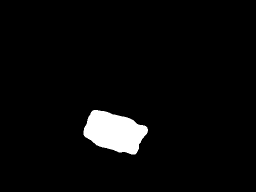}\\
    \includegraphics[width=\linewidth]{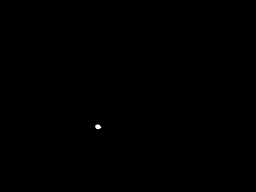}\\
    \includegraphics[width=\linewidth]{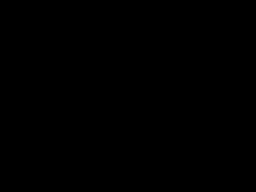}\\
    \includegraphics[width=\linewidth]{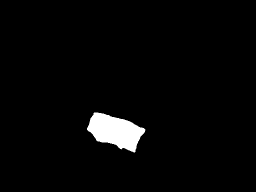}\\
    \includegraphics[width=\linewidth]{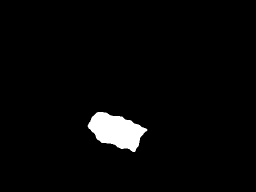}\\
    \includegraphics[width=\linewidth]{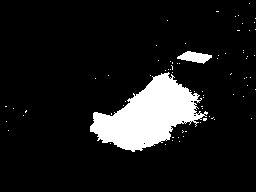}\\
    \includegraphics[width=\linewidth]{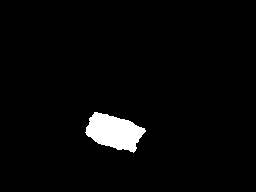}\\
    \includegraphics[width=\linewidth]{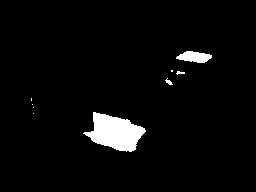}\\
    \includegraphics[width=\linewidth]{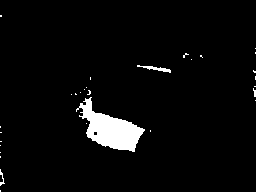}\\
    \includegraphics[width=\linewidth]{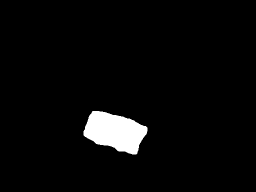}
  \end{subfigure}
  \caption{\textbf{Qualitative segmentation results on the Kubric dataset.} Compared to existing methods, our method can segment transient objects from static scenes more accurately. 
  \vspace{-4mm}
  }
  \label{fig:supplement_visual_seg_compare}
\end{figure*}

\vspace{1mm}
\noindent \textbf{Evaluation on Segmentation.}
Enhanced qualitative segmentation results are presented in~\cref{fig:supplement_visual_seg_compare}, paralleling the discussions in~\cref{sec:evaluation_segmentation} of the main paper. Existing segmentation methodologies that leverage prior knowledge still exhibit limited efficacy, even with the provision of such information. Conversely, heuristics-based approaches, while broader in applicability, fall short in achieving precise segmentation. Our proposed framework merges heuristics with the segmentation model, thus extracting the strengths of both techniques. This integration allows for the effective separation of transient distractors within static scenes, independently of any prior knowledge.

\end{document}